\documentclass{article}

\usepackage{graphicx, color}

\usepackage[a4paper,margin=2cm]{geometry}

\usepackage{caption}
\usepackage{subcaption}
\usepackage{hyperref}
\usepackage{amsmath}
\usepackage{amssymb}
\usepackage{algorithm}
\usepackage{algorithmicx}
\usepackage{algpseudocode}
\usepackage{layouts}
\usepackage[table,xcdraw]{xcolor}
\usepackage{multirow}
\usepackage{graphics}
\usepackage{graphicx}
\usepackage{pdfpages}
\usepackage[toc,page]{appendix}
\usepackage{longtable}
\usepackage{blindtext}
\usepackage{afterpage}

\RequirePackage[backend=bibtex,style=nature,url=false,eprint=false,maxnames=1,isbn=false, doi=false]{biblatex}
\AtEveryBibitem{%
  \clearfield{note}%
}
\AtEveryBibitem{%
  \clearlist{language}%
}
\DeclareFieldFormat*{title}{#1}
\DeclareFieldFormat[book]{title}{#1}

\usepackage{scalerel,xparse}
\NewDocumentCommand\emojisun{}{
    \includegraphics[scale=0.11]{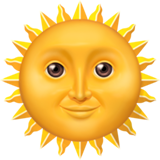}
}

\let\oldabstract\abstract
\let\oldendabstract\endabstract
\makeatletter
\renewenvironment{abstract}
{%
               {\list{}{\addtolength{\leftmargin}{3em} % change this value to add or remove length to the the default
                        \listparindent 1.5em%
                        \itemindent    \listparindent%
                        \rightmargin   \leftmargin%
                        \parsep        \z@ \@plus\p@}%
                \item\relax}%
               {\endlist}%
\oldabstract}
{\oldendabstract}
\makeatother

\begin{document}

\begin{titlepage}

\newcommand{\HRule}{\rule{\linewidth}{0.5mm}} % Defines a new command for the horizontal lines, change thickness here

\center % Center everything on the page

%----------------------------------------------------------------------------------------

%	HEADING SECTIONS

%----------------------------------------------------------------------------------------

\includegraphics[width=\linewidth]{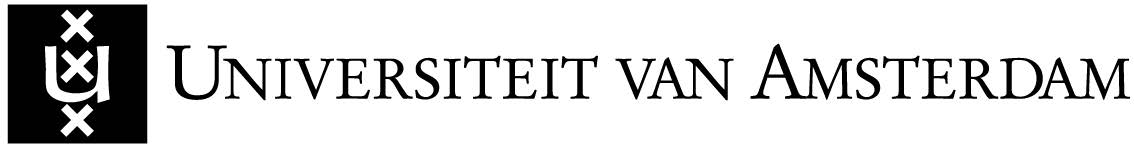}\\[2.5cm]

\textsc{\Large MSc Artificial Intelligence}\\[0.2cm]

\textsc{\Large Master Thesis}\\[0.5cm]

%----------------------------------------------------------------------------------------

%	TITLE SECTION

%----------------------------------------------------------------------------------------

\HRule \\[0.4cm]

{ \huge \bfseries Knowledge Generation \\ \Large Variational Bayes on Knowledge Graphs \\ [0.4cm] } % Title of your document

\HRule \\[0.5cm]

%----------------------------------------------------------------------------------------

%	AUTHOR SECTION

%----------------------------------------------------------------------------------------

by\\[0.2cm]

\textsc{\Large Florian Wolf}\\[0.2cm] %you name

{12393339}\\[1cm]

%----------------------------------------------------------------------------------------

%	DATE SECTION

%----------------------------------------------------------------------------------------

{\Large \today}\\[1cm] % Date, change the \today to a set date if you want to be precise

{48 Credits}\\ %
{April 2020 - January 2021}\\[1cm]
%{Period in which the research was carried out}\\[1cm]%

%----------------------------------------------------------------------------------------

%	COMMITTEE SECTION

%----------------------------------------------------------------------------------------

\begin{minipage}[t]{0.4\textwidth}

\begin{flushleft} \large

\emph{Supervisor:} \\

Dr Peter \textsc{Bloem} \\ Thiviyan \textsc{Thanapalasingam} \\ Chiara \textsc{Spruijt} % Supervisor's Name

\end{flushleft}

\end{minipage}

~

\begin{minipage}[t]{0.4\textwidth}

\begin{flushright} \large

\emph{Assessor:} \\

Dr Paul \textsc{Groth}\\

\end{flushright}

\end{minipage}\\[2cm]

%----------------------------------------------------------------------------------------

%	LOGO SECTION

%----------------------------------------------------------------------------------------

% \framebox{\rule{0pt}{2.5cm}\rule{2.5cm}{0pt}}\\[0.5cm]
% \begin{figure}[H]
%     \centering
%     \begin{minipage}{.5\textwidth}
%         \includegraphics[width=2.5cm, left]{data/images/uva.png} % Include a department/university logo - this will require the graphicx package
%     \end{minipage}
%     \right
\begin{minipage}{\textwidth}
    \centering
    \includegraphics[height=5cm]{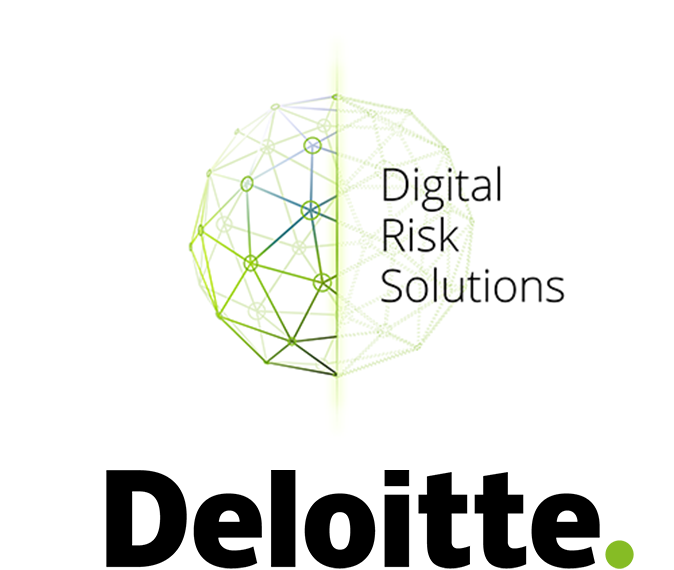} % Include a department/university logo - this will require the graphicx package
\end{minipage}
% \end{figure}

% \textsc{\large \red{institute name}}\\[1.0cm] % 

%----------------------------------------------------------------------------------------

\vfill % Fill the rest of the page with whitespace

\afterpage{\null\newpage}
\pagenumbering{gobble}

\end{titlepage}

\newpage
\pagenumbering{roman}

\begin{abstract}
% We generate Knowledge! \cite{kipf_contrastive_2020}

This thesis is a proof of concept for the potential of Variational Auto-Encoder (VAE) on representation learning of real-world Knowledge Graphs (KG). Inspired by successful approaches to the generation of molecular graphs, we experiment and evaluate the capabilities and limitations of our model, the Relational Graph Variational Auto-Encoder (RGVAE), characterized by its permutation invariant loss function. The impact of the modular hyperparameter choices, encoding though graph convolutions, graph matching and latent space prior, are analyzed and the added value is compared. The RGVAE is first evaluated on link prediction, a common experiment indicating its potential use for KG completion. The model is ranked by its ability to predict the correct entity for an incomplete triple. The mean reciprocal rank (MRR) scores on the two datasets FB15K-237 and WN18RR are compared between the RGVAE and the embedding-based model DistMult.
To isolate the impact of each module, a variational DistMult and a RGVAE without latent space prior constraint are implemented. We conclude, that neither convolutions nor permutation invariance alter the scoring. The results show that between different settings, the RGVAE with relaxed latent space, scores highest on both datasets, yet does not outperform the DistMult.

Graph VAEs on molecular data are able to generate unseen and valid molecules as a results of latent space interpolation. We investigate, if the RGVAE can yield similar results on relational KG data. The experiment is twofold, first the distance between the latent representation of two triples is linearly interpolated, then each latent dimension is explored in a $95$\% confidence interval of its Normal distribution. The interpolations reveal, how successful the RGVAE is at disentangling the latent space and assigning each latent dimension to data characterizing features. Both interpolation experiments show that the RGVAE learns to reconstruct the adjacency matrix but fails to disentangle and assign the right node and edge attributes. 

The assumption of an uninformative latent representation is confirmed in the last experiment of knowledge generation. For this experiment, we present a new validation method for generated triples from the FB15K-237 dataset. The relation type-constrains of generated triples are filtered and matched with entity types. The observed rate of valid generated triples is insignificantly higher than the random threshold. All generated and valid triples are unseen in both train and test set. A comparison between different latent space priors, using the $\delta$-VAE method, reveals insights in the behavior of the RGVAE's parameter distribution and indicates a decoder collapse. Finally we analyze the limiting factors of our approach compared to molecule generation and propose solutions for the decoder collapse and successful representation learning of multi-relational KGs. 
\end{abstract}
\newpage

\renewcommand{\abstractname}{Acknowledgements}
\begin{abstract}
%  Thanks Mum!
  Like the moon, who needs the sun to shine, I am but a reflection of the beautiful souls who supported me throughout this thesis.

I am forever grateful for the support from Deloitte, for the kindness, feedback and inspiration I received from my wonderful colleagues at Digital Risk Solutions. Especially I want to thank Patrick Hafkenscheid for his guidance and Marc Vandonk for believing in my talent.

I thank Peter Bloem and Thiviyan Thanapalasingam for putting me on the right track and for their unlimited patience during the supervision of this thesis. Further thank-yous to my classmates Kiara, Niels and Basti for our fruitful discussions and for helping me overcome my academic self-doubts. Gratitude also to the UvA for being the Hogwarts of my childhood dreams.

Infinito agradecimiento a Papi y Mami, no solo por el apoyo durante \'{e}stos \'{u}ltimos meses, pero por las oportunidades, la educac\'ion y el amor que me dieron toda mi vida ¡Los amo mucho!

Finally to my girlfriend, who kept me company in those endless nights of writing, who prepped my meals when I couldn't, who woke up early every day just to wish me good morning. 
No matter if highs or lows, crying or laughing, I love every single moment with you. 

\begin{center}
    Love you Renuka, you are my sunshine. \\\emojisun 
\end{center}
\end{abstract}

\newpage
\tableofcontents
\newpage
\listoffigures
\listoftables
\newpage
\pagenumbering{arabic}

\section{Introduction}

% Here comes a beautiful introduction. %Promise!

To begin with, we shall clarify the intended ambiguity of this work's title. Before approaching the question behind this thesis in the context of Representational Learning (RL), we would like to present it from a philosophical point of view of AI safety and potential.
One could argue, that we live in times of the fastest advances in science and technology in the history of humanity, therefore making us the \textit{Knowledge Generation}. While we continuously keep researching and accumulating knowledge, we have historically not been willing to share our knowledge with any other species in this Universe. All scientific milestones are from us, for us. 

The rise of AI marked a turning point of this tradition. For the first time we invest in sharing our knowledge and with systems which can act on superhuman dimension. While machine learning models might not be considered a species, they have learned to drive our cars and in fact beaten human intelligence in the game GO \cite{silver_mastering_2017}. This is not only seen as progress but also as danger. The world's richest man, Elon Musk, both build his fortune on AI and respects it as humanity's biggest risk. 

\begin{figure}[H]
    \centering
    \includegraphics[height=.21\textwidth, keepaspectratio]{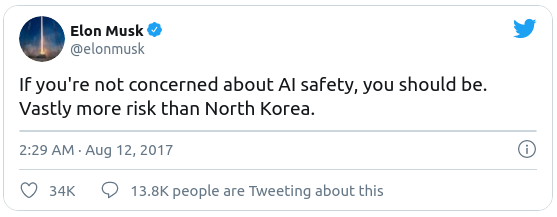}
    \caption{Elon Musk in a tweet on AI. Source \cite{noauthor_elon_nodate}}
    \label{fig1:Elon}
\end{figure}

% \begin{figure}[H]
%     \centering
%     \includepdf[pages=-,pagecommand={},width=\textwidth]{data/images/Elon75a4.pdf}
%     \caption{Elon Musk Tweet on AI. Source \cite{twitter}}
%     \label{fig1:Elon}
% \end{figure}

Closing the circle to the ambiguity of the title and relating to the popular concern, that AI might reach a point where it does not need humans anymore to keep evolving, we ask the crucial question: Can AI generate knowledge?

\subsection{Motivation}

A key area of AI is RL, where the model learns to identify and disentangle characteristics and features of the data. Understanding the semantics of the data is specifically important for unsupervised learning of generative models. 
% computer vision
The task of generating data has been widely explored for images. Computer vision has reached a point where a simple image can be semantically segmented, where objects can be detected and classified and even relations between entities inferred \cite{kipf_contrastive_2020}.

% molecular graphs
Advances in the parallel field of graph generation have received less attention, yet showed promising results. Data stored as a graph has a high density of information and rich semantics, which makes it attractive for variational inference. The recent success of Simonovsky et al. \cite{simonovsky_graphvae_2018} on the generation and completion of molecules represented in graph structure, initially inspired our research. Next to molecules, graphs can be used to store knowledge. While real world Knowledge Graphs (KG) have a far higher complexity than molecule graphs, the proposed generative model, Variational Auto-Encoder (VAE) also has proven its capacity to learn from huge datasets with high variance. Inspired by Simonovsky work and motivated by the vision of generating knowledge, we explore the possibilities and limitation of KG generation with VAEs.

\subsection{Expected Contribution}

The main contributions of this thesis are threefold. The main objective is to proof the hypothesis that a graph VAE can capture, disentangle and reproduce the underlying semantics of a real world KG, secondly. Further we contribute with a novel implementation of a graph matching algorithm and a validation method for generated triples. 

In Simonovsky \cite{simonovsky_graphvae_2018} work, small subgraphs with multiple edges are used to represent molecule graphs. In contrast we proof the hypothesis by generating the smallest possible graph of two nodes, also representable as single triple. The VAE is tested and evaluated in several experiments, including link prediction, latent space interpolation and accuracy of generating valid triples.
% Proof of concept KGVAE

% draw parallels to molecules and explore the differences
We compare our results to related KGs methods and investigate the impact of different hyperparameter. The main focus is be on the influence of the graph matching loss function, the encoding through graph convolutions and stochastic inference. Further, we aim to reproduce the success of molecule generation and point therefore continuously the similarities and differences to our work out.

% Implementation of graph matching in batches
On a lower level, we hope to contribute with our implementation of Cho's \textit{et al.} max-pooling graph matching algorithm for tensor batches. While the algorithm has been cited and implemented numerous times, a working implementation, compatible with deep learning libraries, has to the best of our knowledge not yet been published.  

% Method for syntax cohearence of generated triples from real world KG
Lastly we introduce a high-level method for evaluating the validity of generated data, which compares the type constrain of the generated triple's predicate with its entity types and reports accuracy. This is made possible by expanding the existing dataset FB15K-237 with entity types from its original KG Freebase. While this scoring method is error prone, it does give an insight into the level representational potential of the model and the syntax coherence of the generated triples. Future work can use this evaluation method to track progress and compare to the baseline.

% This thesis is aimed to be a proof of concept, providing insight into the capability of the VAE on generating KG triples and to indicate if further research in this direction would be meaningful.

\subsection{Research Question}

\begin{center}
    \texttt{How successful is a graph VAE in representation learning of real world KG compared to molecule graph data and what is the impact of each major hyperparameter?}
    \label{sec1:requestion}
\end{center}

% \emojismile

% Without further ado -

\section{Related Work}
This section presents previous work which inspired and laid the foundation for this thesis. Relevant publication on topics related to this thesis are presented in terms of methods and results. We focus on the fields of relational graph convolutions, graph encoders, and embedding-based link prediction.

\subsection{Relational Graph Convolutions}
We define a graph as $G=(\mathcal{V}, \mathcal{E})$  with a set of nodes $\mathcal{V}$ and a set of edges $\mathcal{E}$. The set of edges, with each edge connecting node $x$ and $y$, is defined by $\left\{(x, y) \mid(x, y) \in \mathcal{V}^{2} \wedge x \neq y\right\}$ while the constraint $x \neq y$ prohibits self-connections or self-loops, which is optional depending on the graphs function. Moreover, nodes and edges can have features, which contribute additional information about the nodes and their connection. In the literature, these features can be describing attributes and properties, in the context of this work we also use them as indicators to unique entities. Graph convolutions, make use of both these properties and the spectral information in a graphs adjacency matrix. In graph theory spectral properties are the characteristic polynomial, eigenvalues, and eigenvectors of the Laplacian and adjacency matrix \cite{chung1997spectral}. Two popular tasks to evaluate the performance of a neural network on graphs, are node classification and link prediction. The first is a classification problem where the model predicts the class of a node. Link prediction is the task of completing a triple by correctly predicting the missing entity at either head or tail the triple. A in-depth explanation follows in section \ref{ssec4:lpmetrics}.

% Present the relational graph convolution model paper by Kipf and maybe others
In Kipf \textit{et al.} paper on graph convolutions \cite{kipf_semi-supervised_2017} a novel Graph Convolution Network (GCN) for semi-supervised classification is introduced. The model takes as input the adjacency matrix and optionally a feature matrix of the graph and predicts the classes of the nodes. Graph convolutions acts directly on the graph structure and are linearly scalable with the number of nodes. The GCN takes as input the adjacency matrix $A \in \mathbb{R}^{n \times n}$ with $n$ being the number of nodes in the graph. In the case of undirected graphs, the adjacency matrix is symmetric. The output is a matrix $H \in \mathbb{R}^{n \times d_h}$ where $d_h$ are the hidden dimensions or in case od the last layer, the number of classes to predict over. 
While the authors compare different propagation methods for the graph convolutions, their propagating rule using a first-order approximation of spectral graph convolutions, outperforms all other implementations. Propagation denotes the transformation of the input data between layers of a model. Kipf approximates the eigenvalues of the Laplacian with first order Chebyshev polynomials and circumvents the computationally expensive Eigendecomposition. The renormalization trick normalized the adjacency matrix and adds it to an identity matrix of same size. This keeps the eigenvalues in a range between $[0,2]$ which again leads to a stable training, avoiding numerical instabilities and vanishing gradients during learning. Additionally the feature information of neighboring nodes is propagated in every layer what shows improvement in comparison to earlier methods, where only label information is aggregated.
Kipf and Welling perform node classification on the three citation-network datasets, Citeseer, Cora and Pubmed as well as on the KG dataset NELL. In all classification tasks, their results outperform other recently proposed methods in this field and proves to be computationally more efficient than its competition. For more details on the implementation of graph convolutions we refer to section \ref{ssec:gcn}. 

\begin{figure}[h]
    \centering
    \includegraphics[width=0.55\textwidth]{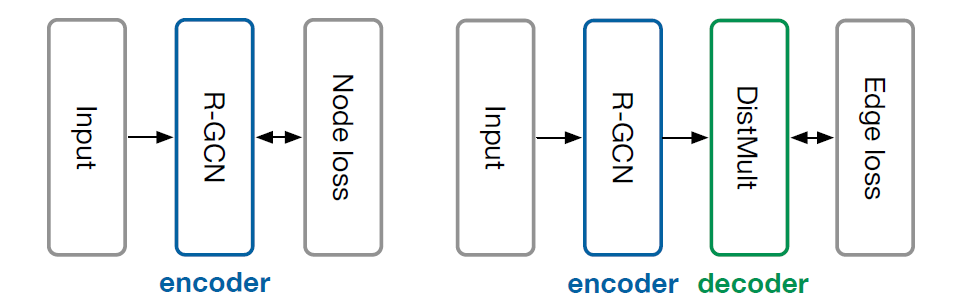}
    \caption{RGCN with encoder-only for node classification and encoder-decoder architecture for link prediction experiments. Source \cite{gangemi_modeling_2018}.}
    \label{fig:RGCN}
\end{figure}

% Kipfs second paper 
In their publication \textit{Modeling Relational Data with Graph Convolutional Networks} Schlichtkrull \textit{et al.} propose a relational graph convolutional network (RGCN) and evaluate it on link prediction on the FB15K-237 and WN18 dataset and node classification on the AIFB, MUTAG, BGS and AM datasets \cite{gangemi_modeling_2018}. The RGCN, with its encoder properties, is used by itself as node classifier, yet for link prediction it is coupled with a DistMult model acting as decoder which scores triples encoded by the RGCN see figure \ref{fig:RGCN}. We go into details of the embedding-based DistMult model in section \ref{ssec:embedlp}.

The RGCN works on dense graphs stored as triples, creating a hidden state for each node. A novel message passing network is layer-wise propagated with the hidden states of the entities. As regularization the authors propose a \textit{basis-} and \textit{block-wise} decomposition. while the first  aims at an effective weight sharing between different relation types, the second  can be seen as a sparsity constraint on the relation type's weight. The model outperforms embedding based model on the link prediction task on the FB15K-237 dataset and scores competitive on the WN18 dataset. In the node classification task, the model sets state of the art results on the datasets AIFB and AM, while scoring competitive on the remaining. The authors conclude, that the model has difficulties encoding higher-degree hub nodes on datasets with many entities and low amount of classes. This is noticeable as it relates to the WN18RR \cite{battaglia_relational_2018}, one of the two datasets used in this thesis.

\subsection{Graph VAE}
% Present different papers with graph VAEs
We have seen how graph convolutional neural networks can be combined in an encoder-decoder architecture, resulting in a generative model suitable for unsupervised learning. We present three recent publications with different methods and use cases of a graph generative model, in particular a VAE.

\begin{figure}[h]
    \centering
    \includegraphics[width=0.5\textwidth]{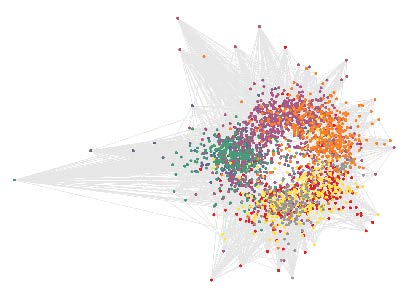}
    \caption{Visualization of the VGAE's latent representation of the Core citation network. Colors express the disentanglement of node classes. Source \cite{kipf_variational_2016}}
    \label{fig:kipfGVAE}
\end{figure}

% Kipfs VGAE
Kipf \textit{et al.} introduce the the Variational Graph Autoencoder (VGAE), a framework for unsupervised learning on graph-structured data \cite{kipf_variational_2016}. This generative model uses a GCN as encoder and a simple inner product module as decoder. Similar to the GCN, the VGAE incorporates node features, which significantly improves its performance on link prediction tasks compared to related models. The VGAE uses a two-layer GCN to encode the mean and the logvariance for the stochastic module to sample the latent space representation, more specifically, a latent vector per node. Referring to the above described GCN, the VGAE encoder outputs a latent matrix $H \in \mathbb{R}^{n \times 2d_z}$ with $d_z$ denoting the latent dimension. The activation of the inner product of this latent matrix yields the reconstruction of the adjacency matrix. Figure \ref{fig:kipfGVAE} shows how the model learns to cluster  nodes according to their class, without these labels being provided to the model during training.
This visualization shows that the VGAE learns successful learn an implicit representation of the data.
The VGAE with added features outperforms state of the art methods (to the time of publication) Spectral Clustering \cite{tang2011leveraging} and Deepwalk \cite{perozzi2014deepwalk} in the task of link prediction on the datasets Cora, Citeseer and Pubmed. The authors point out, that a Gaussian prior might be a poor choice combined with the inner-product decoder.

% TODO I think it's important to state the difference between GVAE and GraphVAE very clear (that is the GVAE learns a latent vector per node and the GraphVAE learn a lv for the whole graph).

\begin{figure}[h]
    \centering
    \includegraphics[width=0.9\textwidth]{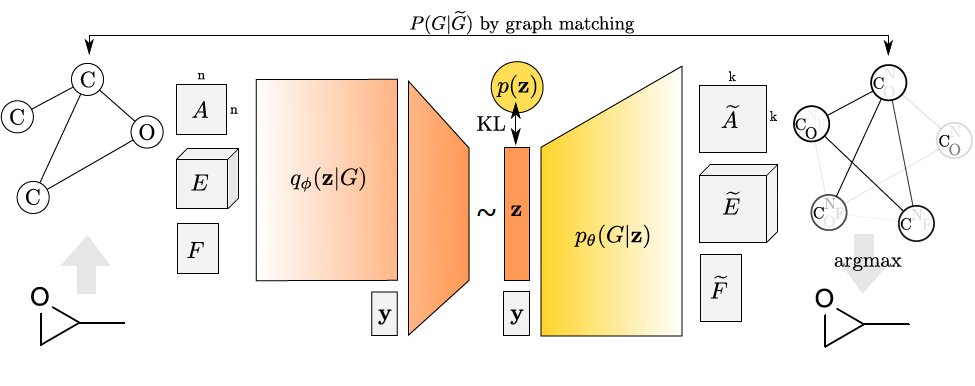}
    \caption{Model architecture of the GraphVAE. Source \cite{simonovsky_graphvae_2018}.}
    \label{fig:graphvaefull}
\end{figure}

Simonovsky \textit{et al.} introduce the GraphVAE, which generates a probabilistic fully-connected molecule graph of a predefined maximum size
in a one-shot approach \cite{simonovsky_graphvae_2018}. In this context fully-connected denotes that all nodes are connected within a graph, in contrast to citation networks where subgraphs can be disconnected from each other. While molecule graphs have a lower node and edge count than citation networks, their edges and nodes are attributed, which constrains each connection. The model includes a standard graph matching algorithm, which finds the optimal permutation between the predicted graph and the ground truth and. The reconstruction loss considers the permutation instead of the raw prediction. In contrast to the previously presented publications, the input to this model is a threefold and sparse graph, defined as $G=(A, E, F)$ with $A$ being the adjacency matrix, $E$ the edge attribute matrix and $F$ the node attribute matrix, with $E$ and $F$ being one-hot encoded. Considering that this method lays the foundation for this thesis, we adopt this notation for our own methods in section \ref{sec:mthods}. Figure \ref{fig:graphvaefull} shows the architecture of the GraphVAE. The encoder is a feed forward network with edge-conditioned graph convolutions \cite{simonovsky2017dynamic}, which takes as input the target graph $G$ with $n$ nodes. After the convolutions the result is flattened and conditioned on the node labels $y$. A fully-connected neural network encodes the stochastic latent representation, which is constrained by Standard Gaussian prior distribution. Note that in contrast to the GCN, which encodes one latent vector per node, the GraphVAE instead encodes a latent representation of the whole graph. This latent representation is again conditioned on the node labels $y$ and propagated through the decoder in form of a fully-connected neural network. The decoder reconstructs the latent representation to the graph prediction. The threefold decoder output is matched with the target using graph matching algorithm, which we discuss further in section \ref{ssec:graphmatch}. The matched and permuted graph is then used for the reconstruction term of the GraphVAE loss. It should be noted, that, while the size of the target and prediction graph are fixed, they do not necessarily have to match. While this approach seems promising, it is limited by the maximum graph size, which has been experimented with up to a node count of $40$.

The model is trained on the QM9 dataset, containing the graph structure of 134k organic molecules with experiments on latent space dimension in the range of $[20,80]$. On the free generation task, about $50\%$ of the generated molecules are chemically valid and thereof remarkably $60\%$ are not included in the training dataset. When testing the model for robustness, it showed little disturbance when adding Gaussian noise to the input graph $G$. The authors conclude that the problem of generating graphs
from a continuous embedding was addressed successfully and that the GraphVAE performs better on small molecules, implying a low node count per graph.

Until here the presented models generate graphs in a single propagation through the model. For completeness we also present a successful approach for graph generation in autoregressive manner. Belli \textit{et al.} introduce named approach for image-conditioned graph generation of road network graphs \cite{belli_image-conditioned_2019}. While we focus on the generative model, their contribution ranges wider, namely the introduction of the graph-based roadmap dataset \textit{Toulouse Road Network} and the task specific distance metric \textit{StreetMover}. The authors propose the Generative Graph Transformer (GGT) a deep autoregressive model that makes use of attention mechanisms on images, to tackle the challenging task of road network extraction from image data. The GGT has a encoder-decoder architecture, with a CNN as encoder, taking the grayscale image as input signal and predicting a conditioning vector. The decoder is a self-attentive transformer, which takes as input the encoded condition vector and a hidden representation of the adjacency matrix $A$ and feature vector $X$ of the previous step. The adjacency matrix here indicates the links between steps and the features are normalized coordinates. A multi-head operator outputs the hidden representation of $A$ and $X$ which finally are decoded by a MLP to the graph representation. For the first step, a empty hidden representation is fed into the decoder. The model terminates the recurrent graph generation by predicting a end-of-sequence token, which signalizes the end of the graph. During learning, the generated graphs are matched to the target graphs using the \textit{StreetMover} metric, based on the Sinkhorn distance. The authors attribute \textit{StreetMover} as a scalable, efficient and permutation-invariant metric for graph comparison. The successful results of the experiments performed, show that this novel approach is suitable for the task of road network extraction and could yield similar success in graph generation task of different fields.
While this publication does not directly align with the previously presented work, we find it of added value to present alternative approaches on our topic.

% \begin{itemize}
%     \item Belli recurrent VAE
%     \item GraphVAE paper
%     \item Variational Graph Auto-Encoders
% \end{itemize}

\subsection{Embedding-Based Link Prediction}
\label{ssec:embedlp}
Finalizing this chapter, we look at embedding-based methods on KGs.
Compared to the previously presented research, embedding models have a much simpler architecture and can be trained computationally very efficient on large graphs. Embedding-based models can only operate on triples, meaning a KG is represented as a set of triples with indices pointing to the unique entity and relation in the graph. Despite their simplicity, they achieve great results on node and link prediction tasks.

Already in 2013 Bordes \textit{et al.} introduced in their paper \textit{Translating Embeddings for Modeling Multi-relational Data} the low-dimensional embedding model TransE \cite{bordes_translating_2013}. The core idea of this model is that relations can be represented as translations in the embedding space. Entities are encoded to a low-dimension embedding space and the relation is represented as vector between the head and tail entity. The assumption is that for correct triples the model learns to reduce the Euclidean distance between head and tail entity by placing them closer together in the embedding space. This results in correct triples having a lower norm of the relational vector than corrupted triples. Using this property, the model can predict the missing entity in link prediction.

The models loss function takes a set of corrupted triples for every triples in the training set and subtracts the translation vector of the corrupted triple in embedding space from the translation vector of the correct triple with added margin. To minimize the loss, the model has to place entities of correct triples closer together in embedding space. We think of a triple as $(s,r,o)$ and $(e_s,e_r,e_o)$ as its embedded representation, $d()$ the Euclidean distance, $\gamma$ the positive margin and $S$ and $S^{\prime}$ as sets of correct and corrupt triples, the loss function of TransE is 

\begin{equation}
    \mathcal{L}=\sum_{S} \sum_{S^{\prime}} \left[\gamma+d(\boldsymbol{s}+\boldsymbol{r}, \boldsymbol{o})-d\left(\boldsymbol{s}^{\prime}+\boldsymbol{r}, \boldsymbol{o}^{\prime}\right)\right].
\end{equation}

The model is trained on a subset of the KGs Freebase and Wordnet, which is also the source for the datasets used in this thesis. TransE's link prediction results on both head and tail outperformed other competing methods of the time, such as RESCAL \cite{nickel_three-way_nodate}.

In 2015, Yang \textit{et al.} proposed a similar, yet better performing KG embedding method \cite{yang_embedding_2015}. Their model DistMult captures relational semantics by matrix multiplication of the embedded entity representation and uses a bilinear learning objective. The main difference to TransE is the bilinear scoring function $d^{b}()$. Bilinear is indicated by the exponent $b$ and connotes the functions score-invariance of swapping the triples head and tail entity. For the embedding space representation of subject and object $e_s$ and $e_{o}$ and a diagonal matrix $\operatorname{diag}(e_{r})$ with the embedded relation $e_r$ on the diagonal, the scoring function is

\begin{equation}
    d^{b}\left((e_s,e_r,e_o)\right)=e_s \operatorname{diag}(e_{r}) e_o.
    \label{eq2:distmult}
\end{equation}

The publication goes on to explore the options of embedding-based rule extraction from KGs. Concluding, the authors state that the prediction scores achieved with the embeddings learned from the bilinear objective not only outperform the state of the art in link prediction but can also capture compositional semantics of relations and extract Horn rules using compositional reasoning.

In a more recent publication, Ruffinelli \textit{et al.} present a comprehensive review of KG embedding models such as TransE and DistMult, coupled with state of the art techniques in deep learning. The authors start by pointing out the similarities and differences of most models. While all methods share the same embedding approach, they differ in their scoring function and their original hyperparameter search. The authors perform a quasi-random hyperparameter search on the five models RESCAL, TransE, DistMult, ComplEx and ConvE, which each use a characteristically different loss function. They are compared by their MRR and Hits@$10$ scores on the two datasets FB15K-237 and WN18. Since these metrics and datasets are used later on in our research, they are explained in section \ref{ssec5:data}(datasets) and \ref{ssec4:lpmetrics}(metrics). The tuned models report a higher MRR score of up to $24\%$ compared to their first reported performance. The authors conclude that simple KG embedding methods can show strong performance when trained with state of the art techniques what indicates that higher complexity is not necessary. The optimal model configurations, which were found by a random search of the hyperparameter space, are included in this publication. 

% Graph Embeddings\\
% TransE represents entities in in low-dimensional embedding. The relationships between entities are represented by the vector between two entities \cite{bordes_translating_2013}.
% (How are different relation between the same entities represented?)

% OntoUSP\\
% This method learns a hierarchical structure to better represent the relations between entities in embedding space.

\section{Background}

% In this section we go over related work and relevant background information for our model and experiments. The depth of the explanation is adopted to the expected prior knowledge of the reader. The reader is supposed to know the basics of machine learning and deep learning, including probability theory and basic knowledge on neural networks and their different architectures. Basic principles such as forward pass, backpropagation and convolutions are expected to be understood. Further the use and functionality of deep learning modules such as the model, the optimizer and the terms target and prediction should be known. This also includes being familiar with the training and testing pipeline of deep learning.

This section derives and explains the techniques, which form the backbone of this research. For fundamental background on machine learning and probability theory we refer the reader to Bishops book \cite{bishop_pattern_2006}. We begin with introducing the VAE and its differences to a normal autoencoder. Further we show how convolutional layers can act on graphs and how these layers can be used in an encoder model, the GCN. Building on these modules, we present the main model for this thesis, the Relational Graph VAE (RGVAE). Finally, we present a popular graph matching algorithm, which is intended to match prediction and target graph \cite{paulheim_knowledge_2016}.

\subsection{Knowledge Graph}

% Knowledge Graphs are great! The best in the world.
% Knowledge Graphs which we be using
% We focus on the generation of KGs.
% Representation of KG as adjacency, edge feature and node feature matrix
'Knowledge graph' has become a popular term. Yet, the term is so broad, that its definitions varies depending on domain and use case \cite{ehrlinger2016towards}. In this thesis we focus on KGs in the context of relational machine learning.

% What is a KG
Both datasets of this thesis are derived from large-scale KG data in RDF format. The Resource Description Framework (RDF), originally introduced as infrastructure for structured metadata, is used as general description framework for the semantic-web applications \cite{miller1998introduction}. It involves a schema-based approach, meaning that every entity has a unique identifier and possible relations limited to a predefined set. The opposite schema-free approach is used in OpenIE models, such as AllenNLP \cite{gardner_allennlp_2018}, for information extraction. These models generate triples from text based on NLP parsing techniques which results in an open set of relations and entities. In this thesis a schema-based framework is used and triples are denote as $(s,r,o)$. A exemplar triple from the FB15K-237 dataset is

\begin{center}
    \texttt{/m/02mjmr, /people/person/place\_of\_birth, /m/02hrh0\_}.
\end{center}

A human readable format of the entities is given by the \textit{id2text} translation of Wikidata \cite{vrandevcic2014wikidata}.

\begin{itemize}
    \item Subject $s$:   \texttt{/m/02mjmr Barack Obama}
    \item Relation/Predicate $r$:    \texttt{/people/person/born-in}
    \item Object $o$:     \texttt{/m/02hrh0\_ Honolulu}
\end{itemize}

For all triples, $s$ and $o$ are part of a set of entities, while $r$ is part of a set of relations. This is sufficient to define a basic KG \cite{nickel_review_2016}.

% Hierarchy, entities, classes
Schema-based KG can include type hierarchies and type constraints. Classes group entities of the same type together, based on common criteria, e.g. all names of people can be grouped in the class 'person'. Hierarchies define the inheriting structure of classes and subclasses. Picking up our previous example, 'spouse' and 'person' would both be a subclass of 'people' and inherit its properties. At the same time the class of an entity can be the key to a relation with type constraint, since some relations can only be used in conjunction with entities fulfilling the constraining type criteria.

% Ontology - semantics
These schema based rules of a KG are defined in its ontology. Here properties of classes, subclasses and constraints for relations and many more are defined. Again, we have to differentiate between KGs with open-world or closed-world assumption. In a closed-world assumption all constraints must be sufficiently satisfied before a triple is accepted as valid. This leads to a huge ontology and makes it difficult to expand the KG. On the other hand open-world KGs such as Freebase, accept every triple as valid, as long as it does not violate a constrain. This again leads inevitably to inconsistencies within the KG, yet it is the preferred approach for large KGs. In context of this thesis we refer to the ontology as semantics of a KG, we research if our model can capture the implied closed-world semantics of an open-world KG \cite{nickel_review_2016}.

% % Sparse and dense representation
% Lastly, we point out one major difference between KGs, namely their representation. RDF KGs are represented as set of triples, consisting of a unique combination of numeric indices. Each index linking to the corresponding entry in the entity and relation vocabulary. This is called dense representation and benefits from fast computation due to an optimized use of memory.

% In contrary the dense representation of a triple is the sparse representation. Here a binary square matrix also called the adjacency matrix, indicates a link between two entities. To identify the node, each node in the adjacency matrix has a one-hot encoded entity-vocabulary vector. All one-hot encoded vectors are concatenated to a node attribute matrix.
% In simple cases, like citation networks this is a sufficient representation. In the case of Freebase, we need an additional edge-attribute matrix, which indicates the relation  of each link. The main benefit of this method is the representation of subsets of triples, also referred to as subgraphs, with more than one relation and the possibility to perform graph convolutions. 

% Usecases of KG

% Knowledge graphs have very different formats. The datasets we be sorting with are in rdf format.
% This format can include an defined onthology or not.
% This means the KG consists of triples subject, relation, object.
% when indexing these triples. we get a dense representation of the KG.
% About sparse KGs

\subsection{Graph VAE}

Since the graph VAE is a adaptation of the original VAE, we start by introducing the original version, which is unrelated to graph data. Furthermore we present each of the different modules, which compose the final model. This includes the different graph encoders as well as sparse graph loss functions. We define the notation upfront for this chapter, which touches upon three different fields. For the VAE and MLP we consider data in vector format. A bold variable denotes the full vector, e.g. $\mathbf{x}$ and variable with index denotes the element at that index, e.g. for the vector element at index $i$ we denote $x_i$. Graphs are represented in matrices, which are denoted in capital letters. $A$ typically denotes the adjacency matrix, from there on paths split and we use different notations for different methods. While $X$ is described as feature matrix, we have to specify further when it comes to Simonovsky's GraphVAE, where $E$ is the edge attribute and $F$ the node attribute matrix. The reason we change from features to attributes, is due to the singularity, also one-hot encoding of attributes per node/edge, in contrast to features which can be numerous per node.

\subsubsection{VAE}
\label{ssection:VAE}
% The VAE as first presented by \cite{kingma_auto-encoding_2014} is an unsupervised generative model consisting of an encoder and a decoder. The architecture of the VAE differs from a common autoencoder by having a stochastic module between encoder and decoder. Instead of directly using the output of the encoder, a distribution of the latent space is predicted from which we sample the input to the decoder. The reparameterization trick allows the model to be differentiable. By placing the sampling module outside the model we get a deterministic model which can be backpropagated.

The VAE as first presented by \cite{kingma_auto-encoding_2014} is an unsupervised generative model in the form of an autoencoder, consisting of an encoder and a decoder. Its architecture differs from a common autoencoder by having a stochastic module between encoder and decoder. The encoder can be represented as recognition model with the probability $p_{\theta}(\mathbf{z} \mid \mathbf{x})$ with $x$ being the variable we want to inference and $z$ being the latent representation given an observed value of $x$. The encoder parameters are represented by $\theta$. Similarly, we denote the decoder as $p_{\theta}(\mathbf{x} \mid z)$, which given a latent representation $z$ produces a probability distribution for the possible values, corresponding to the input of $x$. This is the base architecture of all our models in this thesis.

The main contribution of the VAE is to be a stochastic and fully backpropagateable generative model. This is possible due to the reparametrization trick. Sampling from the latent prior distribution, creates stochasticity inside the model, which can not be backpropagated and makes training of the encoder impossible. By placing the stochastic module outside the model, it becomes fully backpropagatable. We use the predicted encoding as mean and variance for a Gaussian normal distribution, from which we then sample $\epsilon$, which acts as external parameter and does not need to be backpropagated and updated.

\begin{figure}[h]
    \centering
    \includegraphics[width=0.75\textwidth]{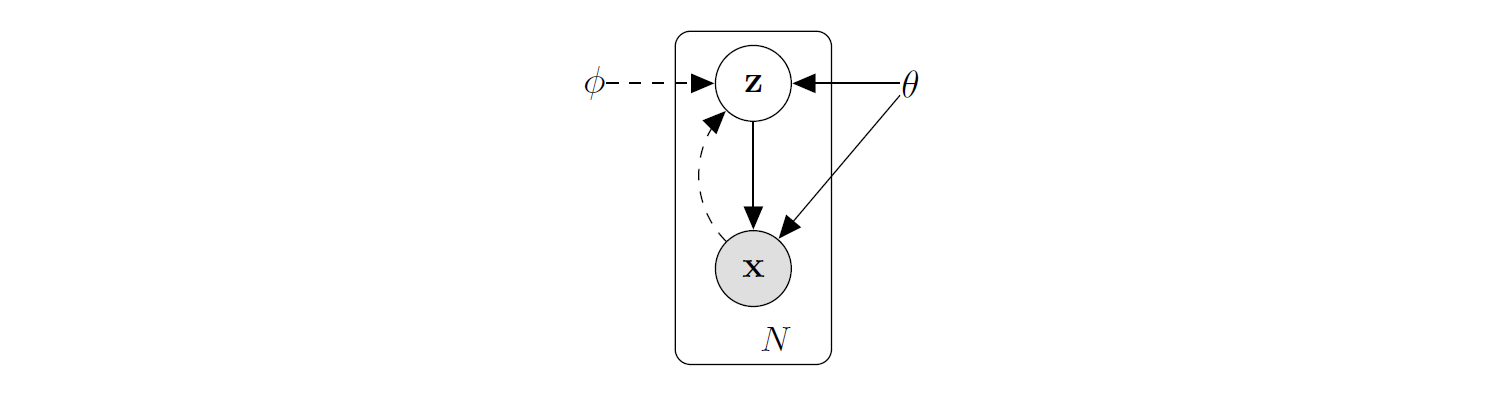}
    \caption{Representation of the VAE as Bayesian network, with solid lines denoting the generator $p_{\theta}(z)p_{\theta}(\mathbf{x} \mid z)$ and the dashed lines the posterior approximation $q_{\phi}(\mathbf{z} \mid \mathbf{x})$ \cite{kingma_auto-encoding_2014}.}
    \label{fig:varinference}
\end{figure}

Figure \ref{fig:varinference} shows, that the true posterior $p_{\theta}(\mathbf{z} \mid \mathbf{x})$ is intractable. To approximate the posterior, we assume a Standard Gaussian prior $p(\mathbf{z})$ with a diagonal covariance, which gives us the approximated posterior

\begin{equation}
    \log q_{\phi}\left(\mathbf{z} \mid \mathbf{x}\right)=\log \mathcal{N}\left(\mathbf{z} ; \mu(\mathbf{x}), \sigma(\mathbf{x})^{2} \mathbf{I}\right) .
\end{equation}
    
Now variational inference can be performed, which allows both $\theta$ the generative and $\phi$ the variational parameters to be learned jointly. Using Monte Carlo estimation of $q_{\phi}(\mathbf{z} \mid x_1)$ we get the variational estimated lower bound (ELBO)

% Monte Carlo estimate implies ----> only one sample, cuz we sample!

\begin{equation}
    \mathcal{L}\left({\theta}, {\phi} ; x_i \right) = -D_{K L}\left(q_{\phi}\left(\mathbf{z} \mid x_i \right) \| p_{\theta}(\mathbf{z})\right) + \mathbb{E}_{q_{\phi}\left(\mathbf{z} \mid x_i\right)}\left[\log p_{\theta}\left(x_i \mid \mathbf{z}\right)\right] .
    \label{eq3:elbo}
\end{equation}

We call the first term the regularization term, as it encourages the approximate posterior to be close to the Gaussian normal prior. This term can be integrated analytically and does not require an estimation. The KL-divergence is a similarity measure between two distributions, resulting in zero for two equal distributions. Thus, the model gets penalized for learning a encoder distribution $p_{\theta}(\mathbf{z} \mid \mathbf{x})$ which is not Standard Gaussian.

Higgins \textit{et al.} present a constrained variational framework, the $\beta$-VAE \cite{higgins_beta-vae_2016}. This framework proposes an additional hyperparameter $\beta$ which acts as factor on the regularization term. The new ELBO including $\beta$ is

\begin{equation}
    \mathcal{L}\left({\theta}, \phi ; x_i\right)=-\beta \left(D_{K L}\left(q_{\phi} \left(\mathbf{z} \mid x_i\right) \| p_{theta}(\mathbf{z})\right)\right)+\mathbb{E}_{q_{\phi} \left(\mathbf{z} \mid x_i \right)} \left[\log p_{\theta}\left(x_i \mid \mathbf{z}\right)\right] .
    \label{eq3:elboBeta}
\end{equation}

For $\beta = 0$ the original VAE is restored and for $\beta>0$ the influence of the regularization term on the ELBO is emphasized. Thus the model prioritizes to learn the approximate posterior $p_{\theta}(\mathbf{z} \mid \mathbf{x})$ even closer to the Standard Gaussian prior. In the literature this results in a disentanglement of the latent space qualitatively outperforms the original VAE in representation learning of image data.

The second term represents the reconstruction error, which requires an estimation by sampling. This means using the decoder to generate samples from the latent distribution. In the context of the VAE, these probabilistic samples are the models output, thus, the reconstruction error the similarity between prediction and target\cite{kingma_auto-encoding_2014}.

Once the parameters $\phi$ and $\theta$ of the VAE are learned, the decoder can be used on its own to generate new samples from $p_{\theta}(\mathbf{x} \mid z)$. Conventionally, a latent input signal is sampled from $\mathbb{N^{d_z}}$ with $d_z$ being the latent dimension. In the case of discrete binary data, each element of the generated sample is used as probability parameter $p$ for a Bernoulli $\mathbb{B}(1,p)$, from which the final output is sampled. In the case of categorical data, e.g. one-hot encoding, the final output is either sampled from a Categorical distribution with the prediction as probability parameters of each class, or simply sampled with the Argmax operator \cite{kingma_introduction_2019}.

\subsubsection{MLP}
\label{ssec:mlp}

% History introduction
The Multi-Layer Perceptron (MLP) was the first of its kind, introducing a machine-learning model with a hidden layer between the input and the output. 
% Invented by who when
Its properties as universal approximator has been discovered and widely studied since 1989.
While we presume that the reader interested in the topic of this thesis does not require a definition of the MLP, it is included for completeness, as we also define the GCN encoder, which both act as encoder and decoder our final model, and contribute different hyperparameters.

% Functionality
% In its basic structure it takes a one dimensional input, fully-connected hidden layer, activation function and finally output layer with normalized predictions.
The MLP takes a linear input vector of the form $x_1,...,x_D$ which is multiplied by the weight matrix $\mathbf{W^{(1)}}$ and then activated using a non-linear function $h(\dot)$, which results in the hidden representation of $\mathbf{x}$. Due to its simple derivative, mostly the rectified linear unit (ReLU) function is used as activation. The hidden units get multiplied with the second weight matrix, denoted $\mathbf{w^{(2)}}$ and finally transformed by a sigmoid function $\sigma(\dot)$, which produces the output. Grouping weight and bias parameter together we get the following equation for the MLP

\begin{equation}
    y_{k}(\mathbf{x}, \mathbf{w})=\sigma\left(\sum_{j=0}^{M} w_{k j}^{(2)} h\left(\sum_{i=0}^{D} w_{j i}^{(1)} x_{i}\right)\right)
\end{equation}

for $j=1, \ldots, M$ and $k=1, \ldots, K$, with $M$ being the total number of hidden units and $K$ of the output.

% Application
Since the sigmoid function returns a probability distribution for all classes, the MLP can have the function of a classifier. Instead of the initial sigmoid function, it was found to also produce good results for multi label classification activating the output through a Softmax function instead. Images or higher dimensional tensors can be processed by flattening them to a one dimensional tensor. This makes the MLP a flexible and easy to implement model \cite{bishop_pattern_2006}.

\subsubsection{Graph convolutions}
\label{ssec:gcn}

% TODO pretty write this
%  Intro on convolutions
Convolutional layers benefit from the symmetry of data, correlation between neighboring datapoints. Convolutional Neural Nets (CNN) are powerful at classification and object detection on image. Neighboring pixel in an image are not independent and identically distributed {i.i.d.) but rather are highly correlated. Thus, patches of datapoints let the CNN infer and detecting local features. The model can further merged those to high-level features, e.g. a face in an image \cite{bishop_pattern_2006}. Similar conditions hold for graphs. Nodes in a graph are not i.i.d. and allow inference of missing node or link labels. 

% How do Graph convs work???
Different approaches for graph convolutions have been published. Here we present the graph convolution network (GCN) of \cite{kipf_semi-supervised_2017}.
We consider $f(X,A)$ a GCN with an undirected graph input ${G}=(\mathcal{V}, \mathcal{E})$, where $v_{i} \in \mathcal{V}$ is a set of $n$ nodes and  $\left(v_{i}, v_{i}\right) \in \mathcal{E}$ the set of edges. The input is a sparse graph representation, with $X$ being a node feature matrix and $A \in \mathbb{R}^{n \times n}$ being the adjacency matrix, defining the position of edges between nodes. In the initial case of no self-loops, the adjacency's diagonal is filled resulting in $\vec{A}=A+I_{n}$. The graph forward pass through the convolutional layer $l$ is then defined as

% equation
\begin{equation}
    H^{(l+1)}=\sigma\left(\sum_{i \in n} \frac{\vec{A_{:,i}}}{\|\vec{A_{:,i}}\|} H^{(l)} W^{(l)}\right).
\end{equation}

\footnote{For symmetric normalization we use $\tilde{D}^{-\frac{1}{2}} \tilde{A} \tilde{D}^{-\frac{1}{2}}$ with $\tilde{D}_{i i}=\sum_{j} \tilde{A}_{i j}$}

The adjacency is row-wise normalized for each node. $W^{(l)}$ is the layer-specific weight matrix and contains the learnable parameters. $H$ returns then the hidden representation of the input graph \cite{gangemi_modeling_2018}.
The GCN was first introduced as node classifier, predicting a probability distribution over all classes for each node in the input graph. Thus, the output dimensions are $Z \in \mathbb{R}^{n \times d_z}$ for the GCN prediction or latent representation matrix $Z$. Let $\hat{A}$ be the normalized adjacency matrix, then the full equation for a two layer GCN is 

\begin{equation}
    Z=f(X, A)=\operatorname{softmax}\left(\hat{A} \operatorname{ReLU}\left(\hat{A} X W^{(0)}\right) W^{(1)}\right).
\end{equation}

% \subsubsection{RGCN}

% GCN which takes further input of edge attribute matrix.

% Either present\\
% Dynamic Edge-Conditioned Filters in Convolutional Neural Networks on Graphs\\
% or nixx
% % Realational Graph Convolution Net (RGCN) was presented in \cite{kipf_semi-supervised_2017} for edge prediction. This model takes into account features of nodes. Both the adjacency and the feature matrix are matrix-multiplied with the weight matrix and then with them-selves. The resulting vector is a classification of the nodes.

\subsubsection{Graph VAE}
\label{ssec:GVAE}
% Encoder options: MLP RGCN
% Decoder MLP
% One shot: creating adjacency and feature matrix at once.

% the model we are  use for challenging KG datasets
We use the presented modules to compose the RGVAE. While the approaches from the literature for graph generative models differ in terms of the model and graph representation, we focus on the GraphVAE architecture presented by Simonovsky \cite{simonovsky_graphvae_2018}, A sparse graph model with graph convolutions.

% TODO rewrite and describe label condition as addition.

Simonovsky's GraphVAE follows the characterizing encoder decoder architecture.
The encoder $q_{\phi}(\mathbf{z} \mid {G})$ takes a graph ${G}$ as input, on which graph convolutions are applied. After the convolutions the hidden representation is flattened and concatenated with the node label vector $y$. A MLP encodes the mean $\mu(\mathbf{z})$ and logvariance $\sigma^2(\mathbf{z})$ of the latent space distribution. Using the reparametrization trick the latent representation is sampled.

For the decoder $p_{\theta}(\mathbf{x} \mid {G})$ the latent representation is again concatenated with the node labels. The decoder architecture for this model is a MLP with the same dimension as the encoder MLP but in inverted order, which outputs a flat prediction of $\tilde{G}$, which is split and reshaped in to the sparse matrix representation.

Simonovsky's GraphVAE \cite{simonovsky_graphvae_2018} is optimized for molecular data. Our aim is to set a proof of concept with the RGVAE for multi-relational KGs, thus, the structure of the GraphVAE is adopted but reduced to a minimum viable product by dropping the conditioning on the node labels and instead using the node attributes as pointers towards the corresponding entity in $\mathcal{E}$. By using node attributes as unique pointers, we exclude any class or type information about the entity. Simplifying further we drop the convolutional layer and directly flatten $G$ as input for the MLP encoder. To isolate the impact of graph convolutions as encoder for the RGVAE, we make the choice between MLP or GCN as encoder a hyperparameter.

\begin{figure}[H]
    \centering
    \includegraphics[scale=0.6,page=1]{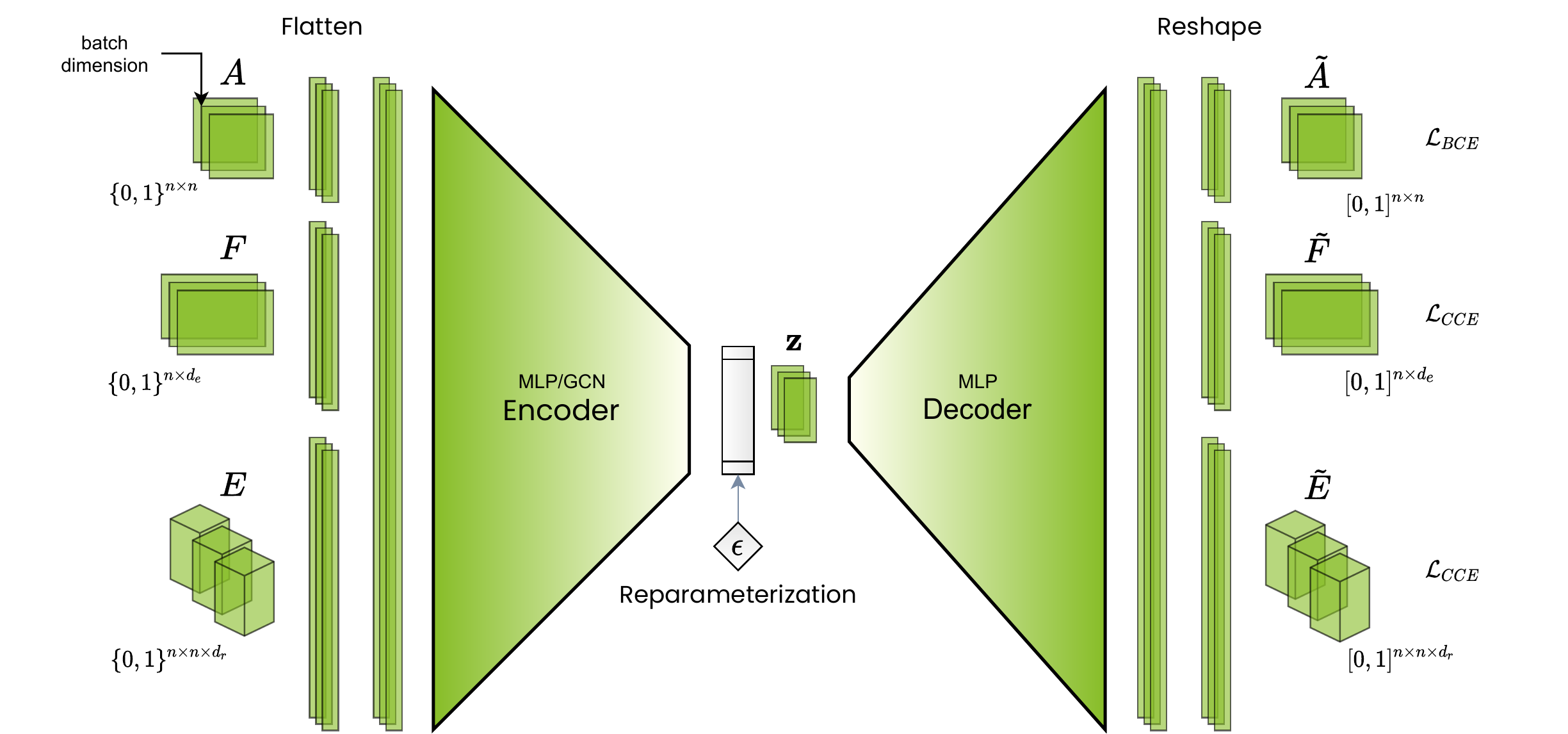}
    \caption{Architecture of the RGVAE.}
    \label{fig3:GVAE}
\end{figure}

% The encoder can either be a MLP, a GCNN or an RGCN. The same holds for the decoder with the addition that model architecture needs to be inverted. An version of a Graph VAE presented in \cite{simonovsky_graphvae_2018}. This model combines both the previous methods. The input graph undergoes relational graph convolutions before it is flattened and projected into latent space. After applying the reparametrization trick, a simple MLP decoder is used to regenerate the graph. In addition the model concatenates the input with a target vector $y$, which represents ???. The same vector is concatenated with the latent tensor. ***Elborate why they do that***.

% Reverse MLP

% Loss function

% What information can you capture when sparse compared to dense?

% Graphs can be generated recursively or in an one-shot approach. This paper uses the second approach and generates the full graph in one go. ***Cite?***
% This model be the starting point for our research.

% \subsubsection{One Shot vs. Recursive}
% One shot: MNIST vs recursive on graphs: Belli
In figure \ref{fig3:GVAE} the concept of the RGVAE is displayed. Each datapoint $G(A,E,F)$ is a subgraph from the KG dataset. Note that the model propagates batches instead of single datapoints. The RGVAE can generate graphs $\tilde{G}(\tilde{A},\tilde{E},\tilde{F})$ by sampling from the approximated posterior distribution $p_{\theta}\left( G \mid \mathbf{z}\right)$. Since it predicts on closed sets of relations and entities, the generated subgraphs are either unseen and complement the KG or are already present in the dataset. The subgraph are sparse with $n$ nodes. A single triple being $n=2$ and a subgraph representation $2<n<40$, where $n=40$ was the explored maximum for the GraphVAE \cite{simonovsky_graphvae_2018}.  

\subsection{Graph Matching}
\label{ssec:graphmatch}

% Intro to graph matching on sparse graphs.
In this subsection we explain the term permutation invariance and its impact on the RGVAE's loss function. Further we present a $k$-factor graph matching algorithm for general graphs with edge and node attributes and the Hungarian algorithm as solution for the NP-hard problem of linear sum assignment. Finally we derive the full loss of the RGVAE when applying the calculated permutation matrix to the models prediction.

\subsubsection{Permutation Invariance}

% Permutation Invariance
% The position or rotation of a graph can vary. 
% Use graph matching to detect similarities between graphs

A visual example of permutation invariance is the image generation of numbers. If the loss function of the model would not be permutation invariant, the generated image could show a perfect replica of the input number, but being translated by one pixel the loss function would penalize the model. Geometrical permutations can be translation, scale or rotation around any axis. 

In the context of sparse graphs the most common, and relevant permutation for this thesis, is the position of a link in the adjacency matrix. By altering its position through matrix-multiplication of the adjacency and the permutation matrix, the original link can change direction or turn into a self-loop. When matching graphs with more $n>2$, permutation can change the nodes which the link connects. Further it is possible to match graphs only on parts of the graph, $k$-factor, instead of the full graph. In the case of different node counts between target and prediction graph, the target graph can be fully (1-factor) matched with the larger prediction graph. 

In context of this thesis, a model or a function is called permutation invariant, if it can match any permutation of the original. This allows the model a wider spectrum of predictions instead of penalizing it on the element-wise correct prediction of the adjacency.

% OR: An example is in object detection in images. An object can have geometrical permutations such as translation, scale or rotation, none the less the model should be able to detect and classify it. In that case, the model is not limited by permutations and  is there fore permutation invariant.
% In our case the object is a graph and the nodes can take different positions in the adjacency matrix. To detect similarities between graphs we apply graph matching.

\subsubsection{Max-Pool Graph matching algorithm}
% These are three of the state of the art graph matching algorithms.

% \begin{itemize}
%     \item Wasserstein
%     \item Maxpooling
%     \item one more
% \end{itemize}

Graph matching of general (not bipartite) graphs is a nontrivial task. Inspired by Simonovsky's approach \cite{simonovsky_graphvae_2018}, the RGVAE uses the max-pool algorithm, which can be effectively integrated in its loss function. Presented in \cite{cho_finding_2014} in the context of computer vision and successful in matching feature point graphs in an image. It first calculates the affinity between two graphs, considering node and edge attributes, then applies edge-wise max-pooling to reduce the affinity matrix to a similarity matrix. Cho \textit{et al.} praise the max-pool graph matching algorithm as resilient to deformations and highly tolerant to outliers compared to the mean or sum alternatives. The output is a normalized similarity matrix in continuous space of the same shape as the target adjacency matrix, indicating the similarity between each node of target and prediction graph. The similarity matrix is subtracted from a unit matrix to receive the cost matrix, necessary for the final step in the graph matching pipeline. Notable is, that this algorithm also allows $k$-factor matching, with $k \leq 1 < n$. Thus, subgraphs with different number of nodes can be matched. The final permutation matrix is determined by linear sum assignment of the cost matrix, a np-hard problem \cite{diestel2016graph}.

% Max-pooling algorithm comes here !!!
We use the previously presented sparse representation for subgraphs, sampled from a KG. The discrete target graph is $G=(A, E, F)$ and the continuous prediction graph $\widetilde{G}=(\widetilde{A}, \widetilde{E}, \widetilde{F})$. The $A, E, F$ matrices store the discrete data for the adjacency, for node attributes and the node attribute matrix of form $A \in\{0,1\}^{n \times n}$ with $n$ being the number of nodes in the target graph. $E\in\{0,1\}^{n \times n \times d_e}$ is the edge attribute matrix and a node attribute tensor of the shape $F\in\{0,1\}^{n \times d_n}$ with $d_e$ and $d_n$ being the size of the entity and relation dictionary. For the predicted graph with $k$ nodes, the adjacency matrix is $\widetilde{A} \in[0,1]^{k \times k}$, the edge attribute matrix is $\widetilde{E} \in \mathbb{R}^{k \times k \times d_{e}}$ and the node attribute matrix is $\widetilde{F} \in \mathbb{R}^{k \times d_{n}}$.

Given these graphs the algorithm aims to find the affinity matrix $S:(i, j) \times(a, b) \rightarrow \mathbb{R}^{+}$ where $i, j \in G$ and $a, b \in \widetilde{G}$. The affinity matrix returns a score for all node and all edge pairs between the two graphs and is calculated 

\begin{equation}
    \begin{array}{l}
        S((i, j),(a, b)) = \left(E_{i, j, \cdot}^{T}, \widetilde{E}_{a, b, \cdot}\right) A_{i, j} \widetilde{A}_{a, b} \widetilde{A}_{a, a} \widetilde{A}_{b, b}[i \neq j \wedge a \neq b] + \left(F_{i, \cdot}^{T} \widetilde{F}_{a, \cdot}\right) \widetilde{A}_{a, a}[i=j \wedge a=b].
    \end{array}
\label{eq3:s}
\end{equation}

Here the square brackets define Iverson brackets \cite{simonovsky_graphvae_2018}.

While affinity scores resemblances which suggest a common origin, similarity directly refers to the closeness between two nodes.
The next step is to find the similarity matrix $X^* \in[0,1]^{k \times n}$. 
Therefore we iterate a first-order optimization framework and get the update rule

\begin{equation}
    X^*_{t+1} \leftarrow \frac{1}{\left\|\mathbf{S} X^*_{t}\right\|_{2}} \mathbf{S} X^*_{t}.
\end{equation}

To calculate $SX^*$ we find the best candidate $X^*_{i,a}$ from the possible pairs of $i \in \mathbb{N}^{[0,n]}$ and $ai \in \mathbb{N}^{[0,k]}$ in the affinity matrix $S$. Heuristically, taking the argmax over all neighboring node pair affinities yields the best result. Other options are sum-pooling or average-pooling, which do not discard potentially irrelevant information, yet have shown to perform worse. Thus, using the max-pooling approach, we can pairwise calculate

\begin{equation}
    \mathbf{Sx}_{i a}=X^*_{i a} \mathbf{S}_{i a ; i a}+\sum^{n}_{j = 0} \max _{0 \leq b < k \mathbb{N}_{a}} X^*_{j b} \mathbf{S}_{i a ; j b}.
\end{equation}

Depending on the matrix size, the number of iterations are adjusted. The resulting similarity matrix $X*$ yields a normalized similarity score for every node pair. The next step if to converting it to a discrete permutation matrix.

\subsubsection{Hungarian algorithm}
\label{ssec3:hung}

%  Find shortest path
Starting with the normalized similarity matrix $X^*$, we reformulate the aim of finding the discrete permutation matrix as a linear assignment problem. Simonovsky \textit{et al.} \cite{silver_mastering_2017}use for this purpose an optimization algorithm, the so called Hungarian algorithm. It original objective is to optimally assign $n$ resources to $n$ tasks, thus $k-n$ rows of the permutation matrix are left empty. The cost of assigning task $i \in \mathbb{N}^{[0,n]}$ to $a \in \mathbb{N}^{[0,k]}$ is stored in $x_{ia}$ of the cost matrix $C \in \mathbb{N}^{n \times k}$. By assuming tasks and resources are nodes and taking $C=1-X^*$ we get the continuous cost matrix $C$. This algorithm has a complexity of $O\left(n^{4}\right)$, thus is not applicable to complete KGs but only to subgraphs with limited number of nodes per graph \cite{date_gpu-accelerated_2016}.

The core of the Hungarian algorithm consist of four main steps, initial reduction, optimality check, augmented search and update. The presented algorithm is a popular variant of the original algorithm and improves the complexity of the update step from $O\left(n^{2}\right)$ to $O\left(n\right)$ and thus, reduces the total complexity to $O\left(n^{3}\right)$. Since throughout this thesis $n=k$ we imply the reduction step and continue with a quadratic cost matrix $C$. The following notation is solely to derive the Hungarian algorithm and does not apply to our graph data. The algorithm takes as input a bipartite graph $G=(V, U, E)$ and the cost matrix $C \in \mathbb{R}^{n \times n}$. $G$ bipartite because it considers all possible edges in the cost matrix in one direction and no self-loops. $V \in \mathbb{R}^n$ and $U \in \mathbb{R}^n$ are the resulting sets of nodes and $E \in \mathbb{R}^{n}$ the set of edges between the nodes. The algorithm's output is a discrete matching matrix $M$. To avoid two irrelevant pages of pseudocode, the steps of the algorithm are presented in the following short summary \cite{mills-tettey_dynamic_nodate}.

\begin{enumerate}
    \item Initialization: \\
    \begin{enumerate}
        \item Initialize the empty matching matrix $M_{0}=\emptyset$.
        \item Assign $\alpha_i$ and $\beta_i$ as follows:
        \begin{align*}
            \forall v_{i} &\in V, \quad &&\alpha_{i}=0 \\
            \forall u_{i} &\in U, \quad &&\beta_{j}=\min _{i}\left(c_{i j}\right)
        \end{align*}
        \end{enumerate}
    \item Loop $n$ times over the different stages:
    \begin{enumerate}
        \item Each unmatched node in $V$ is a root node for a Hungarian tree with completed results in an augmentation path.
        \item Expand the Hungarian trees in the equality subgraph. Store the indices $i$ of $v_i$ encountered in the Hungarian tree in the set $I*$ and similar for $j$ in $u_j$ and the set $J^*$. If an augmentation path is found, skip the next step.
        \item Update $\alpha$ and $\beta$ to add new edges to the equality subgraph and redo the previous step.
        \begin{align*}
            \theta&=\frac{1}{2} \min _{i \in I^{*}, j \notin J^{*}}\left(c_{i j}-\alpha_{i}-\beta_{j}\right) \\
            \alpha_{i} &\leftarrow\begin{cases}{ll}
            \alpha_{i}+\theta & i \in I^{*} \\
            \alpha_{i}-\theta & i \notin I^{*}
            \end{cases} \\
            \beta_{j} &\leftarrow\begin{cases}{ll}
            \beta_{j}-\theta & j \in J^{*} \\
            \beta_{j}+\theta & j \notin J^{*}
            \end{cases} \\
        \end{align*}
        \item Augment $M_{k-1}$ by flipping the unmatched with the matched edges on the selected augmentation path. Thus $M_k$ is given by $\left(M_{k-1}-P\right) \cup\left(P-M_{k-1}\right)$ and $P$ is the set of edges of the current augmentation path.
    \end{enumerate}
    \item Output $M_n$ of the last and $n^{th}$ stage.
\end{enumerate}

\subsubsection{Graph Matching VAE Loss}
\label{ssec3:GVAEloss}
Coming back to our generative model, we now explain how the loss function needs to be adjusted to work with graphs and graph matching, which results in a permutation invariant graph VAE.

The normal VAE maximizes the evidence lower-bound or, in a practical implementation, minimizes the upper-bound on negative log-likelihood. Using the notation of section \ref{ssection:VAE} the graph VAE loss is

\begin{equation}
    \begin{array}{l}
    \mathcal{L}(\phi, \theta ; G)=\mathbb{E}_{q_{\phi}(\mathbf{z} \mid G)}\left[-\log p_{\theta}(G \mid \mathbf{z})\right]+\beta (\operatorname{KL}\left[q_{\phi}(\mathbf{z} \mid G) \| p(\mathbf{z})\right]).
    \end{array}
\end{equation}

The loss function $\mathcal{L}$ is a combination of reconstruction term and regularization term. The regularization term is the KL divergence between a standard normal distribution and the latent space distribution of $\mathbf{z}$. The change to graph data does not influence this term  to graphs. The reconstruction term is the cross entropy between prediction and target, binary for the adjacency matrix and categorical for the edge and node attribute matrices. 

% A sigmoid with logits, E,F, softmax include translation X

% r. Sigmoid activation function is used to compute
% Ae, whereas edge- and node-wise softmax is applied to obtain
% Ee and Fe, respectively. A

The predicted output of the decoder is split in three parts and while $\tilde{A}$ is activated through sigmoid, $\tilde{E}$ and $\tilde{F}$ are activated via edge- and nodewise Softmax. 
For the case of $n<k$, the target adjacency is permuted $A^{\prime}=X A X^{T}$, so that the model can backpropagate over the full prediction. Since $E$ and $F$ are categorical, permuting prediction or target yields the same cross-entropy. Following Simonovsky's approach \cite{simonovsky_graphvae_2018} we permute the prediction, $\widetilde{F}^{\prime}=X^{T} \widetilde{F}$ and $\widetilde{E}_{\cdot, \cdot, l}^{\prime}=X^{T} \widetilde{E}_{\cdot, \cdot, l} X$. Let $l$ be the one-hot encoded edge attribute vector which is permuted. These permuted subgraphs are then used to calculate the maximum log-likelihood estimate \cite{simonovsky_graphvae_2018}:

\begin{equation}
    \begin{split}
        \log p\left(A^{\prime} \mid \mathbf{z}\right) = &1 / k \sum_{a} A_{a, a}^{\prime} \log \widetilde{A}_{a, a}+\left(1-A_{a, a}^{\prime}\right) \log \left(1-\widetilde{A}_{a, a}\right)+ \\ & +1 / k(k-1) \sum_{a \neq b} A_{a, b}^{\prime} \log \widetilde{A}_{a, b}+\left(1-A_{a, b}^{\prime}\right) \log \left(1-\widetilde{A}_{a, b}\right)
    \end{split}
    \label{eq3:GAVElossA}
\end{equation}

\begin{align}
    \log p(F \mid \mathbf{z}) &=1 / n \sum_{i} \log F_{i, \cdot}^{T} \widetilde{F}_{i,}^{\prime} \\
    \log p(E \mid \mathbf{z}) &=1 /\left(\|A\|_{1}-n\right) \sum_{i \neq j} \log E_{i, j,}^{T}, \widetilde{E}_{i, j, \cdot}^{\prime}
    \label{eq3:GAVElossEF}
\end{align}

The normalizing constant $1 / k(k-1)$ takes into account the no self-loops restriction, thus an edge-less diagonal. In the case of self loops this constant is $1 / k^2$. Similar $1 /\left(\|A\|_{1}-n\right)$ for $\log p(E \mid \mathbf{z})$ where $-n$ accounts for the edge-less diagonal and in case of self-loops is discarded, resulting in $1 /\left(\|A\|_{1}\right)$.

\subsection{Ranger Optimizer}
\label{sec3:ranger}

Finalizing this chapter, we explain the novel deep learning optimizer Ranger. Ranger combines Rectified Adam (RAdam), lookahead and optionally gradient centralization. Let us briefly look into the different components.
RAdam is based on the popular Adam optimizer. It improves the learning by dynamically rectifying Adam's adaptive momentum. This is done by reducing the variance of the momentum, which is especially large at the beginning of the training. Thus, leading to a more stable and accelerated start \cite{liu_variance_2020}.
The Lookahead optimizer was inspired by recent advances in the understanding of loss surfaces of deep neural networks, thus proposes an approach where, a second optimizer estimates the gradients behavior for the next steps on a set of parallel trained weights, while the number of 'looks ahead' steps is a hyperparameter.
This improves learning and reduces the variance of the main optimizer \cite{zhang_lookahead_2019}.
The last and most novel optimization technique, Gradient Centralization, acts directly on the gradient by normalizing it to a zero mean. Especially on convolutional neural networks, this helps regularizing the gradient and boosts learning. This method can be added to existing optimizers and can be seen as constrain of the loss function \cite{yong_gradient_2020}.
Concluding we can say that Ranger is a state of the art deep learning optimizer with accelerating and stabilizing properties, incorporating three different optimization methods, which synergize with each other. Considering that generative models are especially unstable during training, we see Ranger as a good fit for this research.

% LookAhead was inspired by recent advances in the understanding of loss surfaces of deep neural networks, and provides a breakthrough in robust and stable exploration during the entirety of training.

\section{Methods}
This section describes the methods used for the implementation of the RGVAE and in the experiments of this thesis. We begin explaining the formatting and preprocessing of the data, and introduce the VanillaRGVAE, focusing on the encodervariations and on the decoder as generator. Furthermore our implementation of the batch-wise max-pooling graph matching algorithm is presented as well as combined with the RGVAE's loss function. Finally, we describe the link prediction pipeline which is the first experiment which the RGVAE is evaluated on. Our model implementation and experiments are written in Python using PyTorch, a high-performance deep-learning library \cite{paszke_pytorch_2019}. All experiments are performed in a fully reproducible manner and the modular implementation of the RGVAE is meant to be reused and further developed in future work. The code is openly available on Github: \footnote{\url{https://github.com/INDElab/rgvae}}

\subsection{Knowledge graph data}
All our presented methods operate on KG data. While data from other graph domains is possible, this work focuses solely on datasets in triple format. We explain the sparse graph representation, which is the input format for our model and how to preprocess the original KG triples to match that format.

\subsubsection{Graph Representation}

% The first step in our pipeline is the representation of the KG in tensor format. In order to represent the graph structure we use an adjacency matrix $A$ of shape $n\times n$ with $n$ being the number of nodes in our graph. The edge attribute or directed relations between the nodes are represented in the matrix $E$ of shape $n\times n\times d_E$ with $d_E$ being the number of edge attributes. Similarly for node attributes we have the matrix $F$ of shape $n\times d_N$ with $d_N$ number of node attributes. The input graph can have less nodes than the maximum $n$ but not more. The diagonal of the adjacency matrix is filled with $1$ if the indexed node exists, and with $0$ otherwise. The number and encoding of the attributes must be predefined and cannot be changed after training. This way we can uniquely represent a KG.

This work uses the graph representation $G(A,E,F)$, where $A$ denotes the adjacency matrix, $E$ the edge feature matrix and $F$ the node feature matrix. This input format for the model architectures as presented in section \ref{ssec:GVAE}. The graph is binary and each matrix batch is stored as separate tensor.

% adjacency matrix
The adjacency matrix $A$ takes the shape $(n\times n)$ with $n$ being the number of nodes in our graph/subgraph. While most of previous work would only allow edges on the upper triangular adjacency matrix and fill the diagonal with ones, we chose a less constrained representation, which we assume is a better fit for representing KGs. In particular, we allow self-loops, meaning a triple where object and subject are the same entity and our relations are directed and can be inverted. Thus $A$ can have a positive signal at any position $A_{i,j}$  $i,j \in \mathbb{R}^{n \times n}$, indicating a directed edge between node of index $i$ and node of index $j$, while $A_{i,j}$ differs from $A_{j,i}$.

% edge attribute matrix
The edge attribute matrix $E$ takes the shape $(n\times n\times d_e)$ with $d_e$ being the number of unique entities in our dataset. For each possible edge in the adjacency matrix we have a one hot encoded vector pointing to the unique relation in the dataset. Stacking these vectors leads to the three dimensional matrix $E$.

% node attribute matrix
The shape of node attributes matrix $F$ is $(n\times d_e)$ with $d_e$ being the number of node attributes describing the nodes. Considering that we split the KG in subgraphs, we use the entity index as node attribute, making it possible to assign every node in a subgraph to the entity in the full KG. Thus, the number of node attributes $d_e$ equals the number unique entities in our dataset. Again the node attributes are one hot encoded vectors, which concatenated result in the two dimensional $F$ matrix.

\subsubsection{Preprocessing}
Our datasets consist of three tab separated value files full of triples for training, validation and testing. The preprocessing steps convert the triples to subgraphs $G(A,E,F)$ and during postprocessing back into triple format as well as a human readable translation. Best practices of research are followed by withholding the test set until the final run.

% set of entities, set of relations 
% triple to graph
% all triples
% true triples dict
% indexing
From all three sets, we create a set of all occurring entities and similar set for the relations. Now we can define our dimensions $d_e$ and $d_r$. For both sets we create two dictionaries \textit{index-2-entity} and \textit{entity-2-index}  which map back and forth between numerical index and the string representation of the entity (similar for the relation set). These dictionaries are used to create a train and test set of triples with numeric indices. Depending if we are in the final testing stage or not, we include all triples from the training and evaluation file in the training set and use the triples in the testing file as test set, or we ignore the triples in the test file and use the evaluation file triples as test set.

%  truetriples dict 
Further we create two dictionaries, \textit{head} and \textit{tail} which for all occurring subject and relation combination, contain all entities which would complete it to a real triple in our dataset (similar for all relation and object combinations). This allows us to filter true triples, which reduces the score bias for link prediction and evaluates the ratio of unseen triples for graph generation. 

%  triple 2 graph
The final step of preprocessing is a function, which takes a batch of numerical triples and converts them to a batch of binary, multidimensional tensors $A$, $E$ and $F$. While this might sound easy for only one triple per graph, it proves more complex for graphs with $n>2$ facing exemption cases such as self loops or an entity occurring in two triples. We solve this by creating a separate set for head and tail entities, then storing the indices of both in a list, starting with the subject set and finally using this list as keys for a dictionary with values in the range to $n$. In both edge cases, this results in an adjacency matrix with a rank lower than $n$. A similar approach, with fewer edge cases to consider, is used to convert the tensor matrices back to triples.

% Graph embeddings? unsupervised approach

\subsection{RGVAE}
The principle of a graph VAE has been explained in section \ref{ssec:GVAE}, which also covers the foundation of our model, the RGVAE. Therefore we focus on the implementation as well as parameter and hyperparameter choice. Since this work is meant to be a proof of concept rather than aimed at outperforming the state of the art, our model is kept as simple as possible and only as complex as necessary. Our approach is modular for both experiment pipeline and model, meaning independence between sequential modules and compatibility with parallel modules. For the encoder we implemented two variations, a fully connected and a convolutional, while for the decoder we opted for a single fully connected network.
%  just explain the code

\subsubsection{Initialization}

The RGVAE is initialized with a set of hyperparameters, which define the input shape. Table \ref{tab:RGVAEhyp} shows a complete list of those parameters and their default values. It is left to mention that we use the Xavier uniform initialization with a gain of $0.01$ to initialize the parameters \cite{glorot2010understanding}.

\begin{table}[H]
\centering
    \begin{tabular}{|l|l|l|}
    \hline
    \rowcolor[HTML]{EFEFEF}
    \multicolumn{1}{|c}{\textsc{Hyerp.}} & \multicolumn{1}{c}{\textsc{Default}} & \multicolumn{1}{c|}{\textsc{Description}} \\\hline
    $n$     & \multicolumn{1}{c|}{$2$} & Number of nodes  \\
    $d_e$   &\multicolumn{1}{c|}{-}   & Total number of entities\\
    $d_r$   &\multicolumn{1}{c|}{-} & Total number of relations\\
    $d_z$ &\multicolumn{1}{c|}{$100$}   & Latent space dimension\\
    $d_h$ &\multicolumn{1}{c|}{$512$}   & Hidden dimension\\
    $dropout$ &\multicolumn{1}{c|}{$0.2$}   & Dropout\\
    $\beta$ & \multicolumn{1}{c|}{$1$}  & $\beta$ value for regularization  \\
    $perminv$ & \multicolumn{1}{c|}{\textbf{True}}  & Permutation invariant loss function  \\
    $clipgrad$ & \multicolumn{1}{c|}{\textbf{True}}  & Learning w/ gradient clipping  \\
    $encoder$ & \multicolumn{1}{c|}{\textbf{MLP}}  & Choice of encoder architecture  \\
    \hline
    \end{tabular}
    \caption{The initial hyperparameters of the RGVAE with default value and description.}
    \label{tab:RGVAEhyp}
\end{table}

\subsubsection{Encoder}
% \\Convolution part
% \\RCGN relation Convolution neural net
% \\MLP encoder
% \\Latent space
% \\reparametrization trick
% \\MLP decoder
% \\Graph matching
% \\Discretization of prediction

% MLP
The proof-of-concept encoder is a MLP as described in section \ref{ssec:mlp}, which takes the flattened concatenated threefold graph $\mathbf{x}=G(A,E,F)$ as batch input. We use the initial parameters to calculate the input

\begin{equation}
    d_{input} = n^2 + n^2*d_r + n*d_e
    \label{eq4:inputdim}
\end{equation}

The main encoder architecture is a 3 layer fully connected network, with both layers using ReLU as activation function. The choice for two hidden layers is based on the huge difference between $d_{input}$ and $d_z$. The first layer has a dimension of $2*d_h$ and the option to use dropout, which by default is set to $0.2$. The second (hidden) layer has the dimension $d_h$ which is by default set to $1024$. After the second ReLU activation, the encoder linearly transforms the hidden state to an output vector of $2 *d_z$. This vector is split and makes the mean and log-variance of size $d_z$ for the reparametrization trick. Sampling $\epsilon$ from an autonomous module, we get the latent representation $\mathbf{z}$.

% % Convolutions
The second option for our RGVAE encoder is a GCN as described in section \ref{ssec:gcn}. We adopt the architecture from \cite{kipf_semi-supervised_2017} namely two layers of graph convolutions with dropout in between. To match the encoder output to the base model, we then add a flattening and a final linear transformation layer. To substitute the feature matrix used in Kipf's work, we reduce the edge attribute matrix $E$ by one dimension and concatenate it with $F$ resulting in $X_{GCN} \in \mathbb{R}^{n \times (d_e+nd_r)}$. The forward pass of the adjacency matrix $A$ and $X_{GCN}$ through the first GCN layer with a hidden dimension of $d_h$ and ReLU as activation function is followed by a dropout layer. It should be mentioned that dropout is only applied during learning, not on evaluation. The second GCN layer takes the output from the previous layer, the two dimensional hidden state and again $A$ as input. Now, instead of having the GCN predict on a number of classes, we use it to output a logits vector of dimension $2d_z$. Therefore we pass the GCN output through a flattening and a linear transformation layer. Similar to the above described encoder we use the reparametrization trick to output the latent reparametrization $\mathbf{z}$. Table \ref{tab4:archcompare} shows the two encoder architectures side by side.

\begin{table}[H]
    \centering
        \begin{tabular}{|l|l|}
        \hline
        \rowcolor[HTML]{EFEFEF}
        \multicolumn{1}{|c}{\textsc{MLP}} & \multicolumn{1}{c}{\textsc{Graph Conv.}} \\\hline
        Flatten($d_{input}$) &   Concatenate \\
        Linear($d_{input},2*d_{h}$) &   Convolution($A,X$) \\
        ReLu() &   ReLU() \\   
        Dropout($0.2$) &   Dropout($0.2$) \\
        Linear($2*d_{h},d_{h}$) &   Convolution($H^{(1)},X$) \\
        ReLu() &   Flatten \\
        Linear($d_{h},2*d_{z}$) &   Linear($d_{H^{(1)}},2*d_z$) \\
        \hline
        \end{tabular}
        \caption{Comparison of the two variants for the encoder of the RGVAE.}
        \label{tab4:archcompare}
    \end{table}

\subsubsection{Decoder}

The RGVAE decoder is a MLP with architecture and dimensions similar but in inverse order to the encoder MLP described in table \ref{tab4:archcompare}. Since we are decoding the latent space, the input dimension is $d_z$ and the output dimension is $d_{input}$ as calculated in equation \ref{eq4:inputdim}. The flat logits output tensor is split threefold and reshaped to the original input shape of $G(A,E,F)$.

To sample from the generated graph we apply the Sigmoid activation function to the logits prediction for the adjacency matrix and use the normalized output as weights for binomial distributions, from which we can sample the discrete $\tilde{A}$. For $\tilde{E}$ and $\tilde{F}$ we take the argmax on the last dimension of both matrices. Each node and edge can have only one attribute, referring to its index in $\mathcal{E}$ and $\mathcal{V}$, thus only the highest predicted value is relevant. The generated sample is a discrete graph $\tilde{G}(\tilde{A},\tilde{E},\tilde{F})$.

\subsubsection{Limitations}

The main limitation of the RGVAE is the quadratic increase of model parameters with the increase of nodes per input graph $\mathcal{O}(n^2)$. The number of parameters to train is directly linked with the GPU memory requirements. Even more computationally expensive is the use of permutation invariant graph matching, with a complexity of $\mathcal{O}(n^3)$. This sets an empirical limitation for the model to small graphs with $2<n<40$ \cite{simonovsky_graphvae_2018}.

% The proposed model is expected to be useful
% only for generating small graphs. This is due to growth
% of GPU memory requirements and number of parameters
% (O(k
% 2
% )) as well as matching complexity (O(k
% 4
% )), with small
% decrease in quality for high values of k. I
% \cite{simonovsky_graphvae_2018}

% \subsubsection{Hyperparameters}
% learning rate
% beta for regularization term
% hidden dimensions
% latent dimensions
% dropout
% n, d_e, d_r
% Convolutions vs no convolutions
% IDEA: Convolutions + MLP

\subsection{RGVAE learning}
% Lay out pipeline.

In this section we present our implementation of how to fit the model to the data. Learning a model on data is a mostly standardized procedure, which includes training and evaluation per epoch. During training, the model forward passes the data, computes the loss, then does a backwards pass and updates its parameters. During evaluation, it is presented a split of the dataset unseen during training. Only the forward pass is done and the loss tracked during evaluation. Up to this point the RGVAE does not differ from the vanilla VAE training. Special is the graph matching function which is applied to the predicted graph and the loss function which takes into account the permutation. Thus, we look deeper into graph matching and derive RGVAE loss.
The training and all experiments are performed on the GPU cluster LISA of the supercomputer SURFsara on the Dutch national e-infrastructure with the support of SURF Cooperative \cite{fengvaleriu}. Each GPU node is powered by a Nvidia titan RX 25GB. We log our experiments and results using \textit{Weights \& Biases}, a cloud-based experiment tracking tool \cite{wandb}.

% GPU requirements.
% LISA
% Nvidia titan RX 25GB 60h
% Experiment log wndb.ai [cite]
% optimizer ranger github repo link

\subsubsection{Max pooling graph matching}

While the pseudocode presented in \cite{cho_finding_2014} is simple and straightforward, it proves complicated to implement this in algorithms for batches and thus, without looping over the indices. Yet, our batch implementation solves these challenges and is more efficient than the direct implementation, which we use for validating our results. Given the target graph $G$ and the predicted graph $\tilde{G}$, the algorithm can be divided in three steps, calculating the five dimensional affinity matrix (the first being the batch dimension), max-pooling the continuous similarity matrix $X^*$ and discretizing $X^*$ to our final permutation matrix $X$.

% Affinity
We use equation \ref{eq3:s} for the first step but instead of adding the two terms to a single output, we return $S$ twofold as $S_r$, five dimensional holding the information of edge affinity and $S_e$, three dimensional with the affinity information of the nodes. In a preprocessing step we zero out the diagonal of $A$, $\tilde{A}$ and for $E$ and $\tilde{E}$ the diagonal of the second and third dimension, to compile with the constrain $[i \neq j \wedge a \neq b]$ of the first term. For the second term we only take into account the diagonal of $\tilde{A}$ to compile with the constrain $[i=j \wedge a=b]$. Pseudocode \ref{alg4:s} shows the implementation, here \textit{diag()} stands for a vector with only the diagonal entries. For the dot product of $E$ and $\tilde{E}$ over the last dimension we implement our own version of \texttt{torch.matmul()} to cope with higher dimensions. The operator $\odot$ denotes element-wise matrix multiplication.

\begin{algorithm}[H]
    \caption{Batch implementation for the affinity between two graphs }
    \hspace*{\algorithmicindent} \textbf{Input:} $G(A,E,F$ and $\tilde{G}(\tilde{A},\tilde{E},\tilde{F})$
    \begin{algorithmic}[1]
        % \Input{$G(A,E,F$ and $\tilde{G}(\tilde{A},\tilde{E},\tilde{F})$}
        \Statex \textbf{First term:} $[i \neq j \wedge a \neq b]$
        \State $E_{term1} = E^T  \tilde{E}$ \Comment{Dot product over the last dimension}
        \State $A_{term1} = A\operatorname{.unsqueeze}(-1)^T (\tilde{A} \odot (\tilde{A}  \tilde{A}^T))\operatorname{.unsqueeze}(-1)$ \Comment{Dot product over the last (empty) dimension}
        \State $S_r = E_{term1} \odot A_{term1}$
        \Statex \textbf{Second term:} $[i=j \wedge a=b]$
        \State $A_{term2} = \operatorname{ones\_like}(\operatorname{diag}(\tilde{A}))^T \operatorname{diag}(\tilde{A})$
        \State $F_{term2} = F^T  \tilde{F}$ \Comment{Dot product over the last dimension}
        \State $S_e = F_{term2} \odot A_{term2}$
        \State \textbf{return} $(S_r, S_e)$
    \end{algorithmic}
    \label{alg4:s}
\end{algorithm}

% Max-pool loop
The next step is the graph matching algorithm is the max-pool loop presented in \cite{cho_finding_2014}. We initialize the similarity matrix as ones $X^* \in 1^{bs\times n \times n}$ with $bs$ denoting the batch size. For a certain number of iterations, Cho \textit{et al.} proposes $40$ but the number should be adjusted to the number of nodes in the graph, we multiply $X^*$ with a reduced version of $S$ and use its Frobenius norm as normalizer. The algorithm \ref{alg4:maxpool} shows our implementation for batches.

\begin{algorithm}
    \caption{Max-pool graph matching for batches}
    \hspace*{\algorithmicindent} \textbf{Input:} $(S_r, S_e)$
    \begin{algorithmic}[1]
        \State Init $X^* \in 1^{bs\times n \times n}$
        \For{$iteration=1,2,\dots$}
            \State $S_{max} = \operatorname{sum}(\operatorname{max}(S_r \odot X^*\operatorname{.unsqueeze}([1,1])))$ \Comment{Sum and max over the last dimension. Unsqueeze two times on the second dimension}
            \State $X^* = X^* \odot S_e + S_{max}$
            \State $X^* = X^* / \operatorname{frobenius\_norm}(X^*$)
        \EndFor
        \State \textbf{return} $X^*$
    \end{algorithmic}
    \label{alg4:maxpool}
\end{algorithm}

To the best of our knowledge, this is the first time this algorithm is implemented in batch style. Thus, we would like to believe that laying out the implementation in detail contribute to the academic value of this thesis.  

% Hungarian
The last step in the graph matching pipeline is the discretization of $X^*$. We adopt Simonovsky's \cite{simonovsky_graphvae_2018} approach which for this purpose uses the Hungarian algorithm as presented in section \ref{ssec3:hung}. To our disappointment and resulting in a bottleneck, no batch nor tensor implementation of the named algorithm has been published so far. Thus, the discrete permutation matrix $X$ is obtained by iterating over the batch dimension \footnote{We convert $X^*$ to \texttt{numpy.array()}, create the cost matrix $X_{cost} = 1 - X^*$ and use the \textit{Scipy} package \cite{2020SciPy-NMeth} \texttt{scipy.optimize.linear\_sum\_assignment} to iteratively calculate the permutation matrix $X$.}. If no permutation is needed, $X$ is the identity matrix of $A$. 

% %  explain the code
% % batch implementation
% % loop implementation to check
% Summing over the neighbors means summing over the whole column
% Normalize matrix with Frobenius Norm

% Batch version:
% Only matmul and dot. keep dimension of S with shape (bs,n,n,k,k)
% When maxpooling, flatten Xs (n,k) for batch dot multiplication. This way (i think) we sum over all j nad b neighbors instead of taking the max.  

\subsubsection{Loss function}
\label{ssec4:loss}

The RGVAE uses the ELBO loss from equation \ref{eq3:elbo}, consisting of two terms, the regularization loss and the reconstruction loss. We present our implementation of both loss terms with graph matching and an alternative loss without graph matching for comparative evaluation of our results 

%  reconstruction ref to loss
%  point out challenges, log, nan
We implement $\log p\left(A^{\prime} \mid \mathbf{z}\right)$ from equation \ref{eq3:GAVElossA} with the second normalizing constant as $1 / k*K$ since we allow self-loops. The permutation matrix $X$ is applied to to the target adjacency, resulting in $A^{\prime}$. For $\log p\left(E^{\prime} \mid \mathbf{z}\right)$ and $\log p\left(F \mid \mathbf{z}\right)$ the permutation is applied to the prediction, which in the case of $E^\prime$ requires our own implementation of matrix multiplication of $d>2$ . Taking into account self-loops we change the normalization constant of $\log p\left(E^{\prime} \mid \mathbf{z}\right)$ to $1 /\left(\|A\|_{1}\right)$. It is left to mention that when implementing $\sum_{i \neq j} \log E_{i, j,}^{T}, \widetilde{E}_{i, j, \cdot}^{\prime}$ in matrix multiplication style, we have to account for the zero values before taking the logarithm. We implement \texttt{torch.sum(torch.sum(torch.log(no\_zero($E$ * $\hat{E}$)),-1)-1)} with \texttt{no\_zero()} being a function which replaces $0$ values with $1$. This implementation of the loss function can be backpropagated with exception of the graph matching part, where the \texttt{numpy} implementation of the Hungarian algorithm prevents backpropagation.

% regularization loss + beta
The regularization loss is given by the KL divergence between the approximated posterior $\log q_{\phi}\left(\mathbf{z} \mid \mathbf{x}\right)$ and the Gaussian prior $p(\mathbf{z})$. The only modification we make to the original loss, is adding a $\beta$ parameter which in values $100<\beta <500$ has shown great results in factorizing the latent space \cite{higgins_beta-vae_2016}. By setting $\beta=1$ we return to the original loss function. This hyperparameter is explored in the experiments.

% present loss function w/o graph matching 
Alternatively and as ground truth we implement the VAE loss from equation \ref{eq3:elbo} for graphs without graph matching. we use binary cross entropy ($BCE$) as reconstruction loss of the adjacency,  and categorical cross entropy ($CE$) for the attribute matrices. The regularization loss is similar to the above presented and also includes the hyperparameter $\beta$. The equation for the ELBO with $\sigma()$ indicating Sigmoid activation is

\begin{equation}
    \mathcal{L}(\phi,\theta:G) = BCE(A,\sigma(\tilde{A})) + CC(E,\tilde{E}) + CC(F,\tilde{F}) - D_{K L}\left(q_{{\phi}}\left(\mathbf{z} \mid G\right) \| p_{{\theta}}(\mathbf{z})\right)
    \label{eg4:normalELBO}
\end{equation}

We train the model on the negative ELBO. Further we use \textit{Ranger} presented in section \ref{sec3:ranger} as optimizer combining three learning optimization methods. Out of the various optimization parameters, we achieved good performance with the default values and  only adjusted the learning rate and the number of lookahead steps. Missing a publication to cite, we refer to the source code of \textit{Ranger}: \footnote{\url{https://github.com/lessw2020/Ranger-Deep-Learning-Optimizer}}

% optimizer ranger github repo link

\subsection{Link prediction and Metrics}
\label{ssec4:lpmetrics}
% LP as main experiment and proof of concept. 
Our first experiment is link prediction, which we gives an insight in the models capacity to reproduce the data. It is intended as proof of concept rather than an attempt to set the state of the art. The results let us draw conclusions on the impact and function of different parameters.
Besides the final link prediction experiment, we perform link prediction on a randomly drawn small subset of the test set during training. This gives us a broader view on the models performance, which otherwise would only be evaluated by the ELBO loss.  

% describe LP as in paper OLD DOG
Link prediction on multi-relational KGs is the task of completing unobserved triples, based on the information gained from the trainings data. To evaluate a model on this task, the most common method is entity ranking in the form of triple completion of unseen triples from the test set. Given a KG $G(\mathcal{E},\mathcal{V})$ we want our model find the right entity out of $\mathcal{E}$ which completes the unseen triple $(s,r,?)$ or $(?,r,o)$ for heads or tail prediction. Thus, the model scores the triple for all possible combination with the entities from $\mathcal{E}$. The rank of the true triple, in descending order, defines the performance of the model \cite{ruffinelli_you_2019}.

In the preprocessing step we create a dictionary with all occurring combinations for all possible triples with missing head or tail. These dictionaries we use to filter out real triples from the scoring. Unfiltered scores are referred to as raw scores. Per link prediction run, the model has to score the number of triples in the test set times the $d_e$ the number of entities in $\mathcal{E}$ times two for head and tail, which mostly results in a number much larger than the size of the actual dataset.

Finally, the metrics for link prediction are the mean reciprocal rank (MRR) of the score for the true triple and the average HITS@$k$ with $k \in [1,3,10]$. We denote $\mathcal{K}_{test}$ the unseen test set and $\left|\mathcal{K}_{\text {test }}\right|$ the number of triples in the test set. The operator $\operatorname{rank()}$ returns in descending order the by the model scored position of the true triple. Head prediction of a triple given relation and object is denoted $(s|r,o)$ and likewise $(o|s,r)$ for tail prediction. The Iverson brackets $[\operatorname{rank}(s \mid r, o) \leq k]$ return $1$ if the scored rank is equal or lower than $k$, else $o$. 
% describe MRR, Hits at n

\begin{equation}
    \begin{aligned}
    \operatorname{MRR} &=\frac{1}{2\left|\mathcal{K}_{\text {test }}\right|} \sum_{(s, r, o) \in \mathcal{K}_{\text {lest }}}\left(\frac{1}{\operatorname{rank}(s \mid r, o)}+\frac{1}{\operatorname{rank}(o \mid s, r)}\right) \\
    \text { Hits@ } k &=\frac{1}{2\left|\mathcal{K}_{\text {test }}\right|} \sum_{(s, r, o) \in \mathcal{k}_{\text {lest }}}([\operatorname{rank}(s \mid r, o) \leq k]+[\operatorname{rank}(o \mid s, r) \leq k])
    \end{aligned}
\label{eq4:MRR}
\end{equation}
% mention we use MRR during training on subset

% Link prediction to make a proof of concept, not achieve SOTA.\\

% First:
% Node classifier by only using encoder.\\
% latent space interpolation to find analogies to smile vector in the latent space of a face VAEs.\\
% Identify if VAE learns semantics. OWL, onthology datasetbatch_size, required.\\
% Wasserstein distance???
% Link prediction?

% \begin{itemize}
%     \item Load Dataset
%     \item Convert to sparse in batches
%     \item forward pass through VAE
%     \item MPGM loss
%     \item Backward pass
%     \item test on val set
%     \item calculate MRR on subset of val set
%     \item draw graphs
% \end{itemize}

\subsection{Variational DistMult}
\label{ssec4:vdistm}

% why do we implement it, which variation
In the context of the link prediction experiment, we implement a control model. This allows a better  interpretation of the results and understanding of the impact of the isolated model parameters. From the wide range of embedding-based models, DistMult reports both good results as well as an efficient architecture. Its bilinear property aligns with the permutation invariance of the RGVAE. Based the original model we implement a variational version of DistMult to isolate the effect of variational inference on multi-relational link prediction.

% architecture link to embed
The DistMult encoder has a linear embedding layer for both sets of entities and relations. The embedded or latent representation is passed through the bilinear scoring function, which is equation \ref{eq2:distmult}. During training the model scores the triples in the training set among a number of corrupted triples. Loss function and optimizer are tunable hyperparameters. The optimal hyperparameter settings for the FB15k-237 dataset are published in Ruffellini \textit{et al.} \cite{ruffinelli_you_2019} and are used for the embedding-based link prediction experiment.

% option to train with elbo or just scores
The model implementation is adopted from Peter Bloem's work \footnote{\url{https://github.com/pbloem/embed}}. Additionally a variational module is implemented, which uses the embedding vector representation as mean and logvariance for a latent distribution from where the latent representation is sampled using the reparametrization trick. When this method is not selected, the stochastic coefficient $\epsilon$ is set to $1$ rather than a random sample from the standard normal distribution. Lastly we implement the ELBO loss, which adds a regularization term, including $\beta$ parameters to the BCE loss, thus we can only use the variational DistMult (VDistMult) in combination with BCE loss. The scoring function of the VDistMult stays identical to the original. We use the parameter settings for both the original and the variational DistMult. 

% optimal parameters

% Control model
% reparametrization trick in middle
% still same scoring function

\label{sec:mthods}

\section{Experiments \& Results}
% Experiments

This section presents the experiments and results aimed at evaluating our proposed graph generative model. First we run a grid-search on the hyperparameter space to find the optimal configuration of the RGVAE. We use the ELBO and MRR as evaluation metric. The best configurations are used to perform link prediction. Here we compare the model performance with MLP versus GCN encoder. We use the VDistMult as control model for link prediction. Finally we run two proof-of-concept experiments. The first generating triples and filter on a entity class constraining relation, thus we get an insight of how much percent of the generated triples are valid. For the experiments we use two multi-relational KG datasets.

% We covered link and node prediction and compared those to SOTA scores. Further we ran experiments on investigating the coherence of the reproduced graph structure. Lastly we measured the adherence of our model to the KG's underlying syntax.

\subsection{Data}
\label{ssec5:data}
For this sake of comparison with state of the art results, we chose the two dataset used in this field of KG link prediction, FB15k237 and WN18rr.

% Training models on each dataset for 333 epochs, without early stopping.

\textbf{FB15K-237} is a successor of the FB15K dataset, first introduced by \cite{bordes_translating_2013}, which suffered of major test leakage, meaning that triples from test set could be inferred by inverting triples from the train set. In FB15K-237, introduced in \cite{toutanova_representing_2015} these triples where removed.
The data was scraped from Freebase, while only the most frequent entities and relations were considered. The huge open-world KG Freebase \cite{bollacker_freebase_2008}, which before its discontinuation had around $1.2$ billion triples and $80$ million entities, was structured by assigning types and classes to entities and type constrains to relations. Thus, a triple can only be formed if the relation constrain matches the entity's type. Freebase was free for everyone to access and expand. This led to inconsistencies, duplicates and highly inconsistent notation, which might have been the reason for its discontinuation. Data dumps of the latest version are still available. Since the launch date of Freebase, advanced systems for storage and query of large-scale maintenance KGs have evolved \cite{cudre2013nosql}.

% textwidth for figures:
% \printinunitsof{in}\prntlen{\textwidth}

% linewidth for figures:
% \printinunitsof{in}\prntlen{\linewidth}

\textbf{WN18RR}, a dataset of synonyms and hypernyms, is a successor of yet another dataset WN18 introduced by \cite{bordes_translating_2013}. Similar to the above the original dataset suffered from test leakage, thus, an updated version without reciprocal triples was introduced by \cite{dettmers_convolutional_2018}. This dataset is characterized by its few relations and large corpus of entities. In contrary to FB15K-237, its triples require a higher level of grammatical understanding and prior knowledge of the entities.

\begin{table}[H]
  \centering
      \begin{tabular}{|l|l|l|l|}
      \hline
      \rowcolor[HTML]{EFEFEF}
      \multicolumn{1}{|c}{\textsc{Dataset}} & \multicolumn{1}{c}{\textsc{Entities}} & \multicolumn{1}{c}{\textsc{Relations}} & \multicolumn{1}{c|}{\textsc{Triples}}\\\hline
      FB15K-237     & \multicolumn{1}{c|}{$14,951$} & \multicolumn{1}{c|}{$237$} & \multicolumn{1}{c|}{$310,116$}\\
      WN18RR   & \multicolumn{1}{c|}{$40,943$} & \multicolumn{1}{c|}{$11$} & \multicolumn{1}{c|}{$93,003$} \\
      \hline
      \end{tabular}
      \caption{Statistics of the FB15K-237 \cite{toutanova_representing_2015} and WN18RR \cite{dettmers_convolutional_2018} datasets.}
      \label{tab5:data}
  \end{table}

\subsection{Hyperparameter Tuning}

In this section the hyperparameters for the VanillaRGVAE are tuned. The VanillaRGVAE has a MLP encoder, a permutation invariant loss function and a Standard Gaussian latent space prior. We run a grid search for the three hyperparameter $\beta$, $d_z$ and $d_h$ for a set of contrastive values. To reduce the computational expenses we train each model for $60$ epochs and  evaluate link prediction on a subset of $50$ triples.

We set the learning rate to $3e^{-5}$ and the maximum batch size fitting on the GPU memory. For $d_h$ we did not see any significant changes for higher values, thus we choose a lower number to reduce the total model parameters. The remaining hyperparameter did influence and the optimal setting vary for each dataset, table \ref{tab:RGVAEhyp} shows the results of our hyperparameter tuning.

% lr empirically and batchszize fixed.

% beta
For the hyperparameter tuning of $\beta \in [0,1,10,100]$ we chose significant values. With $\beta = 0$ we do not constrain our model on the Gaussian prior, thus the latent distribution can take the form of any distribution. This reduces the influence of the variational module and the model becomes closer to an autoencoder. For $\beta = 1$ we get our base model, and for $\beta \in [10,100]$ the $\beta-$VAE version.

\begin{figure}[H]
    \centering
    \begin{subfigure}{.5\textwidth}
      \centering
      \includegraphics[width=.9\linewidth, keepaspectratio]{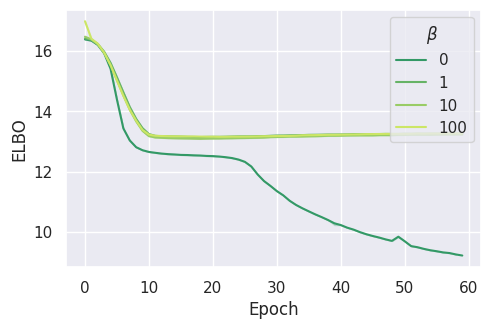}
      \caption{FB15K-237}
      \label{fig5:betafb}
    \end{subfigure}%
    \begin{subfigure}{.5\textwidth}
      \centering
      \includegraphics[width=.9\linewidth, keepaspectratio]{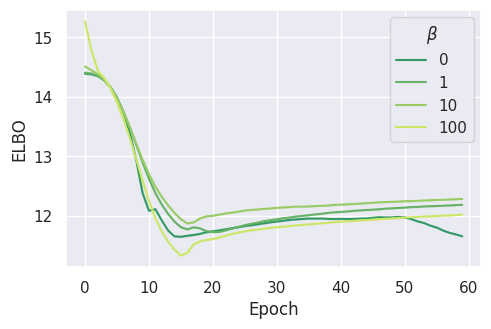}
      \caption{WN18RR}
      \label{fig5:betawn}
    \end{subfigure}
    \caption{Validation loss for RGVAE with $\beta \in [0,1,10,100]$ trained on each dataset.}
    \label{fig5:beta}
\end{figure}

Figure \ref{fig5:beta} shows the validation ELBO for the different $\beta$ values and for both datasets. We notice two interesting outcomes.  

\begin{itemize}
    \item For $\beta = 0$ converges further than the rest.
    \item The remaining values behave identical with $\beta = 100$ performing slightly better. 
\end{itemize}

Since setting $\beta = 0$ yields a VAE with a latent space which is conditionally independent from the data, this setting is not considered valuable. Thus, we chose $\beta = 100$ as default for the following experiments. This also compares with the $\beta$ values proposed by Higgins in \cite{higgins_beta-vae_2016} to achieve a factorization of the latent space.

Especially on the FB15k-237 dataset the $\beta = 0$ configuration converges to a much lower ELBO. Thus, we have the trained models perform link-prediction on a $1\%$ subset of the validation set. Figure \ref{fig5:betafbmrr} indicates an inverse correlation between the ELBO and the MRR score. 

% \begin{figure}[H]
%     \centering
%       \includegraphics[width=.45\textwidth]{graphs/plots/beta_mrr_fb.png}
%       \caption{MRR scores for different $\beta$ values on the dataset FB15k-237.}
%       \label{fig5:betafbmrr}
% \end{figure}

\begin{figure}[H]
  \centering
  \begin{subfigure}{.5\textwidth}
    \centering
    \includegraphics[height=.6\textwidth, keepaspectratio]{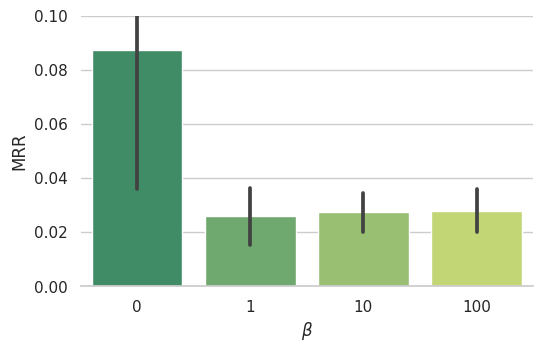}
    \caption{FB15K-237}
    \label{fig5:betamrrfb}
  \end{subfigure}%
  \begin{subfigure}{.5\textwidth}
    \centering
    \includegraphics[height=.6\textwidth, keepaspectratio]{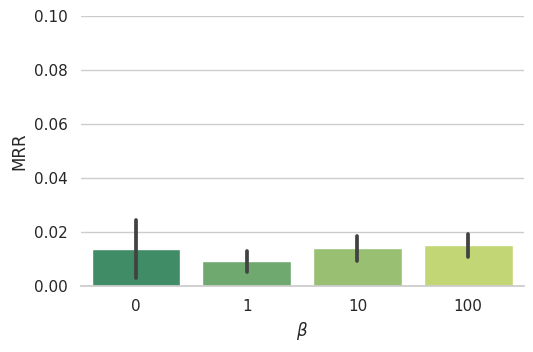}
    \caption{WN18RR}
    \label{fig5:betamrrwn}
  \end{subfigure}
  \caption{MRR scores during training for different $\beta$ values on $1\%$ of the validation set.}
  \label{fig5:betafbmrr}
\end{figure}

% Barplot IN APPENDIX  

% d_z
Experiments on the impact of $d_z$ on the ELBO show little improvement for $10<d_z<100$ and from $100<d_z<1000$ insignificant to no improvement. Thus, we chose $d_z=100$ as default for our experiments.

% d_h did not influcence
Lastly, we evaluate the models hidden dimensions $d_h$ and its influence on the ELBO and the (subset)MRR. We compare between $d_h\in [256, 512, 1024, 2048]$, while the lowest configuration performs slightly worse on the ELBO, there is no significant difference between the remaining three configurations. Considering the models parameter count we chose the $d_h=512$ as default.

\subsection{Link Prediction}

We now get to the most extensive experiment of this thesis. The results of this experiment show, whether the RGVAE is suitable for link prediction. specifically, if it is able to grasp the underlying semantics of the KG data at significantly better than the baseline of random predictions. 
For this experiment the RGVAE is trained until convergence of the ELBO. The setup for the link prediction experiment consists of two iterations for completing each head and tail entity. For each triple in the test set, the RGVAE scores all possible combinations with all entities occurring in the dataset. Since the ranking is in increasing order, we use the negative ELBO to score each triple. This allows the triple with the lowest ELBO to rank the highest. For all test triples and both iterations the scores and the index of the true triple are stored. In a post processing step the raw results are filtered and the scores for real triples, which are not the target are excluded. Further the rank of the correct triple is determined by counting all scores which are higher than the target. For scoring ties with the correct triple, the rank is settled half way. The rank is determined for every test triple and for missing head and tail entity. These ranks are used to calculate the MRR and Hits@$k$ results. 
 
In this experiment, we compare the RGVAE with MLP and GCN encoder. Further we investigate the influence of the variational inference by comparing the variational and original versions of DistMult on link prediction.

\subsubsection{RGVAE}

At this point we compare the convolutional encoder variation of our model, which we denote as cRGVAE. The experiments reveal if the convolutional architecture holds an advantage compared to the simple MLP baseline. Further a randomly initiated and untrained RGVAE is used as control model.

Due to its sparse graph computation, the RGVAE takes about 7 days to evaluate link prediction on the full test set and even 3 days when prediction tasks run parallel on a node with $4$ GPUs. Since the exemplary link prediction during experimenting with different hyperparameter already gave us an idea of the mediocre performance of our model, we chose to spare computation time and power by running link prediction on a randomly drawn one-third of the complete test set. Each run is repeated three times using a different random seed.

The results are visualized in figure \ref{fig5:lp_final}. We chose a visualization over a table, to emphasize the observed differences between encoders and datasets. These results should not be directly compared to other experiments since they were performed, as mentioned above, one-third of the test set. Note that the figures are scales to a range $[0,0.1]$ while all metrics have a maximum of $1$. The dotted line represents the baseline score of an untrained RGVAE with MLP encoder.

Comparing the RGVAE MRR score $0.08$ to the DistMult score $0.3$ on the FB15K-237 dataset our model does not perform competitively on link prediction tasks. The DistMult scores are reproduced within the scope of the embedding-based link prediction experiment, solely on the FB15K-237 dataset, using the by Ruffinelli \textit{et al.} published hyperparameter settings. For the WN18RR dataset the MRR score from \cite{ruffinelli_you_2019} is referenced.
The DistMult scores a MRR $0.45$ compared to the RGVAE score $~0.015$ on WN18RR. 
The baseline MRR score of the untrained RGVAE is $0.006$ for FB15K-237 and $0.0015$ for WN18RR. 

% Better than random?? I hope so.
In this experiment, graph convolutions do not yield an advantage over the MLP encoder. In fact, the RGVAE with GCN encoder even scores slightly worse. All versions of the RGVAE score better than the baseline, proving that the RGVAE learns a representation the data. 

%  FOR Conclusion: this might be because of the implementation of stacking the matrices.
The model scores about three times better on the FB15K-237 dataset than on WN18RR. FB15k-237 is a richer dataset with more triples and a more balanced ratio of entities to relations. WN18RR operates on only 18 relations, what makes the relation most crucial when completing a triple. The architecture of the RGVAE puts twice the emphasis entities, described by the adjacency and the node feature matrices, while the relation is only represented by the overly sparse edge attribute matrix. Thus, we could conclude that our model learns to predict based on the hidden types and topics of the entities. All possible conclusions for this are discussed in chapter \ref{sec:discus}. Relevant for this section is solely, that we chose the FB15K-237 dataset to investigate further, how well the RGVAE's grasps the underlying entity types and triple topics.   

%  REASON: fb has more relations, is a more complete KG. wn only 12 relations and more entities. Model emphasizes entities (adj and node features), relations may be less relevant, wn is more about the relation (same/ not the same). This indicates that our model grasps the hidden types and topics of the FB entities.

 \begin{figure}[H]
  \begin{subfigure}{.55\textwidth}
    \includegraphics[height=1.76in, keepaspectratio]{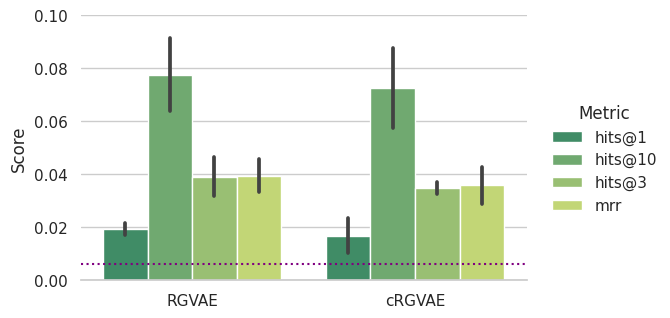}
    \caption{FB15K-237, sota MRR=$0.547$ \cite{pwcFB}.}
    \label{fig5:lpfb}
  \end{subfigure}
  \begin{subfigure}{.5\textwidth}
    \includegraphics[height=1.76in, keepaspectratio]{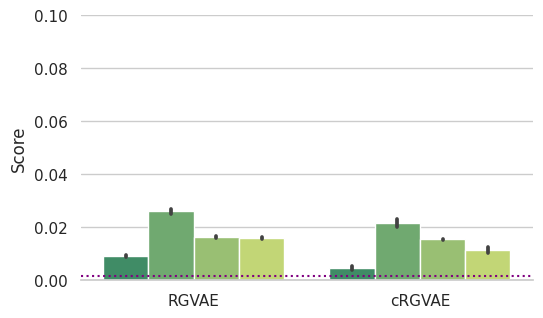}
    \caption{WN18RR, sota MRR=$0.502$ \cite{pwcWN}.}
    \label{fig5:lpwn}
  \end{subfigure}
  \caption{Link prediction scores of the RGVAE. Dotted line represents MRR baseline of an untrained model. }
  \label{fig5:lp_final}
\end{figure}

% TODO add random!!!

% Compare with vs without convolution 
% We use negative elbo as scoring function. Since elbo is aimed to be reduced and LP scores are higher better.

% We try with and without permutation

% We try the model as encoder only NO

% We use 1/3 of the test set only, randomly drawn. Run 3 times?
% Final models only 60 epochs

\subsubsection{Impact of Variational Inference and Gaussian prior}

In order to explain the poor performance of the RGVAE on the task of link prediction, we investigate the impact of the variational inference. Since the RGVAE with relaxed latent space, meaning less variance, indicated higher scores than the version with Gaussian prior, we examine the two variants by means of embedding models. The original DistMult model with optimized parameter serves as control model, while we compare it to the VDistMult, described in section \ref{ssec4:vdistm}, learning the full ELBO versus learning only on the reconstruction loss. By not including the regularization term in the loss the model is no longer bound to the Gaussian prior, which results in a relaxation of the latent space.

We train the three models for $300$ epochs solely on the FB15k-237 dataset and evaluate MRR, Hits@$1$, Hits@$3$ and Hits@$10$. Table \ref{tab5:VarDistM} shows the mean scores with $\mu \pm \sigma^2$ of three runs per model. Note that the exponent on $\sigma^2$ holds for the whole term. We can clearly see that both variational versions of the DistMult perform significantly worse than the original model. Learning on the full ELBO or only the reconstruction loss does not seem to influence the scores in this setting. This indicates, that the models performance on link prediction suffers from using variational inference.

In the last row we show the results of the RGVAE with relaxed latent space. This model was trained with $\beta=0$ thus not constraining the latent space on a Gaussian prior. The model outperforms the versions with hyperparameter choice $\beta>0$ and scores the closest to the DistMult model. The impact of $\beta$ shows in the regularization, which we tracked separately. The maximum values of$D_{K L}$ during the experiment are 

\begin{align*}
  D_{reg} &= \beta D_{K L}\left(q_{\phi}\left(\mathbf{z} \mid G\right) \| p_{\theta}(\mathbf{z})\right) \\
  \max_{\beta = 0}{ D_{K L}} &= 3506 \\
  \max_{\beta = 100}{ D_{K L}} &= 0.0154 \\
  \label{eq5:KLdifferentBeta}
\end{align*}

While the results of the relaxed RGVAE might seem promising, DistMult is a much simpler and faster link predictor, what leads us to the conclusion that the RGVAE in none of the chosen setup scores competitive with embedding-based models on this task. Note that due to the high computation cost of the RGVAE we only run the experiment once on the full dataset, thus no variance is reported.

% TODO: Answer question:Link prediction with control model:

% Trained for 300 epochs

% We see that the variational part messes everything up.

% Table:
% MRR + Hits@all + Loss

\begin{table}[H]
  \centering
      \begin{tabular}{|l|l|l|l|l|}
      \hline
      \rowcolor[HTML]{EFEFEF}
      \multicolumn{1}{|c}{\textsc{Model}} & \multicolumn{1}{c}{\textsc{MRR}} & \multicolumn{1}{c}{\textsc{Hits@$1$}} & \multicolumn{1}{c}{\textsc{Hits@$3$}} & \multicolumn{1}{c|}{\textsc{Hits@$3$}} \\\hline
      DistMult     & \multicolumn{1}{c|}{$0.2854\pm 0.0025$} & \multicolumn{1}{c|}{$0.2\pm 0.001$} & \multicolumn{1}{c|}{$0.3149\pm 0.0038$} & \multicolumn{1}{c|}{$0.4512\pm 0.0053$}  \\
      VDistMult   & \multicolumn{1}{c|}{$(0.517\pm 0.0197)e^{-3}$} & \multicolumn{1}{c|}{$(0.2442\pm 0.1994)e^{-4}$} & \multicolumn{1}{c|}{$(0.8145 \pm 0.3049)e^{-4}$} & \multicolumn{1}{c|}{$(0.399\pm 0.0576)e^{-3}$} \\
      VDistMult w/ ELBO   & \multicolumn{1}{c|}{$(0.6397\pm 0.0357)e^{-3}$} & \multicolumn{1}{c|}{$(0.57\pm 0.3046)e^{-4}$} & \multicolumn{1}{c|}{$(0.1547\pm 0.1023)e^{-3}$} & \multicolumn{1}{c|}{$(0.6351\pm 0.1992)e^{-4}$} \\
      RGVAE w/o ELBO   & \multicolumn{1}{c|}{$0.1412$} & \multicolumn{1}{c|}{$0.0981$} & \multicolumn{1}{c|}{$0.1494$} & \multicolumn{1}{c|}{$0.2275$} \\
      \hline
      \end{tabular}
      \caption{Link prediction scores of DistMult and RGVAE versions on the FB15k-237 dataset.}
      \label{tab5:VarDistM}
  \end{table}

\subsection{Impact of permutation}
% Check if adj matrix adheres to edge attribute matrix.

Furthermore we examine the influence of the permutation invariant loss function described in \ref{ssec4:loss}. During training and subset link prediction, no significant difference was observed between the RGVAE with versus without matching target and prediction graph. Yet, two observations draw our attention, namely:

\begin{itemize}
  \item The amount of nodes permuted per batch converges during training from $100$\% to exactly $80$%.
  \item The RGVAE with permutation invariant loss function learns to predict many variations of adjacency matrix while the standard model predicts similar to the target.
\end{itemize}

The first point indicates that the model learns a set of adjacency matrices, which can be permuted to match the target while optimizations the loss. Note that the generated matrix representation of the triples either has only one edge on the right upper index $A_{0,n}$ or, in the rare case of self-loops in $A_{0,0}$. The number of nodes per graph for these observations is set to $n=2$. Curiosity remains why the model converges to steadily permute $\frac{3}{5}$ of the prediction.
Secondly, we see that even when converged, the model predicts variations od the adjacency matrix very different to the target. The most common is a single edge on $A_{n,0}$ and on $A_{n,n}$. Less common and with lack of explanation are the predictions of an empty, or multi edge adjacency matrix. In contrast to this and as expected, the RGVAE with the standard loss function learns to solely predict edges on $A_{0.0}$. 
Finally we analyze the impact of permutation invariance on the experiment of generating valid triples section \ref{ssec5:syntax}.

% Permutation starts at $100\%$ at the beginning of training and converges to $60\%$.

% Model without predicts adj node always in the upper right just as the target. Model with predicts much more variations of adjacency.

% Graph of permutation during training.
% loss with vs without 

\begin{figure}
  \centering
  \begin{subfigure}{.55\textwidth}
    \includegraphics[height=1.81in, keepaspectratio]{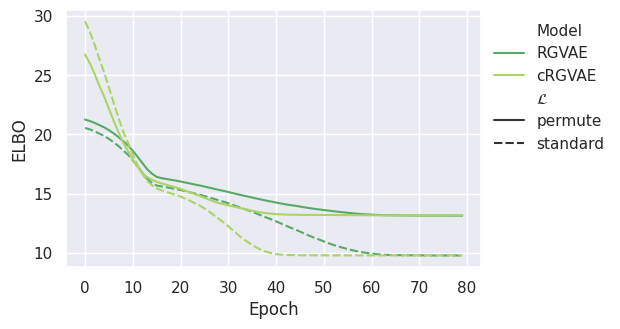}
    \label{fig5:permELBO}
  \end{subfigure}%
  \begin{subfigure}{.5\textwidth}
    \includegraphics[height=1.81in, keepaspectratio]{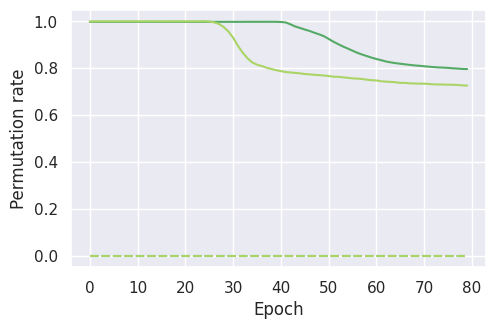}
    \label{fig5:permRate}
  \end{subfigure}
  \caption{RGVAE validation loss (a) and rate of permuted nodes (b) during training.}
  \label{fig5:permInv}
\end{figure}

\subsection{Interpolate Latent Space}

Inspired by the popular results of Higgins, who disentangled the latent space in a way that each dimension mapped to a specific facial feature from data of the FACES dataset \cite{ebner_facesdatabase_2010}. The VAE generates faces controlling feature such as age, gender and emotions by manipulating single latent dimensions \cite{higgins_beta-vae_2016}.  We run this experiment with the RGVAE on the FB15k-237 dataset, using two different interpolation methods. The latent dimension for this experiment is set to $d_{z}=10$ in order to analyze each dimension separately and the interpolation is linear with a step count of $10$.

The first experiment is linear interpolating between two triples. Therefor two valid triples from the train set are encoded into their latent representation. We chose the two semantically related triples to analyze if the linear movement in latent space correlates with an obvious semantic feature. 

\begin{center}
  \texttt{'/m/02mjmr Barack Obama', '/people/person/place\_of\_birth', '/m/02hrh0\_	Honolulu'}
  \texttt{'/m/058w5 Michelangelo', '/people/deceased\_person/place\_of\_death', '/m/06c62	Rome'}
\end{center}

% TODO present the results and link to appendix

\begin{table}[H]
  \centering
  \begin{tabular}{|c|}
  \hline
  \rowcolor[HTML]{EFEFEF} 
  \textsc{RGVAE permutation}\\ \hline
  \texttt{[[France] [/base/petbreeds/city\_with\_dogs/top\_breeds] [Imperial Japanese Army]]}\\
  \texttt{[[Cree Summer] [/tv/tv\_program/program\_creator] [David Chase]]}\\
  \texttt{[[Guitar] [/business/business\_operation/assets] [Paramount Vantage]]}\\
  \texttt{[[Democratic Party] [/music/genre/artists] [Howard Hawks]]}\\
  \texttt{[[Roy Haynes] [/people/person/spouse\_s] [The Portrait of a Lady]]}\\
  \texttt{[[Cleveland Browns] [/film/actor/dubbing\_performances] [David Milch]]}\\
  \texttt{[[Jay-Z] [/film/special\_film\_performance\_type/film\_performance\_type] [Ashley Tisdale]]}\\
  \texttt{[[Phoenix Suns] [/film/film/dubbing\_performances] [Lynn]]}\\
  \texttt{[[James E. Sullivan Award] [/organization/organization/child] [Giant Records]]}\\
  \texttt{[[Boston United F.C.] [/soccer/football\_player/current\_team] [Kensal Green Cemetery]]}\\  
  \hline
  \end{tabular}
\caption{Latent space interpolation between two triples in $10$ steps.}
\label{tab5:ipbtw2}
\end{table}
% Obama triple is not reproduced, not even close.

For the second experiment, we interpolate each latent dimension isolated in a $95\%$ confidence interval of the Standard Gaussian distribution. Starting with the encoded representation of the Obama triple, we incrementally add $z_{i} = -1.96 + j \times s$ with $s = \frac{1.96 * 2}{n_s-1}$ for the number of steps $n_s = 10$. Due to the size of the tables representing the interpolations and the low value they add to the presentation of this work, the result of this experiments can be found in the annex \ref{annexB:95}.

% Further we go ahead and test what happens if we modify one latent dimension at a time with $d_z = 10$ of a triple. TABLE: (s,r,o), x axis dims, y axis steps. $95\%$ Gaussian confidence 

% Can the model assign logical features to latent dimensions?

\subsection{Generator Validation}
\label{ssec5:syntax}

On closed-world schema-based KGs the approach for testing the validity of a new triple is to add it to the existing KG and run a ontology reasoner on it. A inconsistency in the KG appear as \texttt{null}-Class, but only if an axiom is violated. This approach works only for fully constrained KGs and is not scalable, since the reasoner recursively checks every triple for every axiom. Thus, we present an alternative and improvised way to estimate the validity of generated triples.

The FB15K-237 is a subset from the FreeBase KG \cite{bollacker_freebase_2008}, thus, even thought they are not part of the dataset, $14541$ entities have type properties in their original Freebase representation. 
Querying the last official Freebase dump, as proposed by Xie \textit{et al.} \cite{xie2016representation}, we get the types for each entity in the FB15K-237 dataset, With exception of $8$ entities, which could not be found in the query.

Our approach is to randomly generate triples, from signals randomly drawn from a Standard Gaussian distribution. Then to filter those triples on predicates which contain the type \textit{people}, which is within the top 10 most common Freebase types. We differentiate between base-class types subclass types, both can contain the word \textit{people}. The entity \texttt{['/m/02mjmr Barack Obama']} has between many others the type \texttt{[/people/measured\_person]}. Here the base class is \textit{people} and the subclass \textit{measured\_person}. We could filter directly on subclasses, but we chose to give our model more creative freedom and filter for \textit{people} in the full set of types. Using this choice, the generated triples are scored on logic rather than facts. E.g. any person can hypothetically be a \textit{measured\_person}, understanding this implies semantical reasoning, while differentiating between which person is and is not a \textit{measured\_person} implies contextual knowledge. Thus, we use the base type \textit{people} to validate triples.
Furthermore the relations are inconsistent in their notation, partly not only having a head type constrain. Thus, we check only the head entity for the key type. 

\begin{itemize}
  \item From $14541$ entities, $5283$ contain the keyword people, or $36.332$\%.
  \item From $237$ predicates, $25$ contain the keyword people, or $10.549$\%.
  \item From $310116$ triples, $47354$ contain the predicate keyword people, or $15.269$\%.
\end{itemize}

Considering these facts, we calculate the marginal probability of guessing a head entity $s$ of type \textit{people}, given a triple which contains the type \textit{people}. Without prior knowledge $p(s_{p})$ and $p(r_{p})$ are conditionally independent. The probability is calculated as:

\begin{equation}
  p(s_{p} \mid r_{p}) = p(s_{p}) = 0.3633
  \label{eq5:randomValid}
\end{equation}

For this experiment we generate triples until $10^5$ contain the key type. Those filtered triples are validated on the type of the head entity and compared to the full dataset for novelty. Here we again compare the performance between the RGVAE with the two different encoder architectures. Furthermore the models are trained both with regular and permutation invariant loss function. Lastly, the experiments are repeated for sampling the latent signal from $\mathcal{N}(0,1)$ and for sampling from $\mathcal{N}(0,2)$.  We average the accuracy of three runs of generating a valid triple and of this triple being unseen in the dataset. The results for valid triples are shown in figure \ref{fig5:syntax}. The dotted horizontal line indicates the random probability of generating a valid triple, calculated in equation \ref{eq5:randomValid} and $\sigma^2_1$ and $\sigma^2_2$ denote the variance for the different latent space distributions.

\begin{figure}[h]
  \centering
  \includegraphics[height=.3\textwidth, keepaspectratio]{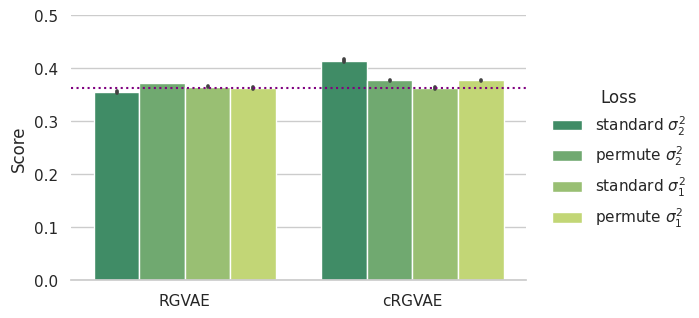}
  \caption{Accuracy of generating valid triples.}
  \label{fig5:syntax}
\end{figure}

To our disappointment the model does not perform significantly better than the baseline of random predictions. Neither the choice loss function nor the doubled variance show a correlation with the accuracy. The only configuration standing out is the RGVAE with convolutional encoder, standard loss function and $\sigma^2=2$, scoring $4$\% higher than the baseline. From all valid generated triples $100\pm 0.001$\% are new and unseen in the dataset. Coming back to \texttt{'/m/02mjmr Barack Obama'}, we filter the unseen triples for the first three appearances of this entity. These are displayed in table \ref{tab5:genTriples} for every variation of the RGVAE and in the same order as in figure \ref{fig5:syntax}. Note that in order to fit the triples to this thesis' text width, some relations with object type specification have been cropped at \texttt{./} as well as entity names with overlength. 

The generated triples confirm the accuracy results. With exception of the triple 
\begin{center}
  \texttt{Barack Obama, /people/person/places\_lived./people/place\_lived/location, Casablanca} 
\end{center}

all remaining triples violate common sense logic. The RGVAE does not differentiate between the types gender, location, movie, person or medicine. It even goes so far to state that Obama was born in \texttt{'Multiple sclerosis'}. While this might sound funny it also clearly indicates, that our model did not learn the underlying semantics of this real world KG.

While investigating the model and the generated triple set, we notice two outcomes. Neither the augmentation of the variance nor the enabling of the permutation invariance has an impact on either of the model with two encoder versions. The regularization loss converges during training to zero, meaning that the model learns a latent representation of the dataset as  nearly perfect Standard Gaussian distribution. Yet, even when sampling latent signal from the exact same distribution, we notice that each model repeatedly predicts combinations of a small subset of entities and relations, e.g. the RGVAE version which did not predict the Obama entity once in a total of $111583$ valid triples. This also aligns with the interpolation results, where we observed a static relation for the full grid search of the latent space. If we look at the gradient and parameter values $\phi$ and $\theta$ of the MLP encoder and decoder, we see a much higher variance and gradients for $\phi$. The decoder shows higher values and variance for a small subset of neighboring parameter, while the remaining parameters converge to a very similar and low value. This indicates that the encoder learns very well to represent each different triple as Standard Gaussian latent representation. The decoder MLP on the opposite seems to ignore most of this representation by assigning vanishing values to the connected parameters. Intuitively it seems that the decoder learns to interpret the part of the latent representation corresponding to the adjacency matrix and minimizes as well as stabilizes the loss of the edge and node attribute matrix by uniformly distributing their probability. This leads to the decoder reconstructing edge and node attribute randomly. Further, depending on the values of $\theta$ when the model finishes learning, the decoder repeatedly predicts the same subset of entities and relation independent of the latent signal $z$. If we look at the flattened representation of our input graph, we see that the part representing the adjacency matrix is way shorter and has a tractable mean of $\frac{1}{4}$ while the mean for the edge and node attribute matrices are $\frac{1}{{4 \times 1345}}$ and $\frac{1}{14951}$. The problem of the RGVAE decoder partly ignoring the latent input and potential solutions are discussed further in section \ref{ssec7:collapse}. 

\begin{table}[H]
  \begin{tabular}{|c|}
  \hline
  \rowcolor[HTML]{EFEFEF} 
  \textsc{RGVAE standard} $\sigma_2^2$\\ \hline
  \texttt{[[Barack Obama]	[/people/person/places\_lived./people/place\_lived/location]	[Casablanca]]}\\
  \texttt{[[Barack Obama]	[/people/person/place\_of\_birth]	[Sarah Silverman]]}\\
  \texttt{[[Barack Obama]	[/people/person/place\_of\_birth]	[The League of Extraordinary Gentlemen]]}\\ \hline
  \rowcolor[HTML]{EFEFEF} 
  \textsc{RGVAE permuted} $\sigma_2^2$\\ \hline
  \texttt{[[Barack Obama]	[/people/ethnicity/geographic\_distribution]	[End of Watch]]}\\
  \texttt{[[Barack Obama]	[/people/profession/specialization\_of]	[Montgomery County]]}\\
  \texttt{[[Barack Obama]	[/people/cause\_of\_death/people]	[WWE Superstars]]}\\ \hline
  \rowcolor[HTML]{EFEFEF} 
  \textsc{RGVAE standard} $\sigma_1^2$\\ \hline
  \texttt{[[Barack Obama]	[/people/person/place\_of\_birth]	[Academy Award for Best Sound Editing]]}\\
  \texttt{[[Barack Obama]	[/people/person/places\_lived./people/place\_lived/location]	[Stan Lee]]}\\
  \texttt{[[Barack Obama]	[/people/person/place\_of\_birth]	[Multiple sclerosis]]}\\ \hline
  \rowcolor[HTML]{EFEFEF} 
  \textsc{RGVAE permuted} $\sigma_1^2$\\ \hline
  \texttt{[[James Brolin]	[/people/person/places\_lived./people/place\_lived/location]	[Barack Obama]]}\\
  \texttt{[[Barack Obama]	[/people/person/spouse\_s./people/marriage/location]	[D.C. United]]}\\
  \texttt{[[Jim Sheridan]	[/people/person/religion]	[Barack Obama]]}\\ \hline
  \rowcolor[HTML]{EFEFEF} 
  \textsc{cRGVAE standard} $\sigma_2^2$\\ \hline
  \texttt{}\\
  \texttt{None}\\
  \texttt{}\\ \hline
  \rowcolor[HTML]{EFEFEF} 
  \textsc{cRGVAE permuted} $\sigma_2^2$\\ \hline
  \texttt{[[Pinto Colvig]	[/people/deceased\_person/place\_of\_burial]	[Barack Obama]]}\\
  \texttt{[[Helena Bonham Carter]	[/people/person/gender]	[Barack Obama]]}\\
  \texttt{[[John Buscema]	[/people/person/spouse\_s./people/marriage/spouse]	[Barack Obama]]}\\ \hline
  \rowcolor[HTML]{EFEFEF} 
  \textsc{cRGVAE standard} $\sigma_1^2$\\ \hline
  \texttt{[[Suhasini Ratnam]	[/people/person/sibling\_s./people/sibling\_relationship]	[Barack Obama]]}\\
  \texttt{[[Barack Obama]	[/people/person/gender]	[Deva]]}\\
  \texttt{[[Barack Obama]	[/people/person/sibling\_s./people/sibling\_relationship]	[Niagara Falls]]}\\ \hline
  \rowcolor[HTML]{EFEFEF} 
  \textsc{cRGVAE permuted} $\sigma_1^2$\\ \hline
  \texttt{[[Jonathan Rhys Meyers]	[/people/person/nationality]	[Barack Obama]]}\\
  \texttt{[[Barack Obama]	[/people/person/sibling\_s./people/sibling\_relationship]	[Motherwell F.C.]]}\\
  \texttt{[[Pinto Colvig]	[/people/deceased\_person/place\_of\_burial]	[Barack Obama]]}\\ \hline
  \end{tabular}
\caption{Generated and unseen knowledge.}
\label{tab5:genTriples}
\end{table}

\subsection{Delta Correction}
\label{ssec5:delta}

% Explain quick delta implementation 
A solution to the observed phenomenon, which is known as decoder collapse, is the use of a $\delta$ parameter, presented by Razavi \textit{et al.} in \cite{razavi_preventing_2018}. The regularization term is expanded by subtracting the $\delta$ parameter and by taking the absolute value. This forces the latent space into a truncated Gaussian distribution. The simple nature of this solution allows us to integrate it in the RGVAE setup and evaluate its impact on the previous experiments.

% Explain new experiments
In a final experiment the RGVAE is trained in $4$ different modes, with $\delta \in [0, 0.6]$ and using the standard versus the graph matching loss function. Since we do not expect this new parameter to have a significant impact on the link prediction results, the most informative and comparable experiments are latent space interpolation and the generation of valid triples.

% show interesting interpolation
Table \ref{tab5:ipbtw2Delta} shows the results for $\delta = 0.6$ and graph matching. The model shows the same behavior as the for $\delta = 0$. The start and end triple do not match the encoded target triples and $9$ out of $10$ interpolation steps predict the same relation. A difference can be seen in the traversing of each latent dimension. Here minimal semantic coherence can be observed \textbf{(Table in Appendix)}.
The new parameter $\delta$ did not raise the ratio of valid generated triples above the baseline.
A difference between standard, appended in the annex \ref{annexA:ipbtw2DeltaNoPerm}, and graph matching loss could also not be observed. The results of the full latent space interpolation for both variations of loss functions are appended in the annex \ref{annexB:95}.

\begin{table}[H]
  \centering
  \begin{tabular}{|c|}
  \hline
  \rowcolor[HTML]{EFEFEF} 
  \textsc{$\delta$-RGVAE permutation}\\ \hline
  \texttt{[[The Gift] [/user/tsegaran/random/taxonomy] [Alyson Hannigan]]}\\
  \texttt{[[Philip K. Dick Award] [/people/deceased] [CSI: Crime Scene Investigation]]}\\
  \texttt{[[The Crying Game] [/user/tsegaran/random/taxonomy] [Stephen Tobolowsky]]}\\
  \texttt{[[Anne Hathaway] [/user/tsegaran/random/taxonomy] [Billie Joe Armstrong]]}\\
  \texttt{[[Mehcad Brooks] [/user/tsegaran/random/taxonomy] [Melissa Leo]]}\\
  \texttt{[[Soap] [/user/tsegaran/random/taxonomy] [United States of America]]}\\
  \texttt{[[2002 Winter Olympics] [/user/tsegaran/random/taxonomy] [Swept Away]]}\\
  \texttt{[[Oliver Platt] [/user/tsegaran/random/taxonomy] [The Good] [the Bad] [the Weird]]}\\
  \texttt{[[Danny DeVito] [/user/tsegaran/random/taxonomy] [Percussion]]}\\
  \texttt{[[Mr. \& Mrs. Smith] [/user/tsegaran/random/taxonomy] [Marie Antoinette]]}\\  \hline
  \end{tabular}
\caption{RGVAE latent space interpolation with $\delta = 0.6$ and graph matching.}
\label{tab5:ipbtw2Delta}
\end{table}

% show parameters if different
Finally, we plot the parameter values, both weight and bias to visualize the collapsing decoder. The values are logscaled and plotter per layer number of encoder and decoder as described in table \ref{tab4:archcompare}. The parameters of the decoder behave equally for both $\delta$ values. All values of the 3rd layer conglomerate around zero revealing a bottleneck of the information stream in the decoder. This indicates that the problem of collapsing decoder was not solved by the addition of the new parameter. The parameter distribution for the RGVAE with standard loss are appended in the annex \ref{annexC:deltaParamsP0} and \ref{annexC:normParamsP0}. The impact of the standard loss function is a more narrow distributed decoder. Since the GCN encoder has different layers, a direct parameter comparison to the MLP encoder is not possible, yet we observe the same parameter distribution for the decoder. Note in figure \ref{fig5:deltaParams} the different scales of the value axis, which are $10^2$ larger for $\delta = 0.6$ than for $\delta = 0$.

\begin{figure}[H]
  \centering
  \begin{subfigure}{\textwidth}
    \includegraphics[width=\textwidth]{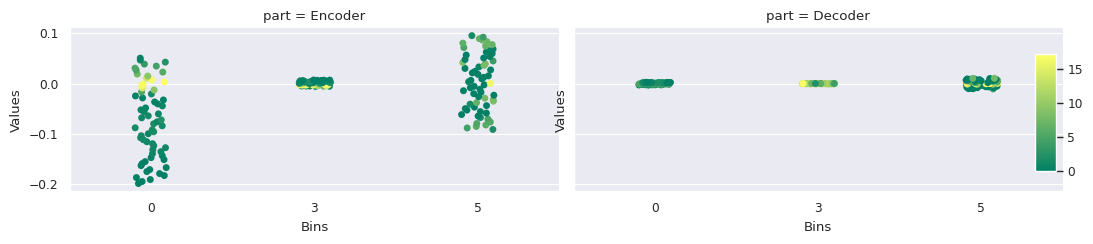}
    \caption{Weight}
    \label{fig5:deltaParamsW}
  \end{subfigure}
  \begin{subfigure}{\textwidth}
    \includegraphics[width=\textwidth]{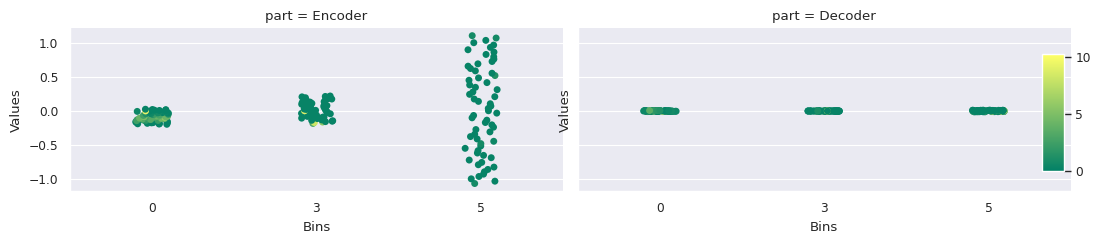}
    \caption{Bias}
    \label{fig5:deltaParamsB}
  \end{subfigure}
\caption{Parameter values per layer of the RGVAE encoder and decoder with $\delta=0.6$.}
\label{fig5:deltaParams}
\end{figure}

\begin{figure}[H]
  \centering
  \begin{subfigure}{\textwidth}
    \includegraphics[width=\textwidth]{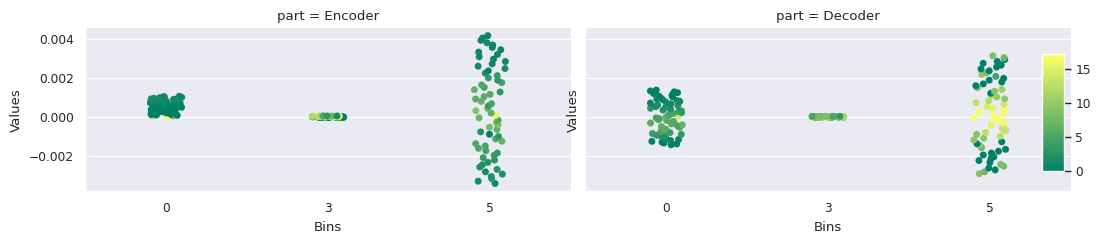}
    \caption{Weight}
    \label{fig5:normParamsW}
  \end{subfigure}
  \begin{subfigure}{\textwidth}
    \includegraphics[width=\textwidth]{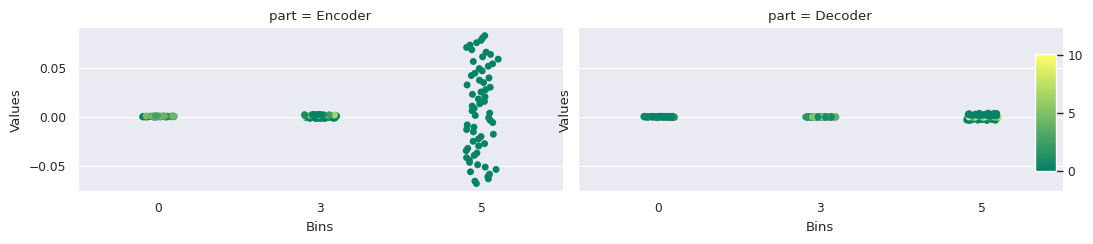}
    \caption{Bias}
    \label{fig5:normParamsB}
  \end{subfigure}
\caption{Parameter values per layer of the RGVAE encoder and decoder with $\delta=0$.}
\label{fig5:normParams}
\end{figure}

% \section{Results}
% \input{sections/section6}

\section{Discussion \& Future Work}

% TODO compare to molecules - point out differences

In this final chapter, we discuss our results from different experiments, emphasizing the comparison to Simonovsky's work on molecule generation. The fact that we did not achieve similarly successful results does not hinder us from analyzing the responsible factors and causalities. Despite the nature of our results we are happy to propose possible solutions and while being of the opinion that asking the right question is more valuable than providing the correct answer, we question the value of these solutions for further research.

\subsection{Experiment Take-away}

% Well well well
% We did link prediction because it is the most common experiment in this field

%  We see loss converges perfectly 
The first relevant observation made is the lower ELBO score to which the RGVAE converges when using $\beta=0$, meaning it is only trained on the reconstruction loss.
We tested the assumption that the model performs better without regularization term. This leads to the assumption that the RGVAE performs better without prior constrain, yet a fully unconstrained VAE, defies the its purpose when it comes to disentangling the latent space and sampling.
The tuned RGVAE shows poor results on the task of link prediction, with slightly worse scores for the convolutional encoder. In order to find the isolate the causalities, we experiment further on link prediction with two variational version of DistMult learning on the full ELBO and only on the reconstruction loss. Both models scored equally bad compared to the original model, indicating that DistMult and embedding-based link predictors in general are not complex enough for variational inference. Following our prior assumption we added the unconstrained RGVAE to this comparison, which led to significantly higher scored than the all versions of the tuned RGVAE and the variation DistMult. Yet, the scores were not competitive with state of the art link predictors of with much lower parameter count, thus we do not recommend further researching in the direction of VAE as link predictor.

The graph matching loss function does not seem to have any impact on the link prediction results, yet we cannot draw a conclusion here since the poor performance is linked to a different causality which might impede the graph matching from adding value. To measure the percentage of permutation we adopt Simonovsky's method of averaging the identity of permutation matrix $X$, which is invariant for permutation equivalent nodes. We see that during training the permutation rate converges to $60$\%.

Comparison of the two datasets, showed a better performance on the FB15K-237 dataset. While both datasets are real-world KGs, WN18RR is composed of synsets with vast discrepancy between the relation and entity count, resulting in a lower chance of generating a valid triple. Thus, we decided to continue the interpolation and graph generation experiments only on the FB15k-237 dataset. 

For the last two experiments we trained the RGVAE with $d_z=10$ for better disentanglement and analysis of each latent dimension.
The interpolation experiment between two triples reveals that the RGVAE in all versions fails to reconstruct the start and end triple. We notice an influence of the graph matching loss, namely the model predicting constantly the same relation versus the standard loss, where the model predicts randomly from a subset of triples. Same can be said for the predicted entities, which form a small subset of what seems to be the most frequent occurrences in the dataset. Empirically we can say that the prediction has little prior conditioning on the latent signal.

These last experiment is designed and evaluated in a similar fashion as Simonovsky did to generate valid molecular graphs. We interpolate the full latent space for $-\sigma^2 < z < \sigma$ and $-2\sigma^2 < z < 2\sigma$. Based on our proposed validation criteria, the RGVAE generated the same rate of valid triples as random. We observe further independence between the decoder and the latent signal. This slightly compares with Simonovsky's result where the generated valid triples had a variance of only $10$\%.
% Interpolation
%  like molecules

% Why does graph matching change that? 
% Higher dimension of freedom when generating graphs using Graph matching.

% All valid generated triples are unseen in both train and test set. 

% Simonovsky also reports low variance of valid triples 10\%

\subsection{Research Answer}

% Answer it right!

% How successful is a VAE in representation learning on real world KG compared to molecule graph data and what is the impact of each major hyperparameter?

%  Overall
To answer the initial research question we first summarize the influence of the main hyperparameter.
%  Convolutions
A convolutional encoder does not seem to contribute to the performance of the RGVAE. This could be caused by the simple implementation of using node and edge attributes both as feature matrix, or be due to the decoder issue, explained in section \ref{ssec7:collapse}, that the prediction is independent of the encoding. 
%  graph matching
Graph matching does have a noticeable impact. The model expresses more creative freedom when predicting the adjacency. Especially for experiments with larger subgraphs we recommend to keep investing in this method.

%  Prior, stochasticity
Variational inference is part of the VAE and allows disentanglement of the latent space, thus, even considering that the unconstrained model scored better, we recommend keeping the VAE structure. Instead of the full stochastic module, we assume the Standard Gaussian prior to be the reason behind the poor representation and recommend comparing alternative prior constrains.
Overall we can answer the research question:

\begin{center}
    \textit{The RGVAE failed in representation learning of real-world KG due to a decoder collapse.}
\end{center}

Yet, metaphorically, this is not the end of the book, but rather the beginning of a new chapter. In the remaining part of this discussion we analyze the underlying problem of the RGVAE and propose a variety of solutions.

\subsection{Decoder Collapse}
\label{ssec7:collapse}

%  what is decoder collapse
To introduce the main part of our conclusion we explain a phenomenon observed when training GANs. A GAN is yet another generative model which based on its potential to generate high resolution images, has drawn much attention in recent years. It consists of a generator, whose task it is to decode a latent signal to a an image, and a discriminator, which given the fake image and the real data has to distinguish between them. Training of GANs is unstable and one of the major problem which occur is \textit{mode collape}. The generator learns to fool the discriminator by generating only a single mode with high precision such that the discriminator classifies it was real image \cite{goodfellow2014generative}.

While VAEs cannot suffer from mode collapse, because they backpropagate over the predicted distributions of all modes, it has a related phenomenon. \textit{Decoder collapse} occurs when the decoder has enough capacity to choose not to consider the latent signal and instead stores the information to reconstruct the data's distribution partly or solely in its parameters. This means that the generated data is not conditioned on the latent signal anymore. The cause of this problem is found in the regularization term $D_{K L}\left(q_{{\phi}}\left(\mathbf{z} \mid \mathbf{x}\right) \| p_{{\theta}}(\mathbf{z})\right)$, more specifically in the constrain on a standard Gaussian prior. The multidimensional encoding of $d_z>1$ the approximated posterior is a mixture of Gaussians, which can only match the multivariate Normal distribution, in the case of all $\mathbf{\mu}_z=0$ and $\Sigma^2_z=\mathbb{I}$, what also implies that no information is encoded. Thus, the VAE has to decide if storing information in the latent vector is necessary to model the dataset distribution $p(x)$. If the penalty for altering the latent Normal distribution outweighs the benefits of the additional information towards the reconstruction loss, the VAE choses for \textit{decoder collapse}. 

%  Cite https://towardsdatascience.com/with-great-power-comes-poor-latent-codes-representation-learning-in-vaes-pt-2-57403690e92b

% when we use a decoder with so much capacity that it chooses to not store information in the latent code at all, a result that leaves important information about our distribution locked up in decoder parameters, rather than neatly extracted as an internal representation.

% the network will only choose to make its z value informative if doing so is necessary to model the full data distribution, p(x). Otherwise, the penalty it suffers for using an informative z will typically outweigh the individual-image accuracy benefit it gets from using it.

% analyze results of interpolation and generation
Exactly this phenomenon can be observed in the triple interpolation and generation experiments. The RGVAE converges rapidly during training, minimizing the regularization loss close to zero and stabilizing the reconstruction loss by learning or matching the adjacency and predicting the most frequent node and edge attributes. This falsely indicates that the model has learned to reproduce the data. Yet, when sampling using only the decoder, the generated triples repeat combinations of a subset of entities and relations which are more frequent in the dataset. During interpolation of the latent space, the model predicts steadily the same relation. Subject and object seem to a certain degree be conditioned on the latent signal, altering while traversing the latent space, yet no clear disentanglement is observed. The final observation on the node attributes is, that the model shows a preference to predicts people's name and film's name entities as subject and location or music genre entities as object.

An interesting question, when observing the RGVAE generate all triples with the same single relation is, if the reason for this behavior can still be attributed to decoder collapse, as it rather collapses only on one mode?

A possible explanation is that the decoder learns to ignore the latent signal only for the relation prediction, still updating the full decoder during training and thus predicting different values after every update. Yet, when done training and used solely as generator, the decoders parameters stay constant and predicting the relation uninformed by any latent signal, which results in the collapse on the same relation prediction.

The results of the last experiment are surprising. While the RGVAE performed significantly better than the untrained baseline on link prediction, the ratio of valid generated triples does not differ from random sampling for any of the different model variations. Further we again notice the predominance of the more frequent data entities and relations.

Why does the RGVAE perform better on link prediction than on generating valid triples, if both tasks require representational understanding of the data?

Crucial in this comparison could be that link prediction is scored by the ELBO, including both encoder and decoder, while triples are generated only by the decoder.

% For link prediction this means that the reconstruction loss is not helpful at all. 
Projecting this finding on the link prediction experiment, we question, why the model scored better than the random baseline. Possible reasons are for once, that the encoder learned to encode the dataset close to a Normal distribution, but when encountering unseen triples, the mean and variance of the encoding vary. A second reason is, that every triple in the test set is corrupted in all possible combinations, including the less frequent entities. The collapsed decoder subsequently keeps generating frequent triples, thus the reconstruction loss between a less frequent combination of entities and the collapsed prediction is be higher than for a frequent combination. Therefore the model learns score the real triple is higher but for the wrong reason. The RGVAE with $\beta = 0$ and therefore unconstrained on the Standard Gaussian prior shows the best results between the different model settings. We can assume that the model learns a latent representation for subject and object. Despite this improvement, is it probable that the decoder still collapses on the reconstruction of the relation index, which explains the remaining scoring gap to the much simpler DistMult model.

\subsection{VAE surgery}
\label{ssec7:solutions}
% Propose solutions: Elbosurgery, adversial(WAE),  recurrent(lossy)

Similar problems in different fields have been encountered for generative VAE applications. While the author of this thesis whishes to have drawn these parallels earlier, all the proposed solutions imply a significant modification of the VanillaVAE, thus would not align with this work's research question. We present three approaches from different literature which tackle the VAE mode collapse in the filed of image and voice generation.

%  ELBO surgery
In a publication of Adobe Research and Google Brain, Hoffman \textit{et al.} propose an elegant modification of the variational evidence lower bound \cite{hoffman2016elbo}. 
More specifically they change the VanillaVAE's regularization term, where instead of imposing a Standard Gaussian prior on the full latent distribution, the prior is imposed on the mixture of all single latent dimension, giving space for more expressive probability distributions such as truncated Gaussians. 
Further, a index-code mutual information term is added, intended to maximize the mutual information between every index of the observation and $z$. While the reconstruction term enforces to encode every feature of $x$ in a corresponding latent dimension, the  information term opposes this by maximizing the mutual information between all  $x_i$ and $z_i$. The information term compares compact to the reconstruction loss, yet it is enough to prevent decoder collapse.

% Lossy Auto Encoder
Kingma \textit{et al.} turn towards a autoregressive solution in their publication \cite{chen_variational_2017}. Their VAE model generates images recursive per pixel, each conditioned on previous point, using both RNN and RCN as decoder. Their solution to the decoder collapse is a normalizing flow, which predicts the encoder posterior $q_{\phi}(z \mid x)$. Besides not suffering from decoder collapse, this model can be set to discard irrelevant information in the data. As downside, the autoregressive nature causes slow image generation.

% Wasserstein or AAE
The last approach we present closes the circle to the introduction of this section. The paper \textit{Wasserstein Auto-Encoders} \cite{tolstikhin_wasserstein_2019} propose a combination between GAN and VAE. A model which uses the VAE's encoder-decoder architecture but instead of the normal regularization term, the posterior distribution is learned and penalized by its Wasserstein distance to the data distribution. This is in fact a generalization of the adversarial loss of generator and discriminator. Since our task of generating triples would benefit from a precise prediction, such as GANs achieve on image generation, we question, if employing adversarial loss also on the reconstruction term would yield better generation results in this field than the exact index-wise loss?

% Two different papers, same principle:

% Use GAN loss for VAE latent space distribution. Generator produces a distribution and discriminator tys to tell if its fake or true.

% Similar approach with Wasserstein distance as regularization loss. 

% Could we also usefully employ adversarial loss on the reconstruction part of the network (that is: have a discriminator try to tell apart input and reconstruction), to get away from the over-focus on exact detail reconstruction that comes with pixel-wise loss

% Remaining: Amortized Inference Regulation and Skip Connections 
Besides these three, numerous other approaches have been proposed. Worth mentioning because of their originality are the idea of adding skip-connections between latent space and the decoder's hidden layer \cite{dieng_avoiding_2019} as well as regularizing the amortized inference \cite{shu_amortized_2019}.  

% delta-VAE <--

% Regulation of the amortized inference.

% Skip connections between latent space and hidden layer of the decoder.

% Two layerVAE

\subsection{Future Work}

There are many avenues to follow for future work. Obviously the first step is to fix the collapsing decoder. The suggested solutions should be compared before making a choice, since none has yet been shown to work on KG VAEs.

Once the decoder reaches the point of predicting the right graph based on its latent representation, the RGVAE should perform significantly better on both link prediction and generating valid triples. Comparing the RGVAE's efficiency at both task, generating triples is efficient and fast whereas link prediction inefficient and therefore disproportionately slow. Optimizing this could yield a bigger challenge than improving the scores. 

Besides the model we should also look at the data, and question the representation. The adjacency matrix can be inferred from the edge attribute matrix and thus, is obsolete. Since it does not contain additional information, does the model perform worse without it? Approaching this thought from a different angle, we could leave the adjacency as it is and represent the node and edge indices as continuous number, as it is done for color scales of images. This would greatly reduce the number of parameters, but certainly also imply new challenges. Important in this context is that the adjacency is represented in sparse format since this the experiments on triples is but a proof of concept for larger subgraphs. 

The model's mayor hurdle seems to be predicting the correct relation index. While interpolating the entire latent space, less than $10$ different relations were generated. A further challenge is the varying count and position of the edges, in contrast to the number of nodes which is constant. Thus, we wonder if a setup of two VAEs, where the first one predicts the adjacency and node attributes and the second recurrent one, while conditioned on the predicted edges of the first VAE, predicts the edge attribute.

Finally and making use of the batch wise implementation of the max-pooling graph matching algorithm presented in this work, we suggest to explore a more efficient and \textit{intelligent} graph matching approach. Would it be possible drastically reduce complexity by training a neural network to predict the permutation matrix based on target and prediction graph?
% with one of the presented solutions, starting with the simplest.

% Is the adjacency matrix even necessary? 

% Basing on the believe that further research will be fruitful, we recommend:

% \begin{itemize}
%     \item Smarter approach for graph matching, let a NN learn the best permutation given the targe and prediction
%     \item Two VAEs, first one-shot only the adjacency, second recurrent edgewise including entity and relation index.
%     \item Disentangle the latent space and condition on text
%     \item NF prior
% \end{itemize}

The author hopes that his questions spark creative ideas for further research on generative models for KGs and looks with excitement towards future findings.

% \textbf{Happy New Year!}

\label{sec:discus}

% \section*{Ideas}
% \input{ideas}

% \section*{Open Issues}
% \input{issues}

\newpage

\printbibliography

@misc{wandb,
title = {Experiment Tracking with Weights and Biases},
year = {2020},
note = {Software available from wandb.com},
url={https://www.wandb.com/},
author = {Biewald, Lukas},
}

@ARTICLE{2020SciPy-NMeth,
  author  = {Virtanen, Pauli and Gommers, Ralf and Oliphant, Travis E. and
            Haberland, Matt and Reddy, Tyler and Cournapeau, David and
            Burovski, Evgeni and Peterson, Pearu and Weckesser, Warren and
            Bright, Jonathan and {van der Walt}, St{\'e}fan J. and
            Brett, Matthew and Wilson, Joshua and Millman, K. Jarrod and
            Mayorov, Nikolay and Nelson, Andrew R. J. and Jones, Eric and
            Kern, Robert and Larson, Eric and Carey, C J and
            Polat, {\.I}lhan and Feng, Yu and Moore, Eric W. and
            {VanderPlas}, Jake and Laxalde, Denis and Perktold, Josef and
            Cimrman, Robert and Henriksen, Ian and Quintero, E. A. and
            Harris, Charles R. and Archibald, Anne M. and
            Ribeiro, Ant{\^o}nio H. and Pedregosa, Fabian and
            {van Mulbregt}, Paul and {SciPy 1.0 Contributors}},
  title   = {{{SciPy} 1.0: Fundamental Algorithms for Scientific
            Computing in Python}},
  journal = {Nature Methods},
  year    = {2020},
  volume  = {17},
  pages   = {261--272},
  adsurl  = {https://rdcu.be/b08Wh},
  doi     = {10.1038/s41592-019-0686-2},
}

@inproceedings{hoffman2016elbo,
  title={Elbo surgery: yet another way to carve up the variational evidence lower bound},
  author={Hoffman, Matthew D and Johnson, Matthew J},
  booktitle={Workshop in Advances in Approximate Bayesian Inference, NIPS},
  volume={1},
  pages={2},
  year={2016}
}

@article{ebner_facesdatabase_2010,
	title = {{FACES}—{A} database of facial expressions in young, middle-aged, and older women and men: {Development} and validation},
	volume = {42},
	issn = {1554-351X, 1554-3528},
	shorttitle = {{FACES}—{A} database of facial expressions in young, middle-aged, and older women and men},
	url = {http://link.springer.com/10.3758/BRM.42.1.351},
	doi = {10.3758/BRM.42.1.351},
	language = {en},
	number = {1},
	urldate = {2021-01-07},
	journal = {Behavior Research Methods},
	author = {Ebner, Natalie C. and Riediger, Michaela and Lindenberger, Ulman},
	month = feb,
	year = {2010},
	pages = {351--362},
}

@article{belli_image-conditioned_2019,
	title = {Image-{Conditioned} {Graph} {Generation} for {Road} {Network} {Extraction}},
	url = {http://arxiv.org/abs/1910.14388},
	urldate = {2020-05-08},
	journal = {arXiv:1910.14388 [cs, stat]},
	author = {Belli, Davide and Kipf, Thomas},
	month = oct,
	year = {2019},
	note = {arXiv: 1910.14388},
}

@article{kingma_introduction_2019,
	title = {An {Introduction} to {Variational} {Autoencoders}},
	volume = {12},
	issn = {1935-8237, 1935-8245},
	url = {https://www.nowpublishers.com/article/Details/MAL-056},
	doi = {10.1561/2200000056},
	language = {English},
	number = {4},
	urldate = {2020-05-07},
	journal = {Foundations and Trends® in Machine Learning},
	author = {Kingma, Diederik P. and Welling, Max},
	month = nov,
	year = {2019},
	note = {Publisher: Now Publishers, Inc.},
	pages = {307--392},
}

@article{vrandevcic2014wikidata,
  title={Wikidata: a free collaborative knowledgebase},
  author={Vrande{\v{c}}i{\'c}, Denny and Kr{\"o}tzsch, Markus},
  journal={Communications of the ACM},
  volume={57},
  number={10},
  pages={78--85},
  year={2014},
  publisher={ACM New York, NY, USA}
}

@inproceedings{perozzi2014deepwalk,
  title={Deepwalk: Online learning of social representations},
  author={Perozzi, Bryan and Al-Rfou, Rami and Skiena, Steven},
  booktitle={Proceedings of the 20th ACM SIGKDD international conference on Knowledge discovery and data mining},
  pages={701--710},
  year={2014}
}

@article{goodfellow2014generative,
  title={Generative adversarial nets},
  author={Goodfellow, Ian and Pouget-Abadie, Jean and Mirza, Mehdi and Xu, Bing and Warde-Farley, David and Ozair, Sherjil and Courville, Aaron and Bengio, Yoshua},
  journal={Advances in neural information processing systems},
  volume={27},
  pages={2672--2680},
  year={2014}
}

@inproceedings{glorot2010understanding,
  title={Understanding the difficulty of training deep feedforward neural networks},
  author={Glorot, Xavier and Bengio, Yoshua},
  booktitle={Proceedings of the thirteenth international conference on artificial intelligence and statistics},
  pages={249--256},
  year={2010}
}

@article{fengvaleriu,
  title={Valeriu Codreanu, SURFsara, Netherlands Ian Foster, UChicago \& ANL, USA Zhao Zhang, TACC, USA},
  author={Feng, Song and Torsten Hoefler, ETH and Li, Switzerland Jessy and Podareanu, Damian and Pu, Qifan and Qiu, Judy and Saletore, Vikram and Smorkalov, Mikhail E and Torres, Jordi}
}

@article{tang2011leveraging,
  title={Leveraging social media networks for classification},
  author={Tang, Lei and Liu, Huan},
  journal={Data Mining and Knowledge Discovery},
  volume={23},
  number={3},
  pages={447--478},
  year={2011},
  publisher={Springer}
}

@article{higgins_beta-vae_2016,
	title = {beta-{VAE}: {Learning} {Basic} {Visual} {Concepts} with a {Constrained} {Variational} {Framework}},
	shorttitle = {beta-{VAE}},
	url = {https://openreview.net/forum?id=Sy2fzU9gl},
	language = {en},
	urldate = {2021-01-02},
	author = {Higgins, Irina and Matthey, Loic and Pal, Arka and Burgess, Christopher and Glorot, Xavier and Botvinick, Matthew and Mohamed, Shakir and Lerchner, Alexander},
	month = nov,
	year = {2016},
}

@article{paszke_pytorch_2019,
	title = {{PyTorch}: {An} {Imperative} {Style}, {High}-{Performance} {Deep} {Learning} {Library}},
	shorttitle = {{PyTorch}},
	url = {http://arxiv.org/abs/1912.01703},
	urldate = {2021-01-02},
	journal = {arXiv:1912.01703 [cs, stat]},
	author = {Paszke, Adam and Gross, Sam and Massa, Francisco and Lerer, Adam and Bradbury, James and Chanan, Gregory and Killeen, Trevor and Lin, Zeming and Gimelshein, Natalia and Antiga, Luca and Desmaison, Alban and Köpf, Andreas and Yang, Edward and DeVito, Zach and Raison, Martin and Tejani, Alykhan and Chilamkurthy, Sasank and Steiner, Benoit and Fang, Lu and Bai, Junjie and Chintala, Soumith},
	month = dec,
	year = {2019},
	note = {arXiv: 1912.01703},
}

@online{nickel_three-way_nodate,
	title = {A {Three}-{Way} {Model} for {Collective} {Learning} on {Multi}-{Relational} {Data}},
	language = {en},
	author = {Nickel, Maximilian and Tresp, Volker and Kriegel, Hans-Peter},
	pages = {8},
}

@article{mills-tettey_dynamic_nodate,
	title = {The {Dynamic} {Hungarian} {Algorithm} for the {Assignment} {Problem} with {Changing} {Costs}},
	language = {en},
	author = {Mills-Tettey, G Ayorkor and Stentz, Anthony and Dias, M Bernardine},
	pages = {19},
	year = {2007},
}

@book{chung1997spectral,
  title={Spectral graph theory},
  author={Chung, Fan RK and Graham, Fan Chung},
  number={92},
  year={1997},
  publisher={American Mathematical Soc.}
}

@article{yong_gradient_2020,
	title = {Gradient {Centralization}: {A} {New} {Optimization} {Technique} for {Deep} {Neural} {Networks}},
	shorttitle = {Gradient {Centralization}},
	url = {http://arxiv.org/abs/2004.01461},
	urldate = {2020-12-24},
	journal = {arXiv:2004.01461 [cs]},
	author = {Yong, Hongwei and Huang, Jianqiang and Hua, Xiansheng and Zhang, Lei},
	month = apr,
	year = {2020},
	note = {arXiv: 2004.01461},
}

@article{paulheim_knowledge_2016,
	title = {Knowledge graph refinement: {A} survey of approaches and evaluation methods},
	volume = {8},
	issn = {22104968, 15700844},
	shorttitle = {Knowledge graph refinement},
	url = {https://www.medra.org/servlet/aliasResolver?alias=iospress&doi=10.3233/SW-160218},
	doi = {10.3233/SW-160218},
	language = {en},
	number = {3},
	urldate = {2020-12-24},
	journal = {Semantic Web},
	author = {Paulheim, Heiko},
	editor = {Cimiano, Philipp},
	month = dec,
	year = {2016},
	pages = {489--508},
}

@article{zhang_lookahead_2019,
	title = {Lookahead {Optimizer}: k steps forward, 1 step back},
	shorttitle = {Lookahead {Optimizer}},
	url = {http://arxiv.org/abs/1907.08610},
	urldate = {2020-12-24},
	journal = {arXiv:1907.08610 [cs, stat]},
	author = {Zhang, Michael R. and Lucas, James and Hinton, Geoffrey and Ba, Jimmy},
	month = jul,
	year = {2019},
	note = {arXiv: 1907.08610
version: 1},
}

@article{liu_variance_2020,
	title = {On the {Variance} of the {Adaptive} {Learning} {Rate} and {Beyond}},
	url = {http://arxiv.org/abs/1908.03265},
	urldate = {2020-12-24},
	journal = {arXiv:1908.03265 [cs, stat]},
	author = {Liu, Liyuan and Jiang, Haoming and He, Pengcheng and Chen, Weizhu and Liu, Xiaodong and Gao, Jianfeng and Han, Jiawei},
	month = apr,
	year = {2020},
	note = {arXiv: 1908.03265},
}

@article{nickel_review_2016,
	title = {A {Review} of {Relational} {Machine} {Learning} for {Knowledge} {Graphs}},
	volume = {104},
	issn = {0018-9219, 1558-2256},
	url = {http://arxiv.org/abs/1503.00759},
	doi = {10.1109/JPROC.2015.2483592},
	number = {1},
	urldate = {2020-12-24},
	journal = {Proceedings of the IEEE},
	author = {Nickel, Maximilian and Murphy, Kevin and Tresp, Volker and Gabrilovich, Evgeniy},
	month = jan,
	year = {2016},
	note = {arXiv: 1503.00759},
	pages = {11--33},
}

@inproceedings{toutanova_representing_2015,
	address = {Lisbon, Portugal},
	title = {Representing {Text} for {Joint} {Embedding} of {Text} and {Knowledge} {Bases}},
	url = {https://www.aclweb.org/anthology/D15-1174},
	doi = {10.18653/v1/D15-1174},
	urldate = {2020-10-22},
	booktitle = {Proceedings of the 2015 {Conference} on {Empirical} {Methods} in {Natural} {Language} {Processing}},
	publisher = {Association for Computational Linguistics},
	author = {Toutanova, Kristina and Chen, Danqi and Pantel, Patrick and Poon, Hoifung and Choudhury, Pallavi and Gamon, Michael},
	month = sep,
	year = {2015},
	pages = {1499--1509},
}

@incollection{bordes_translating_2013,
	title = {Translating {Embeddings} for {Modeling} {Multi}-relational {Data}},
	url = {http://papers.nips.cc/paper/5071-translating-embeddings-for-modeling-multi-relational-data.pdf},
	urldate = {2020-10-22},
	booktitle = {Advances in {Neural} {Information} {Processing} {Systems} 26},
	publisher = {Curran Associates, Inc.},
	author = {Bordes, Antoine and Usunier, Nicolas and Garcia-Duran, Alberto and Weston, Jason and Yakhnenko, Oksana},
	editor = {Burges, C. J. C. and Bottou, L. and Welling, M. and Ghahramani, Z. and Weinberger, K. Q.},
	year = {2013},
	pages = {2787--2795},
}

@article{date_gpu-accelerated_2016,
	title = {{GPU}-accelerated {Hungarian} algorithms for the {Linear} {Assignment} {Problem}},
	volume = {57},
	issn = {0167-8191},
	url = {http://www.sciencedirect.com/science/article/pii/S016781911630045X},
	doi = {10.1016/j.parco.2016.05.012},
	language = {en},
	urldate = {2020-09-27},
	journal = {Parallel Computing},
	author = {Date, Ketan and Nagi, Rakesh},
	month = sep,
	year = {2016},
	pages = {52--72},
}

@misc{bishop_pattern_2006,
	title = {Pattern recognition and machine learning},
	url = {https://cds.cern.ch/record/998831},
	language = {en},
	urldate = {2020-09-25},
	journal = {CERN Document Server},
	author = {Bishop, Christopher M.},
	year = {2006},
	note = {ISBN: 9781493938438 9780387310732
Publisher: Springer},
}

@article{gardner_allennlp_2018,
	title = {{AllenNLP}: {A} {Deep} {Semantic} {Natural} {Language} {Processing} {Platform}},
	shorttitle = {{AllenNLP}},
	url = {http://arxiv.org/abs/1803.07640},
	urldate = {2020-09-24},
	journal = {arXiv:1803.07640 [cs]},
	author = {Gardner, Matt and Grus, Joel and Neumann, Mark and Tafjord, Oyvind and Dasigi, Pradeep and Liu, Nelson and Peters, Matthew and Schmitz, Michael and Zettlemoyer, Luke},
	month = may,
	year = {2018},
	note = {arXiv: 1803.07640},
}

@inproceedings{cho_finding_2014,
	title = {Finding {Matches} in a {Haystack}: {A} {Max}-{Pooling} {Strategy} for {Graph} {Matching} in the {Presence} of {Outliers}},
	isbn = {978-1-4799-5118-5},
	shorttitle = {Finding {Matches} in a {Haystack}},
	url = {http://ieeexplore.ieee.org/lpdocs/epic03/wrapper.htm?arnumber=6909665},
	doi = {10.1109/CVPR.2014.268},
	language = {en},
	urldate = {2020-07-14},
	booktitle = {2014 {IEEE} {Conference} on {Computer} {Vision} and {Pattern} {Recognition}},
	publisher = {IEEE},
	author = {Cho, Minsu and Sun, Jian and Duchenne, Olivier and Ponce, Jean},
	month = jun,
	year = {2014},
	pages = {2091--2098},
}

@article{kipf_variational_2016,
	title = {Variational {Graph} {Auto}-{Encoders}},
	url = {http://arxiv.org/abs/1611.07308},
	urldate = {2020-07-06},
	journal = {arXiv:1611.07308 [cs, stat]},
	author = {Kipf, Thomas N. and Welling, Max},
	month = nov,
	year = {2016},
	note = {arXiv: 1611.07308},
}

@article{yang_embedding_2015,
	title = {Embedding {Entities} and {Relations} for {Learning} and {Inference} in {Knowledge} {Bases}},
	url = {http://arxiv.org/abs/1412.6575},
	urldate = {2020-06-18},
	journal = {arXiv:1412.6575 [cs]},
	author = {Yang, Bishan and Yih, Wen-tau and He, Xiaodong and Gao, Jianfeng and Deng, Li},
	month = aug,
	year = {2015},
	note = {arXiv: 1412.6575},
}

@article{battaglia_relational_2018,
	title = {Relational inductive biases, deep learning, and graph networks},
	url = {http://arxiv.org/abs/1806.01261},
	urldate = {2020-06-18},
	journal = {arXiv:1806.01261 [cs, stat]},
	author = {Battaglia, Peter W. and Hamrick, Jessica B. and Bapst, Victor and Sanchez-Gonzalez, Alvaro and Zambaldi, Vinicius and Malinowski, Mateusz and Tacchetti, Andrea and Raposo, David and Santoro, Adam and Faulkner, Ryan and Gulcehre, Caglar and Song, Francis and Ballard, Andrew and Gilmer, Justin and Dahl, George and Vaswani, Ashish and Allen, Kelsey and Nash, Charles and Langston, Victoria and Dyer, Chris and Heess, Nicolas and Wierstra, Daan and Kohli, Pushmeet and Botvinick, Matt and Vinyals, Oriol and Li, Yujia and Pascanu, Razvan},
	month = oct,
	year = {2018},
	note = {arXiv: 1806.01261},
}

@article{kipf_semi-supervised_2017,
	title = {Semi-{Supervised} {Classification} with {Graph} {Convolutional} {Networks}},
	url = {http://arxiv.org/abs/1609.02907},
	language = {en},
	urldate = {2020-06-15},
	journal = {arXiv:1609.02907 [cs, stat]},
	author = {Kipf, Thomas N. and Welling, Max},
	month = feb,
	year = {2017},
	note = {arXiv: 1609.02907},
}

@article{kingma_auto-encoding_2014,
	title = {Auto-{Encoding} {Variational} {Bayes}},
	url = {http://arxiv.org/abs/1312.6114},
	urldate = {2020-06-11},
	journal = {arXiv:1312.6114 [cs, stat]},
	author = {Kingma, Diederik P. and Welling, Max},
	month = may,
	year = {2014},
	note = {arXiv: 1312.6114},
}

@article{gangemi_modeling_2018,
	title = {Modeling {Relational} {Data} with {Graph} {Convolutional} {Networks}},
	urldate = {2020-06-03},
	booktitle = {The {Semantic} {Web}},
	publisher = {Springer International Publishing},
	author = {Schlichtkrull, Michael and Kipf, Thomas N. and Bloem, Peter and van den Berg, Rianne and Titov, Ivan and Welling, Max},
	year = {2018},
	doi = {10.1007/978-3-319-93417-4_38},
	note = {Series Title: Lecture Notes in Computer Science},
	pages = {593--607},
}

@misc{diestel2016graph,
  title={Graph Theory. 5th. Vol. 173. GTM},
  author={Diestel, Reinhard},
  year={2016},
  publisher={Springer}
}

@online{pwcWN,
author = {paperswithcode},
title  = {Link Prediction on WN18RR},
urldate   = {2021-01},
month = jan,
url    = {https://paperswithcode.com/sota/link-prediction-on-wn18rr}
}

@online{pwcFB,
author = {paperswithcode},
title  = {Link Prediction on FB15k-237},
urldate   = {2021-01},
month = jan,
url    = {https://paperswithcode.com/sota/link-prediction-on-fb15k-237}
}

@InProceedings{simonovsky_graphvae_2018,
	author="Simonovsky, Martin
	and Komodakis, Nikos",
	title="GraphVAE: Towards Generation of Small Graphs Using Variational Autoencoders",
	booktitle="Artificial Neural Networks and Machine Learning -- ICANN 2018",
	year="2018",
	publisher="Springer International Publishing",
	address="Cham",
	pages="412--422",
	isbn="978-3-030-01418-6"
}

@inproceedings{cudre2013nosql,
  title={NoSQL databases for RDF: an empirical evaluation},
  author={Cudr{\'e}-Mauroux, Philippe and Enchev, Iliya and Fundatureanu, Sever and Groth, Paul and Haque, Albert and Harth, Andreas and Keppmann, Felix Leif and Miranker, Daniel and Sequeda, Juan F and Wylot, Marcin},
  booktitle={International Semantic Web Conference},
  pages={310--325},
  year={2013},
  organization={Springer}
}

@article{miller1998introduction,
  title={An introduction to the resource description framework},
  author={Miller, Eric},
  journal={Bulletin of the American Society for Information Science and Technology},
  volume={25},
  number={1},
  pages={15--19},
  year={1998},
  publisher={Wiley Online Library}
}

@article{ehrlinger2016towards,
  title={Towards a Definition of Knowledge Graphs.},
  author={Ehrlinger, Lisa and W{\"o}{\ss}, Wolfram},
  journal={SEMANTiCS (Posters, Demos, SuCCESS)},
  volume={48},
  pages={1--4},
  year={2016}
}

@inproceedings{simonovsky2017dynamic,
  title={Dynamic edge-conditioned filters in convolutional neural networks on graphs},
  author={Simonovsky, Martin and Komodakis, Nikos},
  booktitle={Proceedings of the IEEE conference on computer vision and pattern recognition},
  pages={3693--3702},
  year={2017}
}

@inproceedings{ruffinelli_you_2019,
	title = {You {CAN} {Teach} an {Old} {Dog} {New} {Tricks}! {On} {Training} {Knowledge} {Graph} {Embeddings}},
	url = {https://openreview.net/forum?id=BkxSmlBFvr},
	urldate = {2020-05-14},
	author = {Ruffinelli, Daniel and Broscheit, Samuel and Gemulla, Rainer},
	month = sep,
	year = {2019},
}

@inproceedings{xie2016representation,
  title={Representation learning of knowledge graphs with entity descriptions},
  author={Xie, Ruobing and Liu, Zhiyuan and Jia, Jia and Luan, Huanbo and Sun, Maosong},
  booktitle={Proceedings of the AAAI Conference on Artificial Intelligence},
  volume={30},
  number={1},
  year={2016}
}

@article{kipf_contrastive_2020,
	title = {Contrastive {Learning} of {Structured} {World} {Models}},
	url = {http://arxiv.org/abs/1911.12247},
	urldate = {2020-05-12},
	journal = {ICLR 2020},
	author = {Kipf, Thomas and van der Pol, Elise and Welling, Max},
	month = jan,
	year = {2020},
	note = {arXiv: 1911.12247},
}

@article{shu_amortized_2019,
	title = {Amortized {Inference} {Regularization}},
	url = {http://arxiv.org/abs/1805.08913},
	urldate = {2021-01-09},
	journal = {arXiv:1805.08913 [cs, stat]},
	author = {Shu, Rui and Bui, Hung H. and Zhao, Shengjia and Kochenderfer, Mykel J. and Ermon, Stefano},
	month = jan,
	year = {2019},
	note = {arXiv: 1805.08913},
}

@article{chen_variational_2017,
	title = {Variational {Lossy} {Autoencoder}},
	url = {http://arxiv.org/abs/1611.02731},
	urldate = {2021-01-09},
	journal = {arXiv:1611.02731 [cs, stat]},
	author = {Chen, Xi and Kingma, Diederik P. and Salimans, Tim and Duan, Yan and Dhariwal, Prafulla and Schulman, John and Sutskever, Ilya and Abbeel, Pieter},
	month = mar,
	year = {2017},
	note = {arXiv: 1611.02731},
}

@article{tolstikhin_wasserstein_2019,
	title = {Wasserstein {Auto}-{Encoders}},
	url = {http://arxiv.org/abs/1711.01558},
	urldate = {2021-01-09},
	journal = {arXiv:1711.01558 [cs, stat]},
	author = {Tolstikhin, Ilya and Bousquet, Olivier and Gelly, Sylvain and Schoelkopf, Bernhard},
	month = dec,
	year = {2019},
	note = {arXiv: 1711.01558},
}

@article{dieng_avoiding_2019,
	title = {Avoiding {Latent} {Variable} {Collapse} {With} {Generative} {Skip} {Models}},
	url = {http://arxiv.org/abs/1807.04863},
	urldate = {2021-01-09},
	journal = {arXiv:1807.04863 [cs, stat]},
	author = {Dieng, Adji B. and Kim, Yoon and Rush, Alexander M. and Blei, David M.},
	month = jan,
	year = {2019},
	note = {arXiv: 1807.04863},
}

@inproceedings{razavi_preventing_2018,
	title = {Preventing {Posterior} {Collapse} with delta-{VAEs}},
	url = {https://openreview.net/forum?id=BJe0Gn0cY7},
	language = {en},
	urldate = {2021-01-10},
	author = {Razavi, Ali and Oord, Aaron van den and Poole, Ben and Vinyals, Oriol},
	month = sep,
	year = {2018},
}

@article{silver_mastering_2017,
	title = {Mastering the game of {Go} without human knowledge},
	volume = {550},
	copyright = {2017 Macmillan Publishers Limited, part of Springer Nature. All rights reserved.},
	issn = {1476-4687},
	url = {https://www.nature.com/articles/nature24270},
	doi = {10.1038/nature24270},
	language = {en},
	number = {7676},
	urldate = {2021-01-11},
	journal = {Nature},
	author = {Silver, David and Schrittwieser, Julian and Simonyan, Karen and Antonoglou, Ioannis and Huang, Aja and Guez, Arthur and Hubert, Thomas and Baker, Lucas and Lai, Matthew and Bolton, Adrian and Chen, Yutian and Lillicrap, Timothy and Hui, Fan and Sifre, Laurent and van den Driessche, George and Graepel, Thore and Hassabis, Demis},
	month = oct,
	year = {2017},
	note = {Number: 7676
Publisher: Nature Publishing Group},
	pages = {354--359},
}

@online{noauthor_elon_nodate,
  title = {Elon Musk on Twitter},
  author = {Musk, Elon},
  year = {2017},
  month = aug,
  url = {https://twitter.com/elonmusk/status/896166762361704450},
  urldate = {2021-01-11}
}

@inproceedings{bollacker_freebase_2008,
	address = {New York, NY, USA},
	series = {{SIGMOD} '08},
	title = {Freebase: a collaboratively created graph database for structuring human knowledge},
	isbn = {978-1-60558-102-6},
	shorttitle = {Freebase},
	url = {https://doi.org/10.1145/1376616.1376746},
	doi = {10.1145/1376616.1376746},
	urldate = {2021-01-12},
	booktitle = {Proceedings of the 2008 {ACM} {SIGMOD} international conference on {Management} of data},
	publisher = {Association for Computing Machinery},
	author = {Bollacker, Kurt and Evans, Colin and Paritosh, Praveen and Sturge, Tim and Taylor, Jamie},
	month = jun,
	year = {2008},
	pages = {1247--1250},
}

@article{dettmers_convolutional_2018,
	title = {Convolutional {2D} {Knowledge} {Graph} {Embeddings}},
	url = {http://arxiv.org/abs/1707.01476},
	urldate = {2021-01-12},
	journal = {arXiv:1707.01476 [cs]},
	author = {Dettmers, Tim and Minervini, Pasquale and Stenetorp, Pontus and Riedel, Sebastian},
	month = jul,
	year = {2018},
	note = {arXiv: 1707.01476},
}
\newpage
\renewcommand{\appendixname}{Annex}
\renewcommand{\appendixtocname}{Annex}
\renewcommand{\appendixpagename}{Annex}

\begin{appendices}
    \section{Parameter distributions}
\label{annexC}

\begin{figure}[H]
\centering
\begin{subfigure}{\textwidth}
    \includegraphics[width=\textwidth]{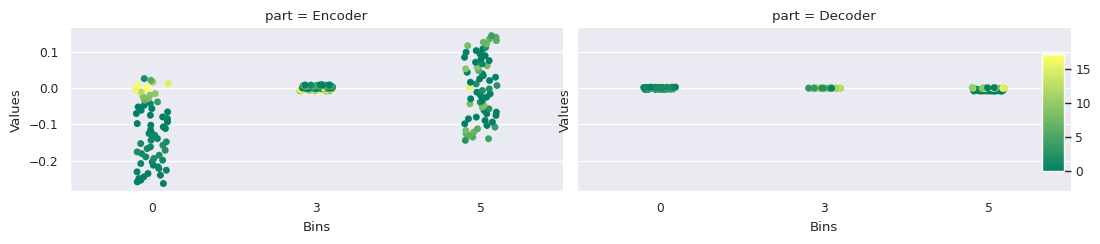}
    \caption{Weight}
    \label{annexC:deltaParamsW}
\end{subfigure}
\begin{subfigure}{\textwidth}
    \includegraphics[width=\textwidth]{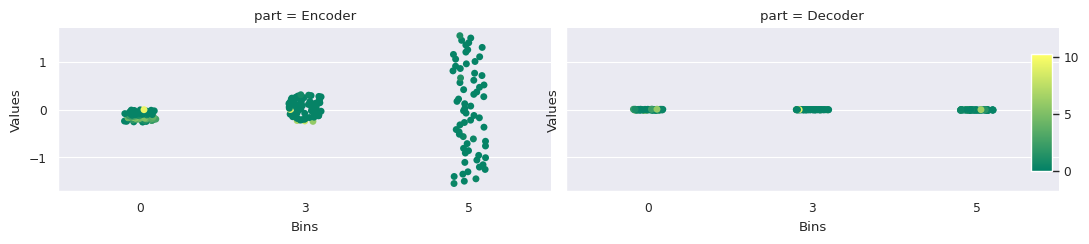}
    \caption{Bias}
    \label{annexC:deltaParamsB}
\end{subfigure}
\caption{Parameter values per layer of the RGVAE with standard loss and $\delta=0.6$.}
\label{annexC:deltaParamsP0}
\end{figure}

\begin{figure}[H]
\centering
\begin{subfigure}{\textwidth}
    \includegraphics[width=\textwidth]{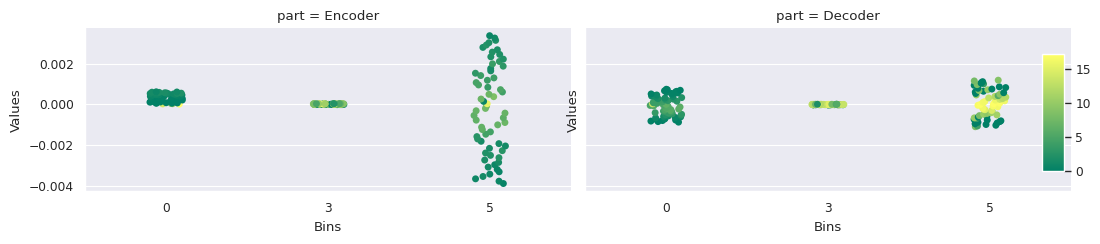}
    \caption{Weight}
    \label{annexC:normParamsW}
\end{subfigure}
\begin{subfigure}{\textwidth}
    \includegraphics[width=\textwidth]{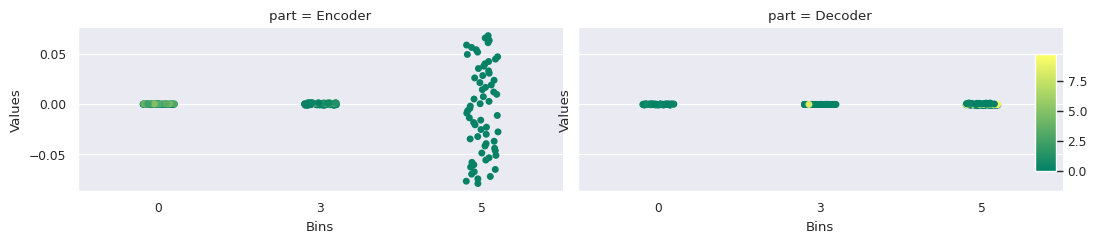}
    \caption{Bias}
    \label{annexC:normParamsB}
\end{subfigure}
\caption{Parameter values per layer of the RGVAE with standard loss and $\delta=0$.}
\label{annexC:normParamsP0}
\end{figure}

    \newpage
    \section{Interpolation}

\subsection{Interpolation between two}

\begin{table}[H]
    \centering
    \begin{tabular}{|c|}
    \hline
    \rowcolor[HTML]{EFEFEF} 
    \textsc{RGVAE standard}\\ \hline
    \texttt{[[Jennifer Jason Leigh] [/film/film/release\_date\_s] [William Wyler]]}\\
    \texttt{[[Keyboard] [/award/award\_category/nominees] [Jessica Biel]]}\\
    \texttt{[[Die Another Day] [/language/human\_language/countries\_spoken\_in] [Stockholm University]]}\\
    \texttt{[[Mission: Impossible II] [/education/educational\_institution] [Intel Corporation]]}\\
    \texttt{[[Techno] [/organization/role/leaders] [Makeup Artist-GB]]}\\
    \texttt{[[The Pianist] [/soccer/football\_team/current\_roster] [Female]]}\\
    \texttt{[[Paul Winchell] [/award/award\_nominee] [Buddhism]]}\\
    \texttt{[[California] [/business/business\_operation/industry] [Marriage]]}\\
    \texttt{[[PFC Beroe Stara Zagora] [/people/person/profession] [National Society of Film Critics Award]]}\\
    \texttt{[[Knight and Day] [/award/award\_category/winners\_winner] [Fox Searchlight Pictures]]}\\    
    \hline
    \end{tabular}
\caption{Latent space interpolation between two triples for RGVAE with standard loss.}
\label{annexA:ipbtw2NoPerm}
\end{table}

\begin{table}[H]
    \centering
    \begin{tabular}{|c|}
    \hline
    \rowcolor[HTML]{EFEFEF} 
    \textsc{$\delta$-RGVAE standard}\\ \hline
    \texttt{[[Willem Dafoe] [/music/instrument/instrumentalists] [Peritonitis]]}\\
    \texttt{[[Tuscaloosa] [/influence/influence\_node/influenced\_by] [Alex Rodriguez]]}\\
    \texttt{[[BAFTA Award for Best Sound] [/people/person/gender] [Fat Possum Records]]}\\
    \texttt{[["Conan OBrien"] [/film/film/country] [World]]}\\
    \texttt{[["They Shoot Horses] [Dont They?"] [/base/popstra/location/vacationers] [Buzz Aldrin]]}\\
    \texttt{[[The Lady] [/award/award\_winning\_work/awards\_won] [Faith Hill]]}\\
    \texttt{[[Laura] [/film/film/country] [Talking Heads]]}\\
    \texttt{[[Peter Morgan] [/film/film/release\_date\_s] [John Amos]]}\\
    \texttt{[[The Butterfly Effect] [/award/award\_nominee] [This Must Be the Place]]}\\
    \texttt{[[Aliens] [/award/award\_winner/awards\_won] [Isle of Man]]}\\
    \hline
    \end{tabular}
\caption{RGVAE latent space interpolation between two triples with $\delta = 0.6$ and standard loss.}
\label{annexA:ipbtw2DeltaNoPerm}
\end{table}

\subsection{Interpolation per latent dimension}
\label{annexB:95}

\begin{longtable}{|c|}
    % \centering
    \hline
    \rowcolor[HTML]{EFEFEF} 
    \textsc{RGVAE permute}\\ \hline
    \rowcolor[HTML]{EFEFEF} 
    \textsc{$i_{d_z}=1$}\\ \hline
    \texttt{[[Estudiantes Tecos] [/film/film/costume\_design\_by] [Sylvester Stallone]]}\\
    \texttt{[[The Man with the Golden Gun] [/film/film/costume\_design\_by] [Bill Fagerbakke]]}\\
    \texttt{[[BAFTA Award for Best Film Music] [/film/film/costume\_design\_by] [Writer-GB]]}\\
    \texttt{[[The Shipping News] [/film/film/costume\_design\_by] [Teri Garr]]}\\
    \texttt{[[Robert De Niro] [/film/film/costume\_design\_by] [Cesar Romero]]}\\
    \texttt{[[Inception] [/film/film/costume\_design\_by] [Hertha BSC Berlin]]}\\
    \texttt{[[Walter F. Parkes] [/film/film/costume\_design\_by] [Mary Lynn Rajskub]]}\\
    \texttt{[[Lorraine Bracco] [/film/film/costume\_design\_by] [Electronic keyboard]]}\\
    \texttt{[[Ben Folds] [/film/film/costume\_design\_by] [1956 Summer Olympics]]}\\
    \texttt{[[Eurasia] [/film/film/costume\_design\_by] [Video]]}\\ \hline
    \rowcolor[HTML]{EFEFEF} 
    \textsc{$i_{d_z}=2$}\\ \hline
    \texttt{[[Piano] [/film/film/costume\_design\_by] [Janes Addiction]]}\\
    \texttt{[[Razzie Award for Worst Director] [/film/film/costume\_design\_by] [Visual Effects Director]]}\\
    \texttt{[[Gone Baby Gone] [/film/film/costume\_design\_by] [Josh Sussman]]}\\
    \texttt{[[Ellen Pompeo] [/film/film/costume\_design\_by] [Joss Whedon]]}\\
    \texttt{[[Bachelor of Arts] [/film/film/costume\_design\_by] [Chicago]]}\\
    \texttt{[[News Corporation] [/film/film/costume\_design\_by] [2012 Summer Olympics]]}\\
    \texttt{[[Ukulele] [/film/film/costume\_design\_by] [85th Academy Awards]]}\\
    \texttt{[[Columbia Pictures] [/film/film/costume\_design\_by] [The Last Picture Show]]}\\
    \texttt{[[Catherine Keener] [/film/film/costume\_design\_by] [Gladys Knight]]}\\
    \texttt{[[Amelia] [/film/film/costume\_design\_by] [Hikaru Utada]]}\\ \hline
    \rowcolor[HTML]{EFEFEF} 
    \textsc{$i_{d_z}=3$}\\ \hline
    \texttt{[[Steve Carell] [/film/film/costume\_design\_by] [Australia]]}\\
    \texttt{[[About a Boy] [/film/film/costume\_design\_by] [United States Dollar]]}\\
    \texttt{[[Academy Award for Best Cinematography] [/film/film/costume\_design\_by] [Defender]]}\\
    \texttt{[[Jayma Mays] [/film/film/costume\_design\_by] [Broadcast Film Critics Association Award]]}\\
    \texttt{[[Arthur C. Clarke] [/film/film/costume\_design\_by] [Comedy-GB]]}\\
    \texttt{[[1968 Summer Olympics] [/film/film/costume\_design\_by] [Quarterback]]}\\
    \texttt{[[Tim Rice] [/film/film/costume\_design\_by] [Electric guitar]]}\\
    \texttt{[[The Faculty] [/film/film/costume\_design\_by] [Linebacker]]}\\
    \texttt{[[Academy Award for Best Screenplay] [/film/film/costume\_design\_by] [Singer-songwriter-GB]]}\\
    \texttt{[[Raoul Walsh] [/film/film/costume\_design\_by] [Julianna Margulies]]}\\ \hline
    \rowcolor[HTML]{EFEFEF} 
    \textsc{$i_{d_z}=4$}\\ \hline
    \texttt{[[Trumpet] [/film/film/costume\_design\_by] [Argentina]]}\\
    \texttt{[[Costa Mesa] [/film/film/costume\_design\_by] [51st Annual Grammy Awards-US]]}\\
    \texttt{[[Amanda Lear] [/education/educational\_degree/people\_with\_this\_degree] [New Jersey]]}\\
    \texttt{[[Jean-Paul Sartre] [/film/film/costume\_design\_by] [Hong Kong]]}\\
    \texttt{[[Charles LeMaire] [/film/film/costume\_design\_by] [United States of America]]}\\
    \texttt{[[13 Assassins] [/film/film/costume\_design\_by] [V for Vendetta]]}\\
    \texttt{[[Academy Award for Best Original Song] [/film/film/costume\_design\_by] [Million Dollar Baby]]}\\
    \texttt{[[Alan Parsons] [/film/film/costume\_design\_by] [Mark Kirkland]]}\\
    \texttt{[[Africa] [/film/film/costume\_design\_by] [General-GB]]}\\
    \texttt{[[Joanna Lumley] [/film/film/costume\_design\_by] [Melanie Griffith]]}\\ \hline
    \rowcolor[HTML]{EFEFEF} 
    \textsc{$i_{d_z}=5$}\\ \hline
    \texttt{[[Brighton Rock] [/film/film/costume\_design\_by] [Brown University]]}\\
    \texttt{[[Claude Rains] [/film/film/costume\_design\_by] [Dhaka]]}\\
    \texttt{[[Drums] [/film/film/costume\_design\_by] [Marty Stuart]]}\\
    \texttt{[[Alameda County] [/film/film/costume\_design\_by] [Striptease]]}\\
    \texttt{[[Bachelor of Arts] [/film/film/costume\_design\_by] [G-Force]]}\\
    \texttt{[[Owen Wilson] [/film/film/costume\_design\_by] [University of London]]}\\
    \texttt{[[Being There] [/film/film/costume\_design\_by] [14th Screen Actors Guild Awards]]}\\
    \texttt{[[Tim McGraw] [/film/film/costume\_design\_by] [Japanese Language]]}\\
    \texttt{[[Brandenburg] [/film/film/costume\_design\_by] [Honolulu]]}\\
    \texttt{[[Rekha] [/film/film/costume\_design\_by] [Defender]]}\\ \hline
    \rowcolor[HTML]{EFEFEF} 
    \textsc{$i_{d_z}=6$}\\ \hline
    \texttt{[[Catherine Deneuve] [/film/film/costume\_design\_by] [Robert Ridgely]]}\\
    \texttt{[[Felix Bastians] [/film/film/costume\_design\_by] [English Language]]}\\
    \texttt{[[Denzel Washington] [/film/film/costume\_design\_by] [German Democratic Republic]]}\\
    \texttt{[[Gina Gershon] [/film/film/costume\_design\_by] [Shanghai Greenland F.C.]]}\\
    \texttt{[[Peter Cook] [/film/film/costume\_design\_by] [Carrie Fisher]]}\\
    \texttt{[[Jim Taylor] [/film/film/costume\_design\_by] [Romantic comedy]]}\\
    \texttt{[[Guitar] [/film/film/costume\_design\_by] [Private university]]}\\
    \texttt{[[Melodica] [/film/film/costume\_design\_by] [Screenwriter]]}\\
    \texttt{[[Broadcast Film Critics Association Award] [/film/film/costume\_design\_by] [Running back]]}\\
    \texttt{[[Southern Methodist University] [/film/film/costume\_design\_by] [Mellotron]]}\\ \hline
    \rowcolor[HTML]{EFEFEF} 
    \textsc{$i_{d_z}=7$}\\ \hline
    \texttt{[[Grammy Award for Best R\&B Performance] [/film/film/costume\_design\_by] [Tunisia]]}\\
    \texttt{[[Comedy-GB] [/film/film/costume\_design\_by] [Louisiana State University]]}\\
    \texttt{[[Yes Man] [/film/film/costume\_design\_by] [Brooklyn]]}\\
    \texttt{[[Justin Chambers] [/film/film/costume\_design\_by] [David Petraeus]]}\\
    \texttt{[[Akiva Schaffer] [/film/film/costume\_design\_by] [United States Dollar]]}\\
    \texttt{[[Masovian Voivodeship] [/film/film/costume\_design\_by] [Germany]]}\\
    \texttt{[[Kevin Betsy] [/film/film/costume\_design\_by] ["One Flew Over the Cuckoos Nest"]]}\\
    \texttt{[[Tombstone] [/film/film/costume\_design\_by] [Drum]]}\\
    \texttt{[[Piranha DD] [/film/film/costume\_design\_by] [United States of America]]}\\
    \texttt{[[Sire Records] [/film/film/costume\_design\_by] [Primetime Emmy Award]]}\\ \hline
    \textsc{$i_{d_z}=8$}\\ \hline
    \texttt{[[Coconut milk] [/film/film/costume\_design\_by] [Goa]]}\\
    \texttt{[[Academy Award for Best Picture] [/base/eating/practicer\_of\_diet/diet] [A Beautiful Mind]]}\\
    \texttt{[[Indiana Jones and the Temple of Doom] [/film/film/costume\_design\_by] [Romania]]}\\
    \texttt{[[Charterhouse School] [/film/film/costume\_design\_by] [Marilyn Manson]]}\\
    \texttt{[[Verizon Communications] [/film/film/costume\_design\_by] [Japan]]}\\
    \texttt{[[Mulan] [/film/film/costume\_design\_by] [Sony BMG Music Entertainment]]}\\
    \texttt{[[Matthew McConaughey] [/film/film/costume\_design\_by] [Chocolat]]}\\
    \texttt{[[Northwestern University] [/film/film/costume\_design\_by] [United States of America]]}\\
    \texttt{[[Survivor] [/film/film/costume\_design\_by] [Malibu]]}\\
    \texttt{[[L.A. Confidential] [/film/film/costume\_design\_by] [Sarah Polley]]}\\ \hline
    \rowcolor[HTML]{EFEFEF} 
    \textsc{$i_{d_z}=9$}\\ \hline
    \texttt{[[Scott Pilgrim vs. the World] [/film/film/costume\_design\_by] [Vegetarianism]]}\\
    \texttt{[[Pina] [/film/film/costume\_design\_by] [Synthesizer]]}\\
    \texttt{[[Ryerson University] [/film/film/costume\_design\_by] [Oldham Athletic A.F.C.]]}\\
    \texttt{[[Tony Geiss] [/film/film/costume\_design\_by] [Maine]]}\\
    \texttt{[[Get Smart] [/film/film/costume\_design\_by] [American Gangster]]}\\
    \texttt{[[59th Golden Globe Awards] [/film/film/costume\_design\_by] [Marriage]]}\\
    \texttt{[[Coimbatore district] [/film/film/costume\_design\_by] [Midfielder]]}\\
    \texttt{[[90th United States Congress] [/film/film/costume\_design\_by] [Male]]}\\
    \texttt{[[Colin Firth] [/film/film/costume\_design\_by] [Israel]]}\\
    \texttt{[[Brian Baumgartner] [/film/film/costume\_design\_by] [American Beauty]]}\\ \hline
    \rowcolor[HTML]{EFEFEF} 
    \textsc{$i_{d_z}=10$}\\ \hline
    \texttt{[[Charles Laughton] [/film/film/costume\_design\_by] [Bosnia and Herzegovina]]}\\
    \texttt{[[Academy Award for Best Design] [/film/film/costume\_design\_by] [Oxford United F.C.]]}\\
    \texttt{[[Winona Ryder] [/location/country/official\_language] [Singapore]]}\\
    \texttt{[[Grammy Award for Best Album for Children] [/film/film/costume\_design\_by] [Blue-eyed soul]]}\\
    \texttt{[[Shooting guard] [/film/film/costume\_design\_by] [National Board of Review Award]]}\\
    \texttt{[[Borat] [/film/film/costume\_design\_by] [The French Connection]]}\\
    \texttt{[[Oh! What a Lovely War] [/film/film/costume\_design\_by] [Kareena Kapoor]]}\\
    \texttt{[[Liv Tyler] [/film/film/costume\_design\_by] [Jamie Lynn Sigler]]}\\
    \texttt{[[John Wayne] [/film/film/costume\_design\_by] [United States Dollar]]}\\
    \texttt{[[Speed Racer] [/film/film/costume\_design\_by] [Grammy Award for Song of the Year]]}\\
    \hline
\caption{Interpolation of each latent dimension for the RGVAE with graph matching.}
\label{annexA:ipdim95}
\end{longtable}

\begin{longtable}{|c|}
    % \centering
    \hline
    \rowcolor[HTML]{EFEFEF} 
    \textsc{RGVAE standard}\\ \hline
    \rowcolor[HTML]{EFEFEF} 
    \textsc{$i_{d_z}=1$}\\ \hline
    \texttt{[[1976 Winter Olympics] [/base/marchmadness/ncaa\_basketball\_tournament/seeds] [Iain Glen]]}\\
    \texttt{[[Judy Greer] [/film/actor/film] [Arkansas]]}\\
    \texttt{[[Vincent Pastore] [/film/film/genre] [Vice President-GB]]}\\
    \texttt{[["Ulysses Gaze"] [/people/person/gender] [ESPN]]}\\
    \texttt{[[Manchester University] [/award/ranked\_item/appears\_in\_ranked\_lists] [Silver medal]]}\\
    \texttt{[[Wesleyan University] [/business/job\_title/people\_with\_this\_title] [Lisa Gay Hamilton]]}\\
    \texttt{[[Iraq] [/music/instrument/family] [Greg Daniels]]}\\
    \texttt{[[Empire of the Sun] [/people/person/profession] [Catherine Keener]]}\\
    \texttt{[[Leslie Bricusse] [/award/award\_nominee] [Jane Fonda]]}\\
    \texttt{[[Forward] [/film/film\_distributor/films\_distributed] [Sandra Bullock]]}\\
    \hline
    \rowcolor[HTML]{EFEFEF} 
    \textsc{$i_{d_z}=2$}\\ \hline
    \texttt{[[Saxophone] [/award/award\_winning\_work/awards\_won] [The Prince of Egypt]]}\\
    \texttt{[[Rang De Basanti] [/location/location/contains] [1992 Summer Olympics]]}\\
    \texttt{[[Small Soldiers] [/award/award\_nominee] [New York City]]}\\
    \texttt{[[Bill Hicks] [/award/award\_nominee] [Paul Williams]]}\\
    \texttt{[[Hairspray] [/user/jg/default\_domain/olympic\_games/sports] [Michael K. Williams]]}\\
    \texttt{[[The Aviator] [/film/actor/film] [Goalkeeper]]}\\
    \texttt{[[Joel Grey] [/award/award\_nominee] [Vladimir Vladimirovich Nabokov]]}\\
    \texttt{[[Vibraphone] [/film/film\_distributor/films\_distributed] ["Jews harp"]]}\\
    \texttt{[[Ice Age: Dawn of the Dinosaurs] [/music/group\_member/membership] [Club Atlas]]}\\
    \texttt{[[Doctor Zhivago] [/film/film/release\_date\_s] [Terror in the Aisles]]}\\
    \hline
    \rowcolor[HTML]{EFEFEF} 
    \textsc{$i_{d_z}=3$}\\ \hline
    \texttt{[[Real Sociedad] [/common/topic/webpage./common/webpage/category] [Palmitic acid]]}\\
    \texttt{[[Soul music] [/people/person/profession] [Alameda]]}\\
    \texttt{[[Viggo Mortensen] [/medicine/disease/risk\_factors] [Love Story]]}\\
    \texttt{[[Finland] [/sports/sports\_position/players] [DVD]]}\\
    \texttt{[[Crazy Heart] [/tv/tv\_producer/programs\_produced] [Los Angeles]]}\\
    \texttt{[[Drew Barrymore] [/tv/tv\_writer/tv\_programs] [Road cycling]]}\\
    \texttt{[[BAFTA Award for Best Screenplay] [/award/award\_nominee] [John Ortiz]]}\\
    \texttt{[[Mumford] [/film/film/release\_date\_s] [Howard Gordon]]}\\
    \texttt{[[Grammy Award for Best Musical Album] [/people/person/places\_lived] [Batman Begins]]}\\
    \texttt{[[Cats \& Dogs] [/sports/sports\_team/colors] [Musician-GB]]}\\
    \hline
    \rowcolor[HTML]{EFEFEF} 
    \textsc{$i_{d_z}=4$}\\ \hline
    \texttt{[[Saint Marys College] [/government/political\_party/politicians\_in\_this\_party] [Joel Coen]]}\\
    \texttt{[[Eva Longoria] [/people/person/profession] [The Lives of Others]]}\\
    \texttt{[[United States Congress] [/award/award\_nominee] [Tony Award for Best Actor]]}\\
    \texttt{[[Midfielder] [/music/instrument/instrumentalists] [Elizabeth]]}\\
    \texttt{[[Brad Pitt] [/film/actor/film] [South Sudan]]}\\
    \texttt{[[Django Unchained] [/user/ktrueman/default\_domain/] [France]]}\\
    \texttt{[[McMaster University] [/music/artist/track\_contributions] [Philippines]]}\\
    \texttt{[[The Stand] [/people/person/profession] [Curly Howard]]}\\
    \texttt{[[Jerome Flynn] [/award/award\_nominee] [Piano]]}\\
    \texttt{[[Brighton Rock] [/location/location/contains] [Forward]]}\\
    \hline
    \rowcolor[HTML]{EFEFEF} 
    \textsc{$i_{d_z}=5$}\\ \hline
    \texttt{[[Jennifer Aniston] [/people/person/place\_of\_birth] [Peru]]}\\
    \texttt{[[Timothy Busfield] [/film/film/release\_date\_s] [Defender]]}\\
    \texttt{[[Nirupa Roy] [/food/food/nutrients] [Wake Forest University]]}\\
    \texttt{[[Kidderminster F.C.] [/award/award\_winning\_work/awards\_won\_winner] [United States Dollar]]}\\
    \texttt{[[BAFTA Award for Best Actress] [/location/statistical\_region] [Forest Lawn Memorial Park]]}\\
    \texttt{[[Babylon 5] [/award/award\_nominee\_nominee] [Hong Kong]]}\\
    \texttt{[[Johann Wolfgang von Goethe] [/film/actor/film] [Sunday Bloody Sunday-GB]]}\\
    \texttt{[[St. Augustine] [/people/marriage\_union\_type/unions\_of\_this\_type] [Columbia Pictures]]}\\
    \texttt{[[Brokeback Mountain] [/award/award\_nominee] [Oscar]]}\\
    \texttt{[[Shanghai Knights] [/film/film/cinematography] [United States Dollar]]}\\
    \hline
    \rowcolor[HTML]{EFEFEF} 
    \textsc{$i_{d_z}=6$}\\ \hline
    \texttt{[[Cape Fear] [/award/award\_nominee] [DVD]]}\\
    \texttt{[[Busta Rhymes] [/film/film/language] [United States Dollar]]}\\
    \texttt{[[The Bourne Identity] [/people/person/profession] [Ally Sheedy]]}\\
    \texttt{[[Taxi Driver] [/people/person/profession] [Ipswich Town F.C.]]}\\
    \texttt{[[France] [/film/actor/film] [Greensboro]]}\\
    \texttt{[[The Constant Gardener] [/award/award\_nominee] [Academy Award for Best Screenplay]]}\\
    \texttt{[[North Atlantic Treaty Organization (NATO)] [/people/person/profession] [Makeup Artist-GB]]}\\
    \texttt{[[Trey Parker] [/time/event/locations] [University of the West Indies]]}\\
    \texttt{[[Mockumentary] [/people/deceased\_person/place\_of\_death] [Newport Beach]]}\\
    \texttt{[[Phillips Academy] [/music/performance\_role/regular\_performances] [Police procedural]]}\\
    \hline
    \rowcolor[HTML]{EFEFEF} 
    \textsc{$i_{d_z}=7$}\\ \hline
    \texttt{[[Gladiator] [/people/person/profession] [Clarinet]]}\\
    \texttt{[[Jenna Fischer] [/sports/sports\_team\_location/teams] [Spain]]}\\
    \texttt{[[Israel] [/base/localfood/seasonal\_month/produce\_available] [Yale University]]}\\
    \texttt{[[Tompkins County] [/organization/organization/place\_founded] [William Shakespeare]]}\\
    \texttt{[[Bas-Rhin] [/film/film/release\_date\_s] [Actor-GB]]}\\
    \texttt{[[Virgin Records] [/location/hud\_foreclosure\_area] ["Peoples Action Party"]]}\\
    \texttt{[[Farida Jalal] [/education/educational\_institution/school\_type] [Primetime Emmy Award]]}\\
    \texttt{[[Piano] [/award/award\_winning\_work/awards\_won] [Marty Stuart]]}\\
    \texttt{[[John C. Reilly] [/location/location/contains] [Jim Broadbent]]}\\
    \texttt{[[Tim Roth] [/film/film/language] [Egypt]]}\\
    \hline
    \rowcolor[HTML]{EFEFEF} 
    \textsc{$i_{d_z}=8$}\\ \hline
    \texttt{[[Ennio Morricone] [/award/award\_winner/awards\_won\_winner] [98th United States Congress]]}\\
    \texttt{[[Mali Finn] [/base/marchmadness/ncaa\_basketball\_tournament/seeds] [Angie Dickinson]]}\\
    \texttt{[[David Walliams] [/music/performance\_role/regular\_performances] [Toy Story 3]]}\\
    \texttt{[[Marriage] [/award/award\_ceremony/awards\_presented] [New York City]]}\\
    \texttt{[[Philip Yordan] [/soccer/football\_team/current\_roster] [English Language]]}\\
    \texttt{[[63rd Primetime Emmy Awards] [/film/film/other\_crew] [Stephen Frears]]}\\
    \texttt{[[Hud] [/film/film/other\_crew] [Potassium]]}\\
    \texttt{[[Arkansas] [/film/film/featured\_film\_locations] [Special Effects Supervisor]]}\\
    \texttt{[[Fran Drescher] [/film/film\_distributor/films\_distributed] [Giant]]}\\
    \texttt{[[Charlotte Gainsbourg] [/travel/travel\_destination/climate] [United Kingdom]]}\\
    \hline
    \rowcolor[HTML]{EFEFEF} 
    \textsc{$i_{d_z}=9$}\\ \hline
    \texttt{[[Bachelor of Business Administration] [/film/director/film] [David Foster]]}\\
    \texttt{[[Andy Samberg] [/people/person/place\_of\_birth] [Portugal]]}\\
    \texttt{[[Academy Award for Best Picture] [/film/film/language] [Jessica Biel]]}\\
    \texttt{[[The Last Emperor] [/education/educational\_institution/school\_type] [Closer]]}\\
    \texttt{[[Hewlett-Packard] [/award/award\_winner/awards\_won\_winner] [Russia]]}\\
    \texttt{[[Joan Allen] [/tv/tv\_producer/programs\_produced] [Daytime Emmy Award for Outstanding Host]]}\\
    \texttt{[[The Royal Tenenbaums] [/olympics/olympic\_sport/athletes] [Mathematics]]}\\
    \texttt{[[EMI] [/people/person/places\_lived] [Jason Isaacs]]}\\
    \texttt{[[DePaul University] [/award/award\_category/nominees] [Martie Maguire]]}\\
    \texttt{[[Managing Director-GB] [/film/actor/film] [Tom Petty and the Heartbreakers]]}\\
    \hline
    \rowcolor[HTML]{EFEFEF} 
    \textsc{$i_{d_z}=10$}\\ \hline
    \texttt{[[Nebraska] [/award/award\_nominee] [2008 Summer Olympics]]}\\
    \texttt{[[Mike Scully] [/location/location/adjoin\_s] [Hal B. Wallis]]}\\
    \texttt{[[Frankenweenie] [/sports/sports\_position/players] [University of Cape Town]]}\\
    \texttt{[[Johnny Green] [/film/film/release\_date\_s] [Bruce Springsteen]]}\\
    \texttt{[[Jamie Lynn Sigler] [/award/award\_nominee\_nominee] [Maggie Smith]]}\\
    \texttt{[[Academy Award for Best Actor] [/base/popstra/celebrity/friendship] [Engineer-GB]]}\\
    \texttt{[[Hope Davis] [/film/film/other\_crew] [London]]}\\
    \texttt{[[Scotland] [/award/award\_nominee\_nominee] [1920 Summer Olympics]]}\\
    \texttt{[[Keanu Reeves] [/music/record\_label/artist] [German Language]]}\\
    \texttt{[[The Perks of Being a Wallflower] [/olympics/olympic\_participating\_country] [United States Dollar]]}\\
    \hline
\caption{Interpolation of each latent dimension for the RGVAE with standard loss function.}
\label{annexA:ipdim95NoPerm}
\end{longtable}

\begin{longtable}{|c|}
    % \centering
    \hline
    \rowcolor[HTML]{EFEFEF} 
    \textsc{$\delta$-RGVAE permute}\\ \hline
    \hline 
    \rowcolor[HTML]{EFEFEF} 
    \textsc{$i_{d_z}=1$}\\ \hline 
    \texttt{[[Amman] [/user/tsegaran/random/taxonomy\_subject/entry] [Bill Cosby]]}\\
    \texttt{[[Kristen Stewart] [/people/person/spouse\_s] [United States of America]]}\\
    \texttt{[[Dean Koontz] [/user/tsegaran/random/taxonomy\_subject/entry] [Fukushima Prefecture]]}\\
    \texttt{[[Western Front] [/user/tsegaran/random/taxonomy\_subject/entry] [Member of Congress]]}\\
    \texttt{[[Cyrano de Bergerac] [/user/tsegaran/random/taxonomy\_subject/entry] [Star Wars Episode III]]}\\
    \texttt{[[Country] [/user/tsegaran/random/taxonomy\_subject/entry] [Lois Smith]]}\\
    \texttt{[[Imelda Staunton] [/user/tsegaran/random/taxonomy\_subject/entry] [DVD]]}\\
    \texttt{[[Acid house] [/film/film/other\_crew] [Sienna Miller]]}\\
    \texttt{[[Underground hip hop] [/user/tsegaran/random/taxonomy\_subject/entry] [Placekicker]]}\\
    \texttt{[[Indie pop] [/user/tsegaran/random/taxonomy\_subject/entry] [Big Love]]}\\
    \hline 
    \rowcolor[HTML]{EFEFEF} 
    \textsc{$i_{d_z}=2$}\\ \hline 
    \texttt{[[UTV] [/base/localfood/seasonal\_month/produce\_available] [Drama]]}\\
    \texttt{[[Crewe Alexandra F.C.] [/film/film/other\_crew] [Wheaton]]}\\
    \texttt{[[Richard Burton] [/user/tsegaran/random/taxonomy\_subject/entry] [Pinellas County]]}\\
    \texttt{[[Cave of Forgotten Dreams] [/film/film/release\_date\_s] [The Lord of the Rings]]}\\
    \texttt{[[2008 Summer Olympics] [/time/event/locations] [Chulalongkorn]]}\\
    \texttt{[[Omarion] [/award/award\_category/disciplines\_or\_subjects] [Taito Corporation]]}\\
    \texttt{[[Providence County] [/government/politician/government\_positions\_held] [Drama]]}\\
    \texttt{[[Economist] [/sports/sports\_team/roster] [Tommy Lee Jones]]}\\
    \texttt{[[Independent Spirit Award for Best Director] [/film/film/written\_by] [Defender]]}\\
    \texttt{[[California] [/user/tsegaran/random/taxonomy\_subject/entry] [Catholicism]]}\\
    \hline 
    \rowcolor[HTML]{EFEFEF} 
    \textsc{$i_{d_z}=3$}\\ \hline 
    \texttt{[[Dick Powell] [/film/film/release\_date\_s] [Alex Borstein]]}\\
    \texttt{[[Jeff Richmond] [/film/film/personal\_appearances] [Never Say Never Again]]}\\
    \texttt{[[The Amazing Spider-Man] [/user/tsegaran/random/taxonomy\_subject/entry] [Animaniacs]]}\\
    \texttt{[[Heavy metal] [/user/tsegaran/random/taxonomy\_subject/entry] [Mark Ronson]]}\\
    \texttt{[[Southeast Asia] [/location/hud\_county\_place/county] [Franklin Medal]]}\\
    \texttt{[[Jon Stewart] [/film/film/country] [Warren Ellis]]}\\
    \texttt{[[Marv Wolfman] [/location/administrative\_division] [95th United States Congress]]}\\
    \texttt{[[John Davis] [/sports/sports\_position/players] [Male]]}\\
    \texttt{[[2009 Tour de France] [/award/award\_nominee] [Cate Blanchett]]}\\
    \texttt{[[Real Betis] [/film/film/release\_date\_s] [Obesity]]}\\
    \hline 
    \rowcolor[HTML]{EFEFEF} 
    \textsc{$i_{d_z}=4$}\\ \hline 
    \texttt{[[Monaco] [/user/tsegaran/random/taxonomy\_subject/entry] [Ryan Seacrest]]}\\
    \texttt{[[MADtv] [/organization/endowed\_organization/endowment] [Toby Jones]]}\\
    \texttt{[[Master of Arts] [/time/event/instance\_of\_recurring\_event] [Conglomerate]]}\\
    \texttt{[[Tim Robbins] [/user/tsegaran/random/taxonomy\_subject/entry] [Homicide: Life on the Street]]}\\
    \texttt{[["The Hitchhikers Guide to the Galaxy"] [/film/film/film\_art\_direction\_by] [Lawrence Bender]]}\\
    \texttt{[[Melvyn Douglas] [/food/food/nutrients] [Kingston University]]}\\
    \texttt{[[Billings] [/user/tsegaran/random/taxonomy\_subject/entry] [Jon Peters]]}\\
    \texttt{[[Brandon] [/people/ethnicity/languages\_spoken] [Motorola]]}\\
    \texttt{[[Tom Stoppard] [/location/statistical\_region] [Kid Rock]]}\\
    \texttt{[[FC Khimki] [/user/tsegaran/random/taxonomy\_subject/entry] [Richard Sylbert]]}\\
    \hline 
    \rowcolor[HTML]{EFEFEF} 
    \textsc{$i_{d_z}=5$}\\ \hline 
    \texttt{[[Ron Paul] [/user/tsegaran/random/taxonomy\_subject/entry] [The Shubert Organization]]}\\
    \texttt{[[Merry Christmas] [Mr. Lawrence] [/people/profession/specialization\_of] [United Kingdom]]}\\
    \texttt{[[St. Mirren F.C.] [/user/tsegaran/random/taxonomy\_subject/entry] [United States Dollar]]}\\
    \texttt{[[Joe Jackson] [/sports/sports\_team/sport] [Iran national football team]]}\\
    \texttt{[[Philips] [/user/tsegaran/random/taxonomy\_subject/entry] [47th Annual Grammy Awards]]}\\
    \texttt{[["Paquito DRivera"] [/film/film\_subject/films] [Forward]]}\\
    \texttt{[[University of Maine] [/base/popstra/celebrity/breakup] [Sonic Youth]]}\\
    \texttt{[[Brian Eno] [/influence/influence\_node/peers] [Tim Conway]]}\\
    \texttt{[[A Room with a View] [/user/tsegaran/random/taxonomy\_subject/entry] [Huntington]]}\\
    \texttt{[[Nicholas and Alexandra] [/film/film/distributors] [Lionsgate Entertainment]]}\\
    \hline 
    \rowcolor[HTML]{EFEFEF} 
    \textsc{$i_{d_z}=6$}\\ \hline 
    \texttt{[[Farrah Fawcett] [/education/university/international\_tuition] [The Bourne Legacy]]}\\
    \texttt{[[William Shakespeare] [/base/saturdaynightlive/snl\_cast\_member/seasons] [Drama]]}\\
    \texttt{[[Serj Tankian] [/film/film/costume\_design\_by] [Ted Levine]]}\\
    \texttt{[[Amitabh Bachchan] [/location/hud\_foreclosure\_area] [LA Galaxy]]}\\
    \texttt{[[Taekwondo] [/location/statistical\_region/rent50\_2] [Drum]]}\\
    \texttt{[[Portugal] [/music/group\_member/membership] [Mountain Time Zone-US]]}\\
    \texttt{[[Saxony] [/user/tsegaran/random/taxonomy\_subject/entry] [2003 invasion of Iraq]]}\\
    \texttt{[[Addis Ababa] [/baseball/baseball\_team/team\_stats] [Tony Award for Best Original Score]]}\\
    \texttt{[[Tijuana] [/user/tsegaran/random/taxonomy\_subject/entry] [Bellevue]]}\\
    \texttt{[[Bruce Timm] [/people/person/gender] [Brian Grazer]]}\\
    \hline 
    \rowcolor[HTML]{EFEFEF} 
    \textsc{$i_{d_z}=7$}\\ \hline 
    \texttt{[[Charles Mingus] [/film/film/distributors] [University of Chicago Law School]]}\\
    \texttt{[[Feroz Khan] [/people/person/sibling\_s] [Marvin Hatley]]}\\
    \texttt{[[Jason Segel] [/film/film/dubbing\_performances] [Psychedelic trance]]}\\
    \texttt{[[Swimming Upstream] [/celebrities/celebrity/celebrity\_friends] [Avocado]]}\\
    \texttt{[[Jockey-GB] [/olympics/olympic\_sport/athletes] [Robert Towne]]}\\
    \texttt{[[Bachelor of Arts] [/organization/organization\_member/member\_of] [Executive Producer]]}\\
    \texttt{[[Fairfield County] [/user/tsegaran/random/taxonomy\_subject/entry] [Saeed Jaffrey]]}\\
    \texttt{[[Catherine Keener] [/base/eating/practicer\_of\_diet/diet] [Bicentennial Man]]}\\
    \texttt{[[Dance-pop] [/location/statistical\_region/places\_exported\_to] [No Doubt]]}\\
    \texttt{[[Tina Turner] [/military/military\_conflict/combatants] [Clarence Brown]]}\\
    \hline 
    \rowcolor[HTML]{EFEFEF} 
    \textsc{$i_{d_z}=8$}\\ \hline 
    \texttt{[[Heavy metal] [/film/special\_film\_performance\_type/film\_performance\_type] [Black comedy]]}\\
    \texttt{[[Jackie Cooper] [/user/tsegaran/random/taxonomy\_subject/entry] [Mali]]}\\
    \texttt{[["The Sorcerers Apprentice"] [/user/tsegaran/random/taxonomy\_subject/entry] [Olympia Dukakis]]}\\
    \texttt{[[Wim Wenders] [/education/university/fraternities\_and\_sororities] [University of London]]}\\
    \texttt{[[Punahou School] [/user/tsegaran/random/taxonomy\_subject/entry] [Aldo Nova]]}\\
    \texttt{[[Rory Cochrane] [/government/government\_office\_category/officeholders] [Samantha Morton]]}\\
    \texttt{[[Robert Byrd] [/award/award\_winning\_work/awards\_won] [Romantic fantasy]]}\\
    \texttt{[[Loudoun County] [/baseball/baseball\_team/team\_stats] [Master of Arts]]}\\
    \texttt{[[Ohio] [/user/tsegaran/random/taxonomy\_subject/entry] [Screen Actors Guild Award]]}\\
    \texttt{[[Nas] [/film/film/prequel] [MTV Video Music Award for Best Hip-Hop Video]]}\\
    \hline 
    \rowcolor[HTML]{EFEFEF} 
    \textsc{$i_{d_z}=9$}\\ \hline 
    \texttt{[[Naismith Basketball] [/user/tsegaran/random/taxonomy\_subject/entry] [William A. Fraker]]}\\
    \texttt{[[Alternative rock] [/user/tsegaran/random/taxonomy\_subject/entry] [Atheism]]}\\
    \texttt{[[Spider-Man 3] [/user/tsegaran/random/taxonomy\_subject/entry] [Visual Effects]]}\\
    \texttt{[[Wong Jing] [/user/tsegaran/random/taxonomy\_subject/entry] [Artist-GB]]}\\
    \texttt{[[Blue Note Records] [/user/tsegaran/random/taxonomy\_subject/entry] [Robert Zemeckis]]}\\
    \texttt{[[African] [Caribbean \& Pacific Group of States] [/music/instrument/family] [The Kite Runner]]}\\
    \texttt{[[Gastroesophageal reflux disease] [/music/group\_member/membership] [United States Dollar]]}\\
    \texttt{[[Alou Diarra] [/people/deceased\_person/place\_of\_death] [Goalkeeper]]}\\
    \texttt{[[Oklahoma!] [/award/award\_nominated\_work] [Iggy Pop]]}\\
    \texttt{[[Writers Guild of America Award] [/organization/endowed\_organization/endowment] [Joe Cocker]]}\\
    \hline 
    \rowcolor[HTML]{EFEFEF} 
    \textsc{$i_{d_z}=10$}\\ \hline 
    \texttt{[[Academy Award for Best Actor] [/user/tsegaran/random/taxonomy\_subject/entry] [Greece]]}\\
    \texttt{[[University of Santo Tomas] [/user/tsegaran/random/taxonomy\_subject/entry] [Charleston]]}\\
    \texttt{[[Asian Development Bank] [/user/tsegaran/random/taxonomy\_subject/entry] [DVD]]}\\
    \texttt{[[Dead Ringers] [/organization/organization/place\_founded] [Allentown]]}\\
    \texttt{[[Academy Award for Best Short Film] [/location/administrative\_division] [Forward]]}\\
    \texttt{[[Anderson] [/base/popstra/celebrity/breakup] [Dancing with the Stars]]}\\
    \texttt{[[Dance-GB] [/user/tsegaran/random/taxonomy\_subject/entry] [High Noon]]}\\
    \texttt{[[FC Carl Zeiss Jena] [/music/genre/artists] [Comcast]]}\\
    \texttt{[[Dundee F.C.] [/government/political\_party/politicians\_in\_this\_party] [Kyushu]]}\\
    \texttt{[[Joan Collins] [/user/tsegaran/random/taxonomy\_subject/entry] [Olivia de Havilland]]}\\
    \hline
    \caption{Interpolation of each latent dimension for the RGVAE with $\delta=0.6$ and graph matching.}
\label{annexA:ipdim95Delta}
\end{longtable}

\begin{longtable}{|c|}
    % \centering
    \hline
    \rowcolor[HTML]{EFEFEF} 
    \textsc{$\delta$-RGVAE standard}\\ \hline
    \hline 
    \rowcolor[HTML]{EFEFEF} 
    \textsc{$i_{d_z}=1$}\\ \hline 
    \texttt{[[Nu metal] [/user/alexander/philosophy/philosopher/interests] [Hans Zimmer]]}\\
    \texttt{[[American Revolutionary War] [/influence/influence\_node/influenced\_by] [Monsters] [Inc.]]}\\
    \texttt{[[IFC Films] [/olympics/olympic\_sport/athletes] [Tony Bennett]]}\\
    \texttt{[[Barbra Streisand] [/music/performance\_role/track\_performances] [Michael Bay]]}\\
    \texttt{[[Avid Technology] [/people/person/religion] [Jules White]]}\\
    \texttt{[[Saturn Award for Best Costume] [/soccer/football\_team/current\_roster] [Nicole Kidman]]}\\
    \texttt{[[Grammy Award for Best Orchestral Performance] [/film/film/release\_date\_s] [Asheville]]}\\
    \texttt{[[Gameloft] [/people/ethnicity/people] [Hewlett-Packard]]}\\
    \texttt{[[Satish Kaushik] [/award/award\_category/winners] [Costa Rica]]}\\
    \texttt{[[Lawrence] [/military/military\_combatant/military\_conflicts] [Charleston]]}\\
    \hline 
    \rowcolor[HTML]{EFEFEF} 
    \textsc{$i_{d_z}=2$}\\ \hline 
    \texttt{[[Jackass: The Movie] [/music/performance\_role/regular\_performances] [Gotha]]}\\
    \texttt{[[Kingston] [/people/person/place\_of\_birth] [Carol Kane]]}\\
    \texttt{[[Eminem] [/film/film/release\_date\_s] [Sabrina]]}\\
    \texttt{[[The Man in the Iron Mask] [/education/educational\_institution/colors] [Oleic acid]]}\\
    \texttt{[[Beitar Jerusalem F.C.] [/award/award\_winning\_work/awards\_won] [Adolph Green]]}\\
    \texttt{[[Tony Gaudio] [/olympics/olympic\_sport/athletes] [Band of Brothers]]}\\
    \texttt{[[Heterosexuality] [/film/film/release\_date\_s] [2012]]}\\
    \texttt{[[Noam Chomsky] [/people/person/spouse\_s] [San Diego Chargers]]}\\
    \texttt{[[Mona Marshall] [/film/film/language] [Reno]]}\\
    \texttt{[[Hyde Park on Hudson] [/organization/organization/headquarters] [The Hurt Locker]]}\\
    \hline 
    \rowcolor[HTML]{EFEFEF} 
    \textsc{$i_{d_z}=3$}\\ \hline 
    \texttt{[[Wallace Beery] [/award/award\_nominee] [MGM Records]]}\\
    \texttt{[[1936 Summer Olympics] [/base/locations/continents/countries\_within] [Pete Postlethwaite]]}\\
    \texttt{[[Godzilla] [/common/topic/webpage] [Inkheart]]}\\
    \texttt{[[Tom Wilkinson] [/film/film/estimated\_budget] [Monroe County]]}\\
    \texttt{[[Chris Martin] [/location/country/second\_level\_divisions] [The Cassandra Crossing]]}\\
    \texttt{[[University of Dayton] [/people/person/languages] [2012 British Academy Film Awards]]}\\
    \texttt{[[Union of South Africa] [/film/film/personal\_appearances] [United Kingdom]]}\\
    \texttt{[[The Pursuit of Happyness] [/award/award\_winning\_work/awards\_won] [Ken Russell]]}\\
    \texttt{[[We Have a Pope] [/award/award\_nominee] [Steven Levitan]]}\\
    \texttt{[[Mindy Kaling] [/organization/organization/child] [Orange County]]}\\
    \hline 
    \rowcolor[HTML]{EFEFEF} 
    \textsc{$i_{d_z}=4$}\\ \hline 
    \texttt{[[Kevin Clash] [/education/educational\_degree/people\_with\_this\_degree] [LGBT]]}\\
    \texttt{[[Colombo] [/sports/sports\_position/players] [A Mighty Wind]]}\\
    \texttt{[[Vibraphone] [/government/legislative\_session/members] [Milton Keynes Dons F.C.]]}\\
    \texttt{[["Ryans Daughter"] [/film/film/language] [Malacca]]}\\
    \texttt{[[United States Department of the Treasury] [/film/film/release\_date\_s] [Jurassic Park]]}\\
    \texttt{[[Tomato] [/award/award\_nominee] [2 Fast 2 Furious]]}\\
    \texttt{[[Harmonica] [/people/person/places\_lived] [Claire Danes]]}\\
    \texttt{[[T. R. Knight] [/people/person/profession] [Male]]}\\
    \texttt{[[Rakesh Roshan] [/film/actor/film] [Legal drama]]}\\
    \texttt{[[Northrop Grumman] [/base/schemastaging/organization\_extra/phone\_number] [Talent show]]}\\
    \hline 
    \rowcolor[HTML]{EFEFEF} 
    \textsc{$i_{d_z}=5$}\\ \hline 
    \texttt{[[John Wood] [/film/film/release\_date\_s] [Changeling]]}\\
    \texttt{[[Gerry Conway] [/award/award\_ceremony/awards\_presented] [June Carter Cash]]}\\
    \texttt{[[College of William and Mary] [/music/group\_member/membership] [Northampton Town F.C.]]}\\
    \texttt{[[Nagesh] [/award/award\_category/winners] [T. Rajendar]]}\\
    \texttt{[[Bryce Dallas Howard] [/film/actor/film] [Tarzan]]}\\
    \texttt{[[UPN] [/organization/organization\_founder/organizations\_founded] [Travis Barker]]}\\
    \texttt{[[The Sum of All Fears] [/award/award\_winner/awards\_won] [Game of Thrones]]}\\
    \texttt{[[Ghent University] [/people/person/profession] [Khmer language]]}\\
    \texttt{[[Nine Network] [/people/deceased\_person/place\_of\_death] [Jim Steinman]]}\\
    \texttt{[[Hip hop music] [/base/marchmadness/ncaa\_basketball\_tournament/seeds] [Erykah Badu]]}\\
    \hline 
    \rowcolor[HTML]{EFEFEF} 
    \textsc{$i_{d_z}=6$}\\ \hline 
    \texttt{[[Mandolin] [/film/film/written\_by] [Atlantic Ocean]]}\\
    \texttt{[[Johann Wolfgang von Goethe] [/award/award\_category/winners] [Mrs. Parker]]}\\
    \texttt{[[Alexander Golitzen] [/music/genre/parent\_genre] [Dysentery]]}\\
    \texttt{[[Universal Motown Records] [/music/performance\_role/regular\_performances] [Psychology]]}\\
    \texttt{[[Tina Turner] [/sports/sports\_league\_draft/picks] [Bryan Forbes]]}\\
    \texttt{[[Carter Burwell] [/award/award\_nominee] [University College London]]}\\
    \texttt{[[Snowboarding] [/base/biblioness/bibs\_location/country] [Howards End]]}\\
    \texttt{[[Anthony Adverse] [/location/location/contains] [United States Dollar]]}\\
    \texttt{[[Socialism] [/film/film\_subject/films] [Male]]}\\
    \texttt{[[Salsa music] [/sports/professional\_sports\_team/draft\_picks] [Led Zeppelin]]}\\
    \hline 
    \rowcolor[HTML]{EFEFEF} 
    \textsc{$i_{d_z}=7$}\\ \hline 
    \texttt{[[Perth Airport] [/education/educational\_degree/people\_with\_this\_degree] [Anna Faris]]}\\
    \texttt{[[Suhasini Mani Ratnam] [/base/schemastaging/person\_extra/net\_worth] [Farmer-GB]]}\\
    \texttt{[[Academy Award for Best Visual Effects] [/film/film/written\_by] [The Deer Hunter]]}\\
    \texttt{[[Kenai Peninsula Borough] [/education/university/domestic\_tuition] [Linebacker]]}\\
    \texttt{[[Dinner for Schmucks] [/sports/sports\_position/players] [The Shubert Organization]]}\\
    \texttt{[[Bass guitar] [/sports/sports\_position/players] [Jonathan Demme]]}\\
    \texttt{[[Chapel Hill] [/music/instrument/instrumentalists] [Hartlepool]]}\\
    \texttt{[[Stanford University] [/music/record\_label/artist] [Miniseries]]}\\
    \texttt{[[Spain] [/education/educational\_institution/students] [Nobel Prize in Literature]]}\\
    \texttt{[[Maurice Gibb] [/award/award\_category/nominees] [Northampton]]}\\
    \hline 
    \rowcolor[HTML]{EFEFEF} 
    \textsc{$i_{d_z}=8$}\\ \hline 
    \texttt{[[Lea Michele] [/film/film/featured\_film\_locations] [Salisbury]]}\\
    \texttt{[[The Lovely Bones] [/award/award\_winning\_work/awards\_won] [Central European Time Zone-US]]}\\
    \texttt{[[Pulmonary embolism] [/people/person/spouse\_s] [Adventure Film]]}\\
    \texttt{[[Curb Your Enthusiasm] [/people/person/gender] [Montpellier HSC]]}\\
    \texttt{[[Damon Lindelof] [/people/person/spouse\_s] [Jonathan Forte]]}\\
    \texttt{[[Seattle] [/award/award\_winning\_work/awards\_won] [Entercom]]}\\
    \texttt{[[Christine Baranski] [/sports/sports\_position/players] [Seeking a Friend]]}\\
    \texttt{[[Working Girl] [/people/person/spouse\_s] [Imperial Japanese Army]]}\\
    \texttt{[[Marriage] [/soccer/football\_team/current\_roster] [Accountancy]]}\\
    \texttt{[[Nevada] [/people/person/spouse\_s] ["2002 NCAA Mens Division I Basketball Tournament"]]}\\
    \hline 
    \rowcolor[HTML]{EFEFEF} 
    \textsc{$i_{d_z}=9$}\\ \hline 
    \texttt{[[John Broome] [/education/educational\_institution/students] [The Spiderwick Chronicles]]}\\
    \texttt{[[Tom Ruegger] [/location/country/second\_level\_divisions] [58th Primetime Emmy Awards]]}\\
    \texttt{[[Sepultura] [/award/award\_nominee] [Homicide: Life on the Street]]}\\
    \texttt{[[The Prize] [/people/person/places\_lived] [Water polo]]}\\
    \texttt{[[Ellen Pompeo] [/award/award\_winner/awards\_won] [Belarus national football team]]}\\
    \texttt{[[Bachelor of Arts] [/music/performance\_role/regular\_performances] [1896 Summer Olympics]]}\\
    \texttt{[[FC Shakhtar Donetsk] [/people/person/place\_of\_birth] [2004 Summer Olympics]]}\\
    \texttt{[[Margot Kidder] [/film/actor/film] [The Mirror Has Two Faces]]}\\
    \texttt{[[California Golden Bears football] [/film/film/produced\_by] [Simon Cowell]]}\\
    \texttt{[[Alternative metal] [/people/person/profession] [New York City]]}\\
    \hline 
    \rowcolor[HTML]{EFEFEF} 
    \textsc{$i_{d_z}=10$}\\ \hline 
    \texttt{[[Cleopatra Records] [/sports/pro\_athlete/teams] [Soprano]]}\\
    \texttt{[[Tara Reid] [/business/job\_title/people\_with\_this\_title] [Newspaper]]}\\
    \texttt{[[Resident Evil] [/education/educational\_degree/people\_with\_this\_degree] [Shine]]}\\
    \texttt{[[Battle of Britain] [/film/film/genre] [Psychology]]}\\
    \texttt{[[Philips] [/music/genre/artists] [MTV Movie Award for Best Villain]]}\\
    \texttt{[[Anarcho-punk] [/people/person/profession] [Navy Blue]]}\\
    \texttt{[[Roger Daltrey] [/people/person/profession] [The Great Gatsby]]}\\
    \texttt{[[Liverpool F.C.] [/film/film/other\_crew] [al-Qaeda]]}\\
    \texttt{[[NCAA Mens Division I Basketball] [/music/instrument/instrumentalists] [Sinusitis]]}\\
    \texttt{[[Clive Owen] [/education/university/local\_tuition] [72nd Academy Awards]]}\\
    \hline
\caption{Interpolation of each latent dimension for the RGVAE with $\delta=0.6$ and standard loss.}
\label{annexA:ipdim95NoPermDelta}
\end{longtable}
    
\end{appendices}

\end{document}